%% file: main.tex
\documentclass[letter,english]{article}

\usepackage[margin=1in]{geometry}

\usepackage[utf8]{inputenc} 
\usepackage[T1]{fontenc}    

\usepackage[unicode=true,
 bookmarks=false,
 breaklinks=false,pdfborder={0 0 1},colorlinks=false]
 {hyperref}
\hypersetup{
 colorlinks,citecolor=blue,filecolor=blue,linkcolor=blue,urlcolor=blue}
\usepackage{url}            
\usepackage{booktabs}       
\usepackage{amsfonts}       
\usepackage{nicefrac}       
\usepackage{microtype}      
\usepackage{xcolor}         
\usepackage{bm}
\usepackage{amsmath}
\usepackage{amssymb}

\usepackage{multirow}
\usepackage{colortbl}
\definecolor{Gray}{gray}{0.85}
\usepackage{enumitem}

\usepackage{amsthm}
\usepackage{cite}  
\usepackage{comment}
\usepackage{natbib}
\usepackage{mathtools}

\usepackage{graphicx}
\usepackage[linesnumbered,ruled,vlined]{algorithm2e}
\usepackage{algorithmic}

\usepackage{float}
\usepackage{cancel}
\usepackage{pifont}
\usepackage{mathrsfs}
\usepackage{bbm}

\usepackage{subcaption}
\usepackage{wrapfig}

\usepackage{dsfont}
\usepackage{color}
\definecolor{yjc}{RGB}{125,0,0}
\definecolor{jiw}{RGB}{10,148,15}
\definecolor{lxs}{RGB}{138,43,226}
 \renewcommand{\hat}{\widehat}
 \renewcommand{\tilde}{\widetilde}
  
\newcommand{\RE}{\normalfont{\textsc{RefineEstimate}}}
\newcommand{\DVR}{\textsf{Fed-DVR-Q}}

\newcommand{\hatQ}{\widehat{Q}}
\newcommand{\barQ}{\overline{Q}}
\newcommand{\hatV}{\widehat{V}}

\newcommand{\hatP}{\widehat{P}}
\newcommand{\barDelta}{\overline{\Delta}}
\newcommand{\hatH}{\widehat{\mathcal{H}}}
\newcommand{\hatT}{\widehat{\mathcal{T}}}
\newcommand{\tildeT}{\widetilde{\mathcal{T}}}
\newcommand{\starQ}{Q^{\star}}
\newcommand{\starQH}{Q^{\star}_{\mathcal{H}}}

\newcommand{\bfzero}{\mathbf{0}}
\newcommand{\SC}{\textsf{SC}}
\newcommand{\CC}{\textsf{CC}}
\newcommand{\ER}{\textsf{ER}}

\allowdisplaybreaks

\input{macros}
\title{The Sample-Communication Complexity Trade-off \\ in Federated Q-Learning}
 \author{
 	Sudeep Salgia\thanks{
        Department of Electrical and Computer Engineering,
        Carnegie Mellon University;
        \texttt{\{ssalgia,yuejiec\}@andrew.cmu.edu}.}   \\
	Carnegie Mellon University
	\and  
	Yuejie Chi\footnotemark[1]\\
        Carnegie Mellon University
 	} 
\date{August 2024; Revised October 2024}

\begin{document}

\theoremstyle{plain} \newtheorem{lemma}{\textbf{Lemma}}
\newtheorem{proposition}{\textbf{Proposition}}
\newtheorem{theorem}{\textbf{Theorem}}
\newtheorem{corollary}{\textbf{Corollary}}
\newtheorem{assumption}{Assumption}
\newtheorem{definition}{Definition}
\newtheorem{claim}{\textbf{Claim}}
\theoremstyle{remark}\newtheorem{remark}{\textbf{Remark}}

\maketitle

\begin{abstract}
    \input{abstract}
\end{abstract}

\noindent\textbf{Keywords:} federated Q-learning, communication complexity, sample complexity, trade-offs

\setcounter{tocdepth}{2}
\tableofcontents

\input{introduction}

\input{problem_formulation}
\input{lower_bound}
\input{algorithm}

\input{simulations}
\input{conclusion}

\section*{Acknowledgement}
This work is supported in part by the grants NSF CCF-2007911, CCF-2106778, CNS-2148212, ECCS-2318441, ONR N00014-19-1-2404 and AFRL FA8750-20-2-0504, and in part by funds from federal agency and industry partners as specified in the Resilient \& Intelligent NextG Systems (RINGS) program.

\bibliographystyle{abbrvnat}
\bibliography{references,bibfileRL}

\newpage

\appendix

\input{appendix}

\input{analysis_upper_bound}

\end{document}

%% file: macros.tex
\newcommand{\E}{\mathbb{E}}

\newcommand{\N}{\mathbb{N}}

\newcommand{\R}{\mathbb{R}}

\newcommand{\cA}{\mathcal{A}}

\newcommand{\cC}{\mathcal{C}}

\newcommand{\cH}{\mathcal{H}}
\newcommand{\cI}{\mathcal{I}}

\newcommand{\cM}{\mathcal{M}}

\newcommand{\cO}{\mathcal{O}}

\newcommand{\cS}{\mathcal{S}}
\newcommand{\cT}{\mathcal{T}}

\newcommand{\cZ}{\mathcal{Z}}

\newcommand{\bfsigma}{\boldsymbol{\sigma}}

\newcommand{\sA}{\mathscr{A}}

\newcommand{\sC}{\mathscr{C}}

\newcommand{\sF}{\mathscr{F}}
\newcommand{\sG}{\mathscr{G}}

\DeclareMathOperator*{\argmax}{arg\,max}

\newcommand{\1}{\mathbbm{1}}

\newcommand{\var}{\text{Var}}

%% file: abstract.tex
We consider the problem of federated Q-learning, where $M$ agents aim to collaboratively learn the optimal Q-function of an unknown infinite-horizon Markov decision process with finite state and action spaces. We investigate the trade-off between sample and communication complexities for the widely used class of intermittent communication algorithms. We first establish the converse result, where it is shown that a federated Q-learning algorithm that offers any speedup with respect to the number of agents in the per-agent sample complexity needs to incur a communication cost of at least an order of $\frac{1}{1-\gamma}$ up to logarithmic factors, where $\gamma$ is the discount factor. We also propose a new  algorithm, called \DVR, which is the first federated Q-learning algorithm to simultaneously achieve order-optimal sample and communication complexities. Thus, together these results provide a complete characterization of the sample-communication complexity trade-off in federated Q-learning.
  

%% file: introduction.tex
\section{Introduction}

Reinforcement Learning (RL)~\citep{Sutton2018RLBook} refers to a paradigm of sequential decision making where an agent aims to learn an optimal policy, i.e., a policy that maximizes the long-term total reward, through repeated interactions with an unknown environment. RL finds applications across a diverse array of fields including, but not limited to, autonomous driving, games, recommendation systems, robotics and Internet of Things (IoT)~\citep{Kober2013RoboticsSurvey, Yurtsever2020SelfDriving, Huang2016Go, Lim2020IoT}.  

The primary hurdle in RL applications is often the high-dimensional nature of the decision space that necessitates the learning agent to have to access to an enormous amount of data in order to have any hope of learning the optimal policy. Moreover, the sequential collection of such an enormous amount of data through a single agent is extremely time-consuming and often infeasible in practice \citep{mnih2016a3c}. Consequently, practical implementations of RL involve deploying multiple agents to collect data in parallel. This decentralized approach to data collection has fueled the design and development of distributed or federated RL algorithms that can collaboratively learn the optimal policy without actually transferring the collected data to a centralized server, while achieving a linear speedup in terms of the number of agents. Such a federated approach to RL, which does not require the transfer of local data, is gaining interest due to lower bandwidth requirements and lower security and privacy risks.
In this work, we focus on federated variants of the vastly popular Q-learning algorithm \citep{Watkins1992QL}, where the agents collaborate to directly learn the optimal Q-function without forming an estimate of the underlying unknown environment. 

A particularly important aspect of designing federated RL algorithms, including federated Q-learning algorithms, is to address the natural tension between sample and communication complexities. At one end of the spectrum lies the na\"ive approach of running a centralized algorithm with an optimal sample complexity after transferring and combining all the collected data at a central server. Such an approach trivially achieves the optimal sample complexity while suffering from a very high and prohibitive communication complexity. On the other hand, several recently proposed algorithms~\citep{Khodadadian2022FederatedQL, Woo2023FedSynQ,Woo2024FedOffline,zheng2024federated} operate in more practical regimes, offering significantly lower communication complexities when compared to the na\"ive approach at the cost of sub-optimal sample complexities. These results suggest the existence of an underlying trade-off between sample and communication complexities of federated RL algorithms. The primary goal of this work is to better understand this trade-off in the context of federated Q-learning by investigating these following fundamental questions.  
\begin{itemize}
\item {\em Fundamental limit of communication:}
{What is the minimum amount of communication required by a federated Q-learning algorithm to achieve any statistical benefit of collaboration?} 

\item {\em Optimal algorithm design:}
{How does one design a federated Q-learning algorithm that simultaneously offers optimal sample and communication complexity guarantees, i.e., operates on the optimal frontier of the sample-communication complexity trade-off?}
\end{itemize}

\subsection{Main results}

In this work, we consider a setup where $M$ distributed agents --- each with access to a local generative model \citep{Kearns1998FiniteSampleQL} --- collaborate to learn the optimal Q-function of an infinite-horizon Markov decision process (MDP) \citep{Puterman2014MDPBook}, which is defined over a finite state space $\cS$ and a finite action space $\cA$, and has a discount factor of $\gamma \in (0,1)$. To probe the communication complexity, we consider a common setup in federated learning called the intermittent communication setting \citep{Woodworth2021IntermittentComm}, where the agents intermittently share information among themselves with the help of a central server.
We provide a complete characterization of the trade-off between sample and communication complexities under the aforementioned setting by providing answers to both the questions. Summarized below, the main result of this work is twofold.
\begin{itemize}
    \item \textit{Fundamental lower bounds on the communication complexity of federated Q-learning.} We establish lower bounds on the communication complexity of federated Q-learning, both in terms of the number of communication rounds and the overall number of bits that need to be transmitted in order to achieve any {\em speedup} in the convergence rate with respect to the number of agents. Specifically, we show that in order for an intermittent communication algorithm to obtain \emph{any} benefit of collaboration, i.e., \emph{any} order of speedup with respect to the number of agents, the number of communication rounds must be least $\Omega \left( \frac{1}{(1- \gamma) \log^2 N} \right)$ and the number of \emph{bits} sent by each agent to the server must be least $\Omega \left( \frac{|\cS| |\cA|}{(1- \gamma) \log^2 N} \right)$, where $N$ denotes the number of samples taken by the algorithm for each state-action pair.
 
    \item \textit{Achieving the optimal sample-communication complexity trade-off of federated Q-learning.} We propose a new federated Q-learning algorithm called Federated Doubly Variance Reduced Q-Learning (dubbed \DVR), that simultaneously achieves order-optimal sample complexity and communication complexity as dictated by the lower bound. We show that \DVR \ learns an  $\varepsilon$-accurate optimal Q-function in the $\ell_{\infty}$ sense with $\tilde{\cO}\left(\frac{|\cS| |\cA|}{M\varepsilon^2 (1-\gamma)^3}\right)$ i.i.d. samples from the generative model at each agent while incurring a total communication cost of $\tilde{\cO}\left( \frac{|\cS| |\cA|}{ 1-\gamma}\right)$ \emph{bits} per agent across $\tilde{\cO}\left( \frac{1}{1-\gamma}\right)$ rounds of communication. Thus, \DVR \ not only improves upon both the sample and communication complexities of existing algorithms, but also is the \emph{first algorithm} to achieve order-optimal sample and communication complexities (See Table~\ref{table:comparison_with_others} for a comparison). 
\end{itemize}

\begin{table*}[t]
\centering
    \begin{tabular}{c|c|c|c}
    \toprule
     \multirow[c]{2}{*}{Algorithm/Reference}   & \multirow[c]{2}{*}{\shortstack{Number of\\ Agents}} & \multirow{2}{*}{\shortstack{Sample \\ Complexity}}  & \multirow{2}{*}{\shortstack{Communication \\ Complexity}}  \\
    &  &  &  \\
    \midrule
    Q-learning \citep{Li2023QLMinimax} & $1$ & $\dfrac{|\cS| |\cA|}{(1-\gamma)^4 \varepsilon^2}$ & N/A \\
    \midrule
    Variance Reduced Q-learning \citep{Wainwright2019VarianceReduced} & $1$ & $\dfrac{|\cS| |\cA|}{(1-\gamma)^3 \varepsilon^2}$ & N/A \\
    \midrule
    Fed-SynQ \citep{Woo2023FedSynQ} & $M$ & $\dfrac{|\cS| |\cA|}{M(1-\gamma)^5 \varepsilon^2}$ & $\dfrac{M}{1-\gamma}$ \\
    \midrule
    \DVR \ (\textbf{This work}) & $M$ & $\dfrac{|\cS| |\cA|}{M(1-\gamma)^3 \varepsilon^2}$ & $\dfrac{1}{1-\gamma}$ \\
    \midrule
    Lower bound (\citep{Azar2013MinimaxPAC}, \textbf{This work}) & $M$ & $\dfrac{|\cS| |\cA|}{M(1-\gamma)^3 \varepsilon^2}$ & $\dfrac{1}{1-\gamma}$ \\
     \bottomrule
    \end{tabular}
    \caption{Comparison of sample and communication complexities (per-agent) of various single-agent and federated Q-learning algorithms for learning an $\varepsilon$-optimal Q-function under the synchronous setting. We hide logarithmic factors and burn-in costs for all results for simplicity of presentation. Here, $\cS$ and $\cA$ represent state and action spaces respectively and $\gamma$ denotes the discount factor. We report the communication complexity only in terms of the number of rounds as existing algorithms assume transmission of real numbers and hence do not report bit-level costs. For the lower bound,~\cite{Azar2013MinimaxPAC} and this work establish the lower bounds for the sample and communication complexities, respectively.  
    }
    \label{table:comparison_with_others}
\end{table*}

\subsection{Related work}

\paragraph{Single-agent Q-learning.} Q-learning has been extensively studied in the single-agent setting in terms of its asymptotic convergence~\citep{Jaakkola1993StochasticIterativeDP, Tsitsiklis1994AsynchronousQL, Szepesvari1997AsymptoticQL, Borkar2000ODE-RL} and its finite-time sample complexity in  synchronous~\citep{EvenDar2004LearningRates, Kearns1998FiniteSampleQL, Beck2012ConstantStepSize, sidford2018near, Wainwright2019ConeContractive, Wainwright2019VarianceReduced, Chen2020FiniteSampleAnalysis, Li2023QLMinimax} and asynchronous settings~\citep{Chen2021Lyapunov, Li2023QLMinimax, Li2021SampleComplexityAsynchronousQL, Qu2020FiniteAsychronousQL,xia2024instance} in terms of convergence in the $\ell_\infty$ sense. On the other hand, regret analysis of Q-learning has been carried out in both online settings~\citep{jin2018q,bai2019provably,menard2021ucb,li2021breaking} and offline settings~\citep{shi2022pessimistic,yan2022efficacy}, to name a few.

\paragraph{Federated and distributed RL.} There has also been a considerable effort towards developing distributed and federated RL algorithms. The distributed variants of the classical temporal difference (TD) learning algorithm have been investigated in a series of studies~\citep{Chen2021OffPolicyTDC, Doan2019TD0LinearFunctionMARL, Doan2021DistributedTD0, Sun2020FiniteDecentralizedTDLinear, Wai2020ConvergenceConsensus, liu2023distributed, Wang2020DecentralizedTDTracking, Zeng2021FiniteDecentralizedSA}. The impact of environmental heterogeneity in federated RL was studied in~\cite{Wang2023FederatedTDHeterogeneity} for TD learning, and in \cite{Jin2022FederatedRLHeterogeneity} when the local environments are known. A distributed version of the actor-critic algorithm was studied by~\citet{Shen2023AsynchronousAdvantageActorCritic} where the authors established convergence of their algorithm and demonstrated a linear speedup in the number of agents in their sample complexity bound.~\citet{Chen2022DecentralizedActorCritic} proposed a new distributed actor-critic algorithm which improved the dependence of sample complexity on the error $\varepsilon$ with a communication cost of $\tilde{\cO}(\varepsilon^{-1})$.~\citet{Chen2021CommEfficientPolicyGrad} proposed a communication-efficient distributed policy gradient algorithm with convergence analysis and established a communication complexity of $\cO(1/(M\varepsilon))$. ~\citet{Xie2023FedKL} adopts a distributed policy optimization perspective, which is different from the Q-learning paradigm considered in this work. Moreover, the algorithm in~\citet{Xie2023FedKL} obtains a linear communication cost, which is worse than that obtained in our work. Similarly,~\citet{Zhang2024FiniteTimeOnPolicy} focuses on on-policy learning and incurs a communication cost that depends polynomially on the required error $\varepsilon$. Several additional studies~\citep{Yang2023FederatedNPG, Zeng2021DecentralizedNPG, Lan2024AsynchronousPG} have also developed and analyzed other distributed/federated variants of the classical natural policy gradient method~\citep{Kakade2001PolicyGradient}.~\citet{Assran2019GossipActorCritic, Espeholt2018Impala, Mnih2016AsynchronousDeepRL} developed distributed algorithms to train deep RL networks more efficiently.  

\paragraph{Federated Q-learning.} Federated Q-learning has been explored relatively recently with theoretical sample and communication complexity guarantees.~\citet{Khodadadian2022FederatedQL} proposed and analyzed a federated Q-learning algorithm in the asynchronous setting, however, its sample complexity guarantee exhibits pessimistic dependencies with respect to salient problem-dependent parameters. 
\cite{Woo2023FedSynQ} provided improved analyses for federated Q-learning under both synchronous and asynchronous settings, and introduced importance averaging to tame the heterogeneity of local behavior policies in the asynchronous setting to further improve the sample complexity, showing that a collaborative coverage of the entire state-action space suffices for federated Q-learning. Moving to the offline setting, \cite{Woo2024FedOffline} proposed a federated Q-learning algorithm for offline RL in the finite-horizon setting and established sample and communication complexities that only require a collaborative coverage of the state-action pairs visited by the optimal policy. \cite{zheng2024federated}, on the other hand, established a linear speedup for federated Q-learning in the online setting from the regret minimization perspective.

\paragraph{Accuracy-communication trade-off in federated learning.} The trade-off between communication complexity and accuracy (or equivalently, sample complexity) has been studied in various federated and distributed learning problems, including stochastic approximation algorithms for convex optimization.~\cite{Duchi2014DistributedEstimation, Braverman2016CommunicationLowerBounds} established the celebrated inverse linear relationship between the error and the communication cost for the problem of distributed mean estimation. Similar trade-offs for distributed stochastic optimization, multi-armed bandits and linear bandits have been studied and established across numerous studies, e.g., \citep{Woodworth2018GraphOracle, Woodworth2021IntermittentComm, Tsitsiklis1987, Shi2021FMAB, Salgia2023LinearBandits}.


%% file: problem_formulation.tex
\section{Background and Problem Formulation}

In this section, we provide a brief background of MDPs, outline the performance measures for federated Q-learning algorithms and describe the class of intermittent communication algorithms considered in this work.

\subsection{Markov decision processes}
\label{sub:MDP}

We focus on an infinite-horizon MDP, denoted by $\cM$, over a state space $\cS$ and an action space $\cA$ with a discount factor $\gamma \in (0,1)$. Both the state and action spaces are assumed to be finite sets. In an MDP, the state $s$ evolves dynamically under the influence of actions based on a probability transition kernel, $P : (\cS \times \cA) \times \cS \mapsto [0,1]$. The entry $P(s'|s,a)$ denotes the probability of moving to state $s'$ when an action $a$ is taken in state $s$. An MDP is also associated with a deterministic reward function $r : \cS \times \cA \mapsto [0,1]$, where $r(s,a)$ denotes the immediate reward obtained for taking action $a$ in state $s$. Thus, the transition kernel $P$ along with the reward function $r$ completely characterize an MDP.

A policy $\pi : \cS \mapsto \Delta(\cA)$ is a rule for selecting actions across different states, where $\Delta(\cA)$ denotes the simplex over $\cA$ and $\pi(a|s)$ denotes the probability of choosing action $a$ in state $s$. Each policy $\pi$ is associated with a state value function and a state-action value function, or the Q-function, denoted by $V^{\pi}$ and $Q^{\pi}$ respectively. Specifically, $V^{\pi}$ and $Q^{\pi}$ measure the expected discounted cumulative reward attained by policy $\pi$ starting from certain state $s$ and state-action pair $(s,a)$ respectively.
Mathematically, $V^{\pi}$ and $Q^{\pi}$ are given as
\begin{align}
    V^{\pi}(s) := \E\left[ \sum_{t = 0}^{\infty}\gamma^{t}r(s_t, a_t) \bigg|\ s_0 = s\right]; \quad 
    Q^{\pi}(s,a) := \E\left[ \sum_{t = 0}^{\infty}\gamma^{t}r(s_t, a_t) \bigg|\ s_0 = s, a_0 = a\right], \label{eqn:q_fn_defn}
\end{align}
where $a_t \sim \pi(\cdot | s_t)$, $s_{t+1} \sim P( \ \cdot \ | s_t, a_t)$ for all $t \geq 0$, and the expectation is taken w.r.t. the randomness in the trajectory $\{s_t, a_t\}_{t =1}^{\infty}$.
Since the rewards lie in $[0,1]$, it follows immediately that both the value function and the Q-function lie in the range $[0, \frac{1}{1-\gamma}]$.

An optimal policy $\pi^{\star}$ is a policy that maximizes the value function uniformly over all the states and it has been shown that such an optimal policy $\pi^{\star}$ always exists~\citep{Puterman2014MDPBook}. The optimal value and Q-functions are those corresponding to that of an optimal policy $\pi^{\star}$, denoted as $V^{\star} := V^{\pi^{\star}}$ and $Q^{\star} := Q^{\pi^{\star}}$ respectively. The optimal Q-function, $Q^{\star}$, is also the unique fixed point of the {\em Bellman optimality operator} $\cT : \cS \times \cA \mapsto \cS \times \cA$, given by
\begin{align}
\cT Q^{\star} = Q^{\star}, \qquad \mbox{where}\quad    (\cT Q)(s,a) = r(s,a) + \gamma \cdot \E_{s' \sim P( \cdot | s,a)}\left[\max_{a'\in \cA} Q(s',a')\right]. \label{eqn:bellman_operator}
\end{align}
The popular Q-learning algorithm~\citep{Watkins1992QL} aims to learn the optimal policy by first learning $Q^{\star}$ as the solution to the above fixed-point equation --- via stochastic approximation  when $\cT$ is only accessed through samples --- and then obtaining a deterministic optimal policy via greedy action selection, i.e., $\pi^{\star}(s) = \argmax_{a} \starQ(s,a)$.

\subsection{Performance measures in federated Q-learning}

We consider a federated learning setup consisting of $M$ agents, where all the agents face a common, unknown MDP, i.e., the transition kernel and the reward function are the same across agents. In addition, we consider the synchronous setting \citep{Wainwright2019ConeContractive}, where each agent has access to an independent generative model or simulator from which they can draw independent samples from the unknown underlying distribution $P(\cdot|s,a)$ for all state-action pair $(s,a)$~\citep{Kearns1998FiniteSampleQL} simultaneously. Let $Z \in \cS^{|\cS| |\cA|}$ be a corresponding random vector whose $(s,a)$-th coordinate is drawn from the distribution $P( \cdot | s,a)$, independently of all other coordinates. We define the random operator $ \hat{\cT}_{Z} : (\cS \times \cA) \mapsto (\cS \times \cA)$ as   
\begin{align}
    ( \hat{\cT}_{Z} Q)(s,a) = r(s,a) + \gamma V(Z(s,a)),
\end{align}
where $V(s') = \max_{a'\in \cA} Q(s',a')$. The operator $\hat{\cT}_{Z}$ can be interpreted as the sample Bellman operator, where we have the relation $\cT Q = \E_{Z}\big[\hat{\cT}_Z Q\big]$ for all Q-functions.  
For a given value of $\varepsilon \in (0,\frac{1}{1-\gamma})$, the objective of agents is to collaboratively learn an $\varepsilon$-optimal estimate (in the $\ell_{\infty}$ sense) of the optimal Q-function of the unknown MDP.

We measure the performance of a federated Q-learning algorithm $\sA$ using two metrics --- sample complexity and communication complexity. For a given MDP $\cM$, let $\hatQ_{\cM}(\sA, N, M)$ denote the estimate of $\starQ_{\cM}$, the optimal Q-function of the MDP $\cM$, returned by an algorithm $\sA$, when given access to $N$ i.i.d. samples from the generative model for each $(s,a)$ pair at all the $M$ agents. The error rate of the algorithm $\sA$, denoted by $\ER(\sA; N, M)$, is defined as
\begin{align}
    \ER(\sA; N, M) := \sup_{\cM  } \E\left[\big\|\hatQ_{\cM}(\sA, N, M) - \starQ_{\cM} \big\|_{\infty}\right],
\end{align}
where the expectation is taken over the samples and any randomness in the algorithm. Given a value of $\varepsilon  \in (0, \frac{1}{1 -\gamma}) $, the sample complexity of $\sA$, denoted by $\SC(\sA; \varepsilon, M)$ is given by
\begin{align}
    \SC(\sA; \varepsilon, M) := |\cS| |\cA| \cdot \min \left\{N \in \N: \ER(\sA; N, M) \leq \varepsilon \right\}.
\end{align}
Similarly, we can also define a high-probability version for any $\delta \in (0,1)$ as follows:
\begin{align*}
    \SC(\sA; \varepsilon, M, \delta) := |\cS| |\cA| \cdot \min\left\{N \in \N:  \Pr \left( \sup_{\cM} \|\hatQ_{\cM}(\sA, N, M) - \starQ_{\cM}\|_{\infty} \leq \varepsilon \right) \geq 1 -\delta \right\}.
\end{align*}
We measure the communication complexity of any federated learning algorithm both in terms of the frequency of information exchange and the total number of bits uploaded by the agents. For each agent $m$, let $C_{\textsf{round}}^m(\sA; N)$ and $C_{\textsf{bit}}^m(\sA; N)$ respectively denote the number of times agent $m$ sends a message, and, the total number of bits uploaded by agent $m$ to the server when an algorithm $\sA$ is run with $N$ i.i.d. samples from the generative model for each $(s,a)$ pair at all the $M$ agent. The communication complexity of $\sA$, when measured in terms of the frequency of communication and the total number of bits exchanged, is given by
\begin{align}
    \CC_{\textsf{round}}(\sA; N)  :=  \frac{1}{M} \sum_{m = 1}^M C_{\textsf{round}}^m(\sA; N); \quad
    \CC_{\textsf{bit}}(\sA; N)  := \frac{1}{M} \sum_{m = 1}^M C_{\textsf{bit}}^m(\sA; N), 
\end{align}
respectively. Similarly, for a given value of $\varepsilon \in (0, \frac{1}{1 -\gamma})$, we can also define $\CC_{\textsf{round}}(\sA; \varepsilon)$ and $\CC_{\textsf{bit}}(\sA; \varepsilon)$  when $\sA$ is run to guarantee an error of at most $\varepsilon$, as well as the high-probability version for any $\delta \in (0,1)$ as $\CC_{\textsf{round}}(\sA; \varepsilon, \delta)$ and $\CC_{\textsf{bit}}(\sA; \varepsilon, \delta)$.


\subsection{Intermittent-communication algorithm protocols}

We consider a popular class of federated learning algorithms with intermittent communication. The intermittent communication setting provides a natural framework to extend single-agent Q-learning algorithms to the distributed setting. As the name suggests, under this setting, the agents intermittently communicate with each other or a central server, sharing their updated beliefs about $\starQ$. Between two communication rounds, each agent updates their belief about $\starQ$ using stochastic approximation iterations based on the locally available data, similar to a single-agent setup.
Such intermittent communication algorithms have been extensively studied and used to establish lower bounds on communication complexity of distributed stochastic convex optimization~\citep{Woodworth2018GraphOracle, Woodworth2021IntermittentComm}. 
 
\begin{algorithm}[ht]
    \caption{A generic federated Q-learning algorithm $\sA$}
    \label{alg:general_fed_alg_with_inter_comm}
    \begin{algorithmic}[1]
        \STATE Input : $T, R, \{\eta_t\}_{t = 1}^T, \cC = \{t_r\}_{r = 1}^R, B$
        \STATE Set $Q_0^m = 0$ for all agents $m$
        \FOR{$t = 1,2,\dots, T$}
        \FOR{$m = 1,2,\dots, M$}
        \STATE Compute $Q_{t-\frac{1}{2}}^m$ according to Eqn.~\eqref{eqn:generic_algo_update}
        \STATE Compute $Q_{t}^m$ according to Eqn.~\eqref{eqn:generic_algo_averaging}
        \ENDFOR
        \ENDFOR
        \STATE \textbf{return} $Q_T$
    \end{algorithmic}
\end{algorithm}

A generic federated Q-learning algorithm with intermittent communication is outlined in Algorithm~\ref{alg:general_fed_alg_with_inter_comm}. It is characterized by the following five parameters: (i) the total number of updates $T$; (ii) the number of communication rounds $R$; (iii) a step size schedule $ \{\eta_t\}_{t = 1}^T$; (iv) a communication schedule $\cC = \{t_r\}_{r= 1}^R$; (v) batch size $B$. 
During the $t$-th iteration, each agent $m$ computes $\{\widehat{\cT}_{Z_b}(Q_{t-1}^{m})\}_{b = 1}^B$, a minibatch of sample Bellman operators at the current estimate $Q_{t-1}^m$, using $B$ samples from the generative model for each $(s,a)$ pair, and obtains an intermediate local estimate using the Q-learning update rule as follows:
\begin{align}
    Q_{t-\frac{1}{2}}^m = (1- \eta_t)Q_{t-1}^m + \frac{\eta_t}{B} \sum_{b = 1}^B \widehat{\cT}_{Z_b}(Q_{t-1}^{m}). \label{eqn:generic_algo_update}
\end{align}
Here, $\eta_t \in (0,1]$ is the step size chosen corresponding to the $t$-th time step. The intermediate estimates are averaged based on a communication schedule $\cC = \{t_r\}_{r = 1}^R$ consisting of $R$ rounds, i.e., 
\begin{align}
    Q_{t}^m = \begin{cases} \frac{1}{M} \sum_{j = 1}^M Q_{t-\frac{1}{2}}^j & \text{ if } t \in \cC, \\ Q_{t-\frac{1}{2}}^m & \text{otherwise.} \end{cases}
    \label{eqn:generic_algo_averaging}
\end{align}
In the above equation, the averaging step can also be replaced with any distributed mean estimation routine that includes compression to control the bit level costs. Without loss of generality, we assume that $Q^m_0 = 0$ for all agent and $t_R = T$, i.e., the last iterates are always averaged. It is straightforward to note that the number of samples taken per agent by an intermittent communication algorithm is $BT$, i.e, $N = BT$ and the communication complexity $\CC_{\textsf{round}} (\sA; N) = R$.

%% file: lower_bound.tex
\section{Communication Complexity Lower Bound}
\label{sec:lower_bound}

In this section, we investigate the first of the two questions regarding the lower bound on communication complexity. The following theorem establishes a lower bound on the communication complexity of a federated Q-learning algorithm with intermittent communication as described in Algorithm~\ref{alg:general_fed_alg_with_inter_comm}.

\begin{theorem}
    Assume that $\gamma \in [5/6, 1)$ and the state and action spaces satisfy $|\cS| \geq 4$ and $|\cA| \geq 2$. Let $\sA$ be a federated Q-learning algorithm with intermittent communication (as described in Algorithm~\ref{alg:general_fed_alg_with_inter_comm}) that is run for $T \geq \max\{16,\frac{1}{1-\gamma}\}$ steps with a step size schedule of either $\eta_t := \frac{1}{1 + c_{\eta}(1-\gamma)t}$ or $\eta_t := \eta$ for all $ 1\leq t \leq T$. If
    \begin{align*}
        R = \CC_{\textsf{round}}(\sA; N) \leq \frac{c_0}{(1- \gamma) \log^2 N}; \text{ or } \CC_{\textsf{bit}}(\sA; N) \leq \frac{c_1 |\cS| |\cA|}{(1- \gamma) \log^2 N}
    \end{align*}
    for some universal constants $c_0, c_1 > 0$, then for all choices of communication schedule, batch size $B$, $c_{\eta} > 0$ and $\eta \in (0,1)$, the error of $\sA$ satisfies
    \begin{align*}
        \ER(\sA; N, M) \geq \frac{C_{\gamma}}{\sqrt{N} \log^3 N}
    \end{align*}
    for all $M \geq 2$ and $N = BT$. Here $C_{\gamma} > 0$ is a constant that depends only on $\gamma$.
    \label{thm:lower_bound}
\end{theorem}

    


The above theorem states that for any federated Q-learning algorithm with intermittent communication  to obtain \emph{any} benefit of collaboration, i.e., for the error rate $\ER(\sA; N, M)$ to decrease w.r.t. the number of agents, it  must have at least $\Omega \left( \frac{1}{(1- \gamma)\log^2 N } \right)$ rounds of communication and transmit $\Omega \left( \frac{|\cS| |\cA|}{(1- \gamma) \log^2 N }\right)$ bits of information per agent, both of which scale inverse proportionally to the effective horizon $\frac{1}{1-\gamma}$ of the MDP. The above lower bound on the communication complexity also immediately applies to federated Q-learning algorithms that offer order-optimal sample complexity, and thereby a linear speedup with respect to the number of agents.
Therefore, this characterizes the converse relation for the sample-communication tradeoff in federated Q-learning. 

The above lower bound on the communication complexity of federated Q-learning is a consequence of the bias-variance trade-off that governs the convergence of the Q-learning algorithm. While a careful choice of step sizes alone is sufficient to balance this trade-off in the centralized setting, the choice of communication schedule also plays an important role in balancing this trade-off in the federated setting. The local steps between two communication rounds induce a positive estimation bias that depends on the standard deviation of the iterates and is a well-documented issue of ``over-estimation'' in Q-learning~\citep{Hasselt2010DoubleQL}. Since such a bias is driven by \emph{local} updates, it does not reflect any benefit of collaboration. During a communication round, the averaging of iterates across agents allows the algorithm an opportunity to counter this bias by reducing the effective variance of the updates through averaging. In our analysis, we show that if the communication is infrequent, the local bias becomes the dominant term and averaging of iterates is insufficient to counter the impact of the positive bias induced by the local steps. As a result, we do not observe any statistical gains when the communication is infrequent. Our argument is inspired by the analysis of Q-learning in~\cite{Li2023QLMinimax} and is based on analyzing the convergence of an intermittent communication algorithm on a specifically chosen ``hard'' MDP instance. The detailed proof is deferred to Appendix~\ref{appendix:proof_lower_bound}.

\begin{remark}[Communication complexity of policy evaluation] Several recent studies \citep{liu2023distributed,tian2024one} established that a single round of communication is sufficient to achieve linear speedup of TD learning for {\em policy evaluation}, which do not contradict with our results focusing on Q-learning for {\em policy learning}. The latter is more involved due to the nonlinearity of the Bellman optimality operator. Specifically, if the operator whose fixed point is to be found is linear in the decision variable (e.g., the value function in TD learning) then the fixed point update only induces a variance term corresponding to the noise. However, if the operator is non-linear, then in addition to the variance term, we also obtain a \emph{bias} term in the fixed point update. While the variance term can be controlled with one-shot averaging, more frequent communication is necessary to ensure that the bias term is small enough.
\end{remark} 

\begin{remark}[Extension to asynchronous Q-learning]
We would like to point out that our lower bound extends to the asynchronous setting \citep{Li2023QLMinimax} as the assumption of i.i.d. noise corresponding to a generative model is a special case of Markovian noise observed in the asynchronous setting.
\end{remark} 

%% file: algorithm.tex
\section{The \DVR \ Algorithm}
\label{sec:fed_dvr_algorithm}

Having characterized the lower bound on the communication complexity of federated Q-learning, we explore our second question of interest --- designing a federated Q-learning algorithm that achieves this lower bound while simultaneously offering an optimal order of sample complexity (up to logarithmic factors). 

We propose a new federated Q-learning algorithm, \DVR, that achieves not only a communication complexity of $\CC_{\textsf{round}} = \tilde{\cO} \left( \frac{1}{1-\gamma} \right)$ and $\CC_{\textsf{bit}} = \tilde{\cO} \left( \frac{|\cS| |\cA|}{1-\gamma} \right)$ but also  order-optimal sample complexity (up to logarithmic factors), thereby providing a tight characterization of the achievability frontier that matches with the converse result derived in the previous section.




\subsection{Algorithm description}

\begin{wrapfigure}{R}{0.6\textwidth}
\begin{minipage}{0.6\textwidth}
\vspace{-1em}
    \begin{algorithm}[H]
    \caption{\DVR}
    \label{alg:fed_dvr_q}
    \begin{algorithmic}[1]
        \STATE Input : Error bound $\varepsilon > 0$, failure probability $\delta > 0$
        \STATE $k \leftarrow 1, Q^{(0)} \leftarrow \bfzero$
        \STATE \texttt{// Set parameters as described in Section~\ref{ssub:parameter_setting}}
        \FOR{$k = 1,2,\dots, K$}
        \STATE $Q^{(k)} \leftarrow \textsc{RefineEstimate}(Q^{(k-1)}, B, I, {L}_k, D_k, J)$
        \STATE $k \leftarrow k + 1$
        \ENDFOR
        \STATE \textbf{return} $Q^{(K)}$
    \end{algorithmic}
\end{algorithm}
\end{minipage}
\end{wrapfigure}

\DVR \ proceeds in epochs. During an epoch $k \geq 1$, the agents collaboratively update $Q^{(k-1)}$, the estimate of $Q^{\star}$ obtained at the end of the previous epoch, to a new estimate $Q^{(k)}$, with the aid of the sub-routine called \textsc{RefineEstimate}. The sub-routine \textsc{RefineEstimate} is designed to ensure that the suboptimality gap, $\|Q^{(k)} - Q^{\star}\|_{\infty}$, reduces by a factor of $2$ at the end of every epoch. Thus, at the end of $K = \cO(\log(1/\varepsilon))$ epochs, \DVR \ obtains an $\varepsilon$-optimal estimate of $Q^{\star}$, which is then set to be the output of the algorithm. Please refer to Algorithm~\ref{alg:fed_dvr_q} for a pseudocode.

\subsubsection{The \textsc{RefineEstimate} sub-routine}
\label{ssub:refine_estimate_routine}

\textsc{RefineEstimate}, starting from $\overline{Q}$, an initial estimate of $Q^{\star}$, uses variance-reduced Q-learning updates \citep{Wainwright2019VarianceReduced,sidford2018near} to obtain an improved estimate of $Q^{\star}$. It is characterized by four parameters --- the initial estimate $\overline{Q}$, the number of local iterations $I$, the re-centering sample size $L$ and the batch size $B$, which can be appropriately tuned to control the quality of the returned estimate. Additionally, it also takes input two parameters $D$ and $J$ required by the compressor $\sC(\cdot; D, J)$, whose description is deferred to Section~\ref{ssub:quantization}. 

The first step in \textsc{RefineEstimate} is to collaboratively approximate $\cT\overline{Q}$ for the variance-reduced updates. To this effect, each agent $m$ builds an approximation of $\cT\overline{Q}$ as follows: 
\begin{align}
    \widetilde{\cT}^{(m)}_{L}(\overline{Q}) := \frac{1}{\lceil {L}/M \rceil} \sum_{l = 1}^{\lceil {L}/M \rceil} \cT_{Z_l^{(m)}}(\overline{Q}), \label{eqn:tilde_T_N_m_def}
\end{align}
where $\{Z_1^{(m)}, Z_2^{(m)}, \ldots, Z_{\lceil {L}/M \rceil}^{(m)}\}$ are $\lceil {L}/M \rceil$ i.i.d. samples with $Z_1^{(m)} \sim Z$.
Each agent then sends a compressed version, $\sC\left( \tildeT^{(m)}_{{L}}(\overline{Q}) - \barQ; D, J \right)$, of the difference $\tildeT^{(m)}_{{L}}(\overline{Q}) - \barQ$ to the server, which collects all the estimates from the agents and constructs the estimate
\begin{align}
    \tildeT_{L}(\overline{Q}) = \overline{Q} + \frac{1}{M} \sum_{m = 1}^M \sC\left( \tildeT^{(m)}_{{L}}(\overline{Q}) - \barQ; D, J \right) \label{eqn:tilde_T_N_def}
\end{align}
and sends it back to the agents for the variance-reduced updates.  
Equipped with the estimate $\tildeT_{L}(\overline{Q})$, \textsc{RefineEstimate} constructs a sequence $\{Q_i\}_{i = 1}^I$ using the following iterative update scheme initialized with $Q_0 = \barQ$. During the $i$-th iteration, each agent $m$ carries out the following update:
\begin{align}
    Q_{i - \frac{1}{2}}^{m} = (1 - \eta) Q_{i-1} + \eta \left[ \widehat{\cT}_i^{(m)}Q_{i-1} - \widehat{\cT}_i^{(m)}\overline{Q} + \tildeT_{L}(\overline{Q}) \right]. \label{eqn:q_l_update_rule}
\end{align}
In the above equation, $\eta \in (0,1)$ is the step size and $\widehat{\cT}_i^{(m)}Q := \frac{1}{B} \sum_{z \in \cZ_{i}^{(m)}} \cT_{z}Q$, where $\cZ_{i}^{(m)}$ is the minibatch of $B$ i.i.d. random variables drawn according to $Z$, independently at each agent $m$ for all iterations $i$. Each agent then sends a compressed version of the update, i.e., $\sC\left( Q_{i- \frac{1}{2}}^{m} - Q_{i-1}; D, J \right )$, to the server, which uses them to compute the next iterate 
\begin{align}
    Q_i = Q_{i-1} + \frac{1}{M} \sum_{m = 1}^M \sC\left( Q_{i- \frac{1}{2}}^{m} - Q_{i-1}; D, J \right ), \label{eqn:Q_l_defn}
\end{align}
and broadcast it to the agents. After $I$ such updates, the obtained iterate $Q_I$ is returned by the sub-routine. A pseudocode of \RE is given in Algorithm~\ref{alg:refine_estimate}.


\begin{algorithm}[!h]
    \caption{\RE $(\barQ, B, I, L, D, J)$}
    \label{alg:refine_estimate}
    \begin{algorithmic}[1]
        \STATE Input: Initial estimate $\barQ$, batch size $B$, number of iterations $I$, re-centering sample size $L$, quantization bound $D$, message size $J$ 
        \STATE \texttt{// Build an approximation for $\cT \barQ$ which is to be used for variance reduced updates}
        \FOR{$m = 1,2, \dots, M$}
        \STATE Draw $\lceil L/M \rceil$ i.i.d. samples from the generative model for each $(s,a)$ pair and evaluate $\tildeT_L^{(m)}(\barQ)$ according to Eqn.~\eqref{eqn:tilde_T_N_m_def}
        \STATE Send $\sC(\tildeT_L^{(m)}(\barQ) - \barQ; D, J)$ to the server 
        \STATE Receive $\frac{1}{M} \sum_{m = 1}^M \sC(\tildeT_L^{(m)}(\barQ) - \barQ; D, J)$ from the server and compute $\tildeT_L (\barQ)$ according to Eqn.~\eqref{eqn:tilde_T_N_def}
        \ENDFOR
        \STATE $Q_0 \leftarrow \barQ$
        \STATE \texttt{// Variance reduced updates with minibatching}
        \FOR{$i = 1,2,\dots, I$}
        \FOR{$m = 1,2,\dots, M$}
        \STATE Draw $B$ i.i.d. samples from the from the generative model for each $(s,a)$ pair 
        \STATE Compute $Q_{i - \frac{1}{2}}^{m}$ according to Eqn.~\eqref{eqn:q_l_update_rule}
        \STATE Send $\sC(Q_{i - \frac{1}{2}}^{m} - Q_{i-1}; D, J)$ to the server 
        \STATE Receive $\frac{1}{M} \sum_{m = 1}^M \sC(Q_i^{m} - Q_{i-1}; D, J)$ from the server and compute $Q_i$ according to Eqn.~\eqref{eqn:Q_l_defn}
        \ENDFOR
        \ENDFOR
        \STATE \textbf{return} $Q_I$
    \end{algorithmic}
\end{algorithm}

\subsubsection{The compression operator}
\label{ssub:quantization}

The compressor, $\sC(\cdot; D, J)$, used in the proposed algorithm \DVR \ is based on the popular stochastic quantization scheme. In addition to the input vector $Q$ to be quantized, the compressor or quantizer $\sC$ takes two input parameters $D$ and $J$: (i) $D$ corresponds to an upper bound on the $\ell_{\infty}$ norm of $Q$, i.e., $\|Q\|_{\infty} \leq D$; (ii) $J$ corresponds to the resolution of the compressor, i.e., number of bits used by the compressor to represent each coordinate of the output vector. 

The compressor first splits the interval $[0, D]$ into $2^J -1$ intervals of equal length where $0 = d_1 < d_2 \ldots < d_{2^J} = D$ correspond to end points of the intervals. Each coordinate of $Q$ is then separately quantized as follows. The value of the $n$-th coordinate, $\sC(Q)[n]$,  is set to be $d_{j_n - 1}$ with probability $\frac{d_{j_n} - Q[n]}{d_{j_n} - d_{j_n -1}}$ and to $d_{j_n}$ with the remaining probability,
where $j_n := \min\{j : d_j < Q[i] \leq d_{j+1}\}$. It is straightforward to note that each coordinate of $\sC(Q)$ can be represented using $J$ bits.

\subsubsection{Setting the parameters}
\label{ssub:parameter_setting}

The desired convergence of the iterates $\{Q^{(k)}\}$ is obtained by carefully choosing the parameters of the sub-routine \RE \ and the compression operator $\sC$. Given a target accuracy $\varepsilon \in (0,1]$ and $\delta \in (0,1)$, the total number of epochs is set to 
\begin{align}\label{eq:def_K}
K = \left\lceil \frac{1}{2}\log_2 \left(\frac{1}{1-\gamma} \right) \right\rceil + \left\lceil \frac{1}{2}\log_2 \left(\frac{1}{(1-\gamma)\varepsilon^2} \right) \right\rceil.
\end{align}
For all epochs $k \geq 1$, we set the number of iterations $I$, the batch size $B$, and the number of bits $J$ of the compressor $\sC$ to be 
\begin{subequations} \label{eq:def_IBJ}
\begin{align}
I & := \left\lceil \frac{2}{\eta(1-\gamma)} \right\rceil, \label{eq:def_I}\\
B & := \left\lceil \frac{2}{M}\left(\frac{12\gamma}{1-\gamma}\right)^2  \log\left(\frac{8KI|\cS||\cA|}{\delta} \right) \right\rceil,  \label{eq:def_B}\\
J & := \left\lceil \log_2\left(\frac{70}{\eta(1-\gamma)}\sqrt{\frac{4}{M} \log\Big(\frac{8KI|\cS||\cA|}{\delta} \Big)} \right)  \right\rceil    \label{eq:def_J}
\end{align}
\end{subequations}
respectively.  The re-centering sample sizes $L_k$ and bounds $D_k$ of the compressor $\sC$ are set to be the following functions of epoch index $k$ respectively:
\begin{align} \label{eq:choice_Lk}
    L_k  :=  \frac{39200}{(1-\gamma)^2} \log\left(\frac{8KI|\cS||\cA|}{\delta} \right) \cdot\begin{cases} 4^k & \text{ if } k \leq K_0 \\ 4^{k-K_0} & \text{ if } k > K_0 \end{cases}; \quad  D_k := 16 \cdot \frac{2^{-k}}{1-\gamma},
\end{align}
where $K_0 = \lceil \frac{1}{2}\log_2 (\frac{1}{1-\gamma}) \rceil$. The piecewise definition of $L_k$ is crucial to obtain the optimal dependence with respect to $\frac{1}{1-\gamma}$, similar to the two-step procedure outlined in~\citet{Wainwright2019VarianceReduced}.



\subsection{Performance guarantees}

The following theorem characterizes the sample and communication complexities of \DVR.

\begin{theorem}
    Consider any $\delta \in (0,1)$ and $\varepsilon \in (0, 1]$. The sample and communication complexities of the \normalfont{\DVR} algorithm, when run with the choice of parameters described in Section~\ref{ssub:parameter_setting} and a learning rate $\eta \in (0,1)$, satisfy the following relations for some universal constant $C_1 > 0$:
    \begin{align*}
        \SC(\DVR; \varepsilon, M, \delta) & \leq \frac{C_1}{\eta M(1-\gamma)^3\varepsilon^2}\log_2\left(\frac{1}{(1-\gamma)\varepsilon}\right)\log\left(\frac{8KI|\cS||\cA|}{\delta} \right), \\
        \CC_{\textsf{round}}(\DVR; \varepsilon, \delta) &\leq \frac{16}{\eta(1-\gamma)} \log_2\left(\frac{1}{(1-\gamma)\varepsilon}\right), \\
        \CC_{\textsf{bit}}(\DVR; \varepsilon, \delta ) &\leq \frac{32|\cS|\cA|}{\eta(1-\gamma)} \log_2\left(\frac{1}{(1-\gamma)\varepsilon}\right) \log_2\left(\frac{70}{\eta(1-\gamma)}\sqrt{\frac{4}{M} \log\left(\frac{8KI|\cS||\cA|}{\delta} \right)}\right).
    \end{align*}
    \label{thm:fed_dvr_performance}
\end{theorem}
A proof of Theorem~\ref{thm:fed_dvr_performance} can be found in Appendix~\ref{appendix:dvr_analysis}. A few implications of the theorem are in order.

\paragraph{Optimal sample-communication complexity trade-off.} As shown by the above theorem, \DVR \ offers a linear speedup in the sample complexity with respect to the number of agents while simultaneously achieving the same order of communication complexity as dictated by the lower bound derived in Theorem~\ref{thm:lower_bound}, both in terms of frequency and bit level complexity. 
Moreover, \DVR \ is the \emph{first} federated Q-learning algorithm that achieves a sample complexity with optimal dependence on all the salient parameters, i.e., $|\cS|, |\cA|$ and $\frac{1}{1-\gamma}$, in addition to linear speedup w.r.t. to the number of agents, thereby bridging the existing gap between upper and lower bounds on the sample complexity for federated Q-learning.
Thus, Theorem~\ref{thm:lower_bound} and~\ref{thm:fed_dvr_performance} together provide a characterization of optimal operating point of the sample-communication complexity trade-off in federated Q-learning.

\paragraph{Role of minibatching.} The commonly adopted approach in intermittent communication algorithm is to use a local update scheme that takes multiple small (i.e., $B = \cO(1)$), noisy updates between communication rounds, as evident from the algorithm design in~\citet{Khodadadian2022FederatedQL, Woo2023FedSynQ} and even numerous FL algorithms for stochastic optimization~\citep{Mcmahan2017FedAvg, Haddadpour2019LUPASGD, Khaled2020LocalSGDHeterogenous}. In \DVR, we replace the local update scheme of taking multiple small, noisy updates by a single, large update with smaller variance, obtained by averaging the noisy updates over a minibatch of samples. The use of updates with smaller variance in variance-reduced Q-learning yields the algorithm its name. While both the approaches result in similar sample complexity guarantees, the local update scheme requires more frequent averaging across clients to ensure that the bias of the estimate, also commonly referred to as ``client drift'', is not too large. On the other hand, the minibatching approach does not encounter the problem of bias accumulation from local updates and hence can afford more infrequent averaging, allowing \DVR \ to achieve optimal order of communication complexity. 

\paragraph{Compression.} \DVR \ is the first federated Q-learning algorithm to analyze and establish communication complexity at the bit level. While all existing studies on federated Q-learning focus only on the frequency of communication and assume transmission of real numbers with infinite bit precision, our analysis provides a more holistic view point of communication complexity  at the bit level, which is of great practical significance. While some recent other studies~\citep{Wang2023FederatedTDHeterogeneity} also considered quantization in federated RL, their objective is to understand the impact of message size on the convergence with no constraint on the frequency of communication, unlike the holistic viewpoint adopted in this work.


%% file: simulations.tex
\section{Numerical Experiments}
\label{appendix:empirical_studies}

In this section, we corroborate our theoretical results through simulations. For the simulations, we consider an MDP with $3$ states and two actions, i.e., $\cS = \{0,1,2\}$ and $\cA = \{0,1\}$. The discount parameter is set to $\gamma = 0.9$. The reward and transition kernel of the MDP is based on the hard instance constructed in Appendix~\ref{appendix:proof_lower_bound}. Specifically, the reward and transition kernel of state $0$ is given by the expression in Eqn.~\eqref{eqn:state_0_dynamics}. Similarly, the reward and transition kernel corresponding to state $1$ and $2$ are identical and given by Eqns.~\eqref{eqn:state_1_dynamics_1} and~\eqref{eqn:state_1_dynamics_2} with $p = 0.8$.

\begin{figure}[ht]
\centering
\subfloat[Sample Complexity]{\label{fig:samp_complex_comp}\centering \includegraphics[scale = 0.4]{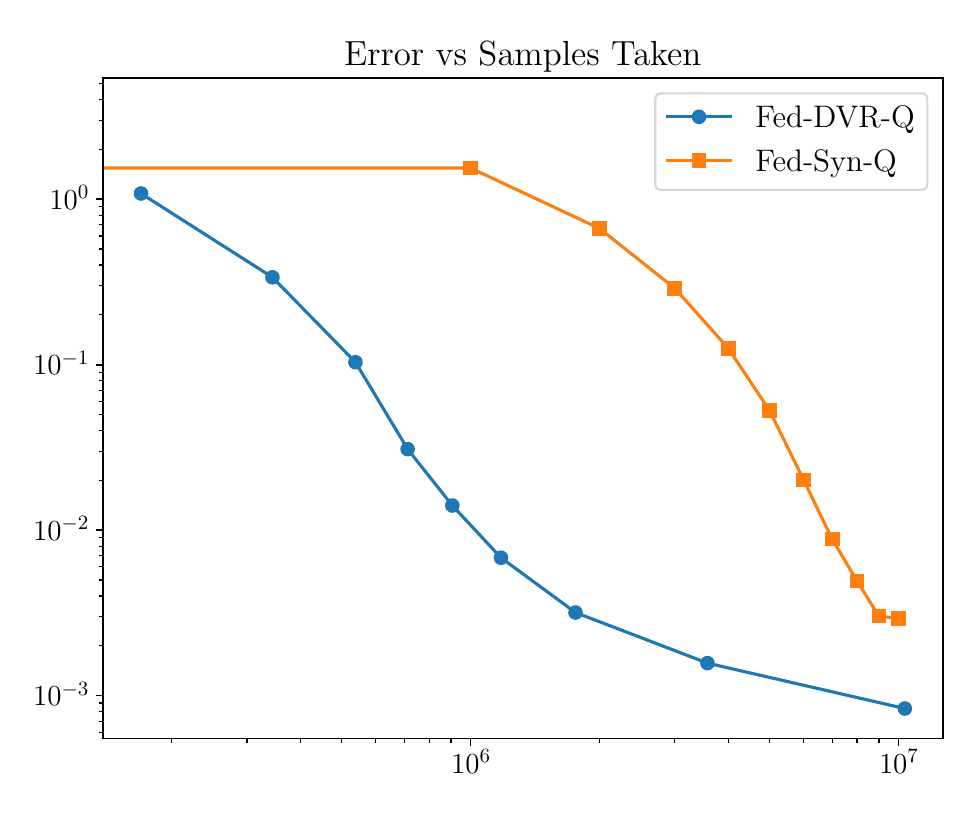}}
~
\subfloat[Communication Complexity]{\label{fig:comm_complex_comp}\centering \includegraphics[scale = 0.4]{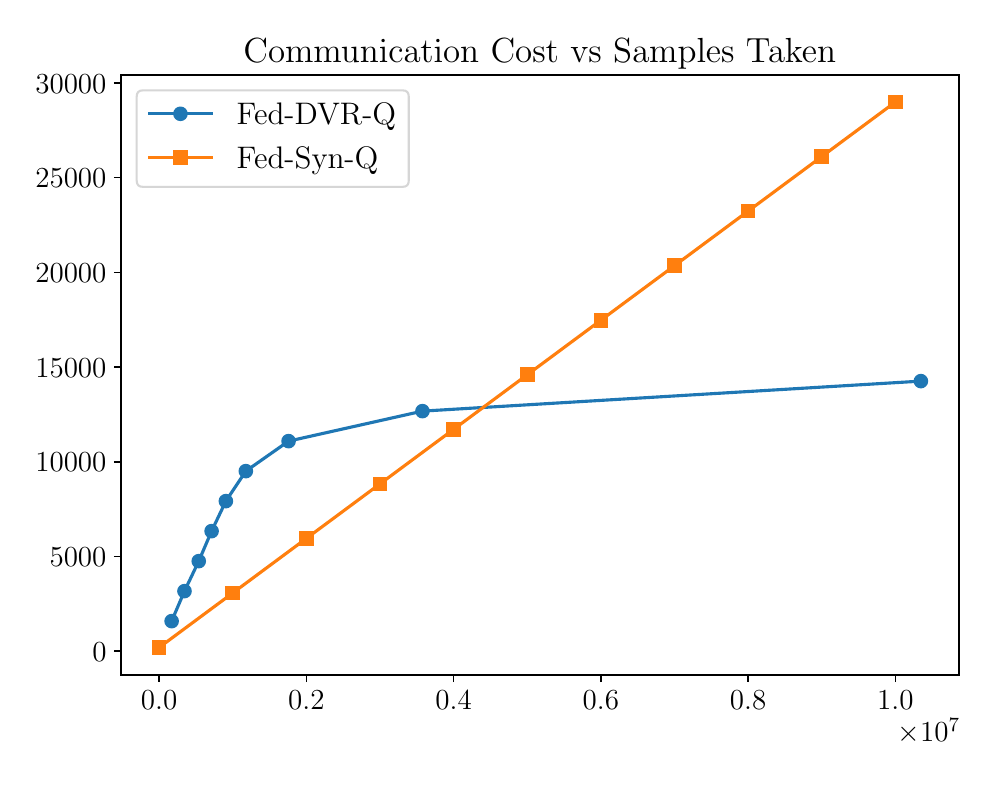}}
~
\caption{Comparison between sample and communication complexities of \DVR \ and the algorithm \textsf{Fed-SynQ} from~\citet{Woo2023FedSynQ}. }
\label{fig:comparison_bw_algos}
\end{figure}

\begin{figure}[ht]
\centering
\subfloat[Sample Complexity]{\label{fig:samp_comp_vs_n_agents}\centering \includegraphics[scale = 0.4]{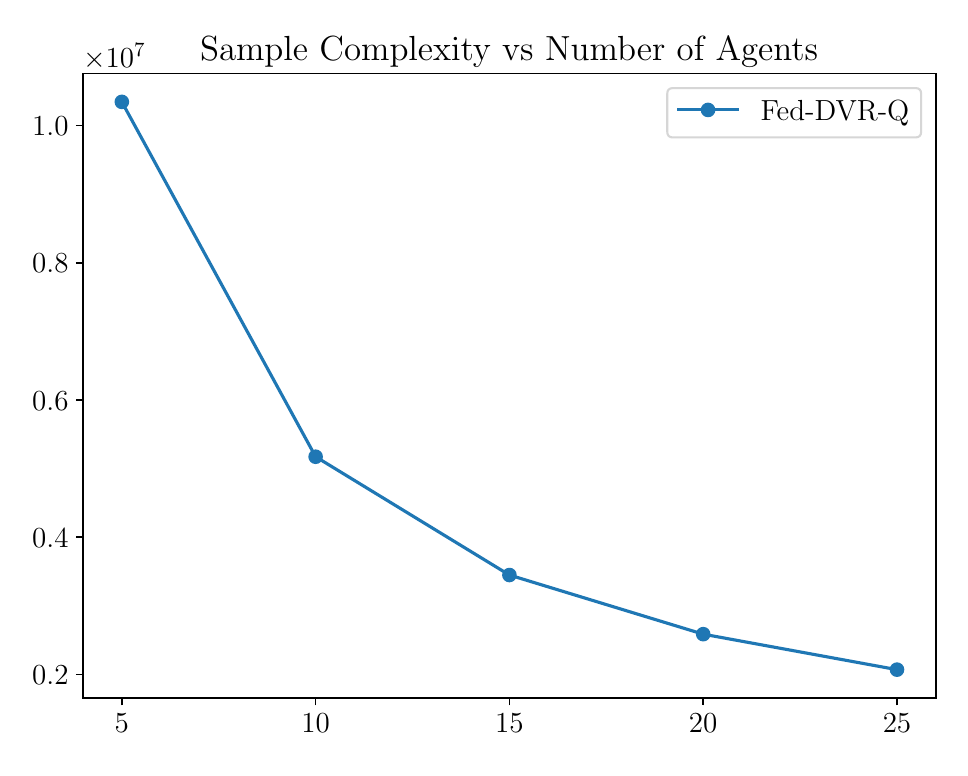}}
~
\subfloat[Communication Complexity]{\label{fig:comm_comp_vs_n_agents}\centering \includegraphics[scale = 0.4]{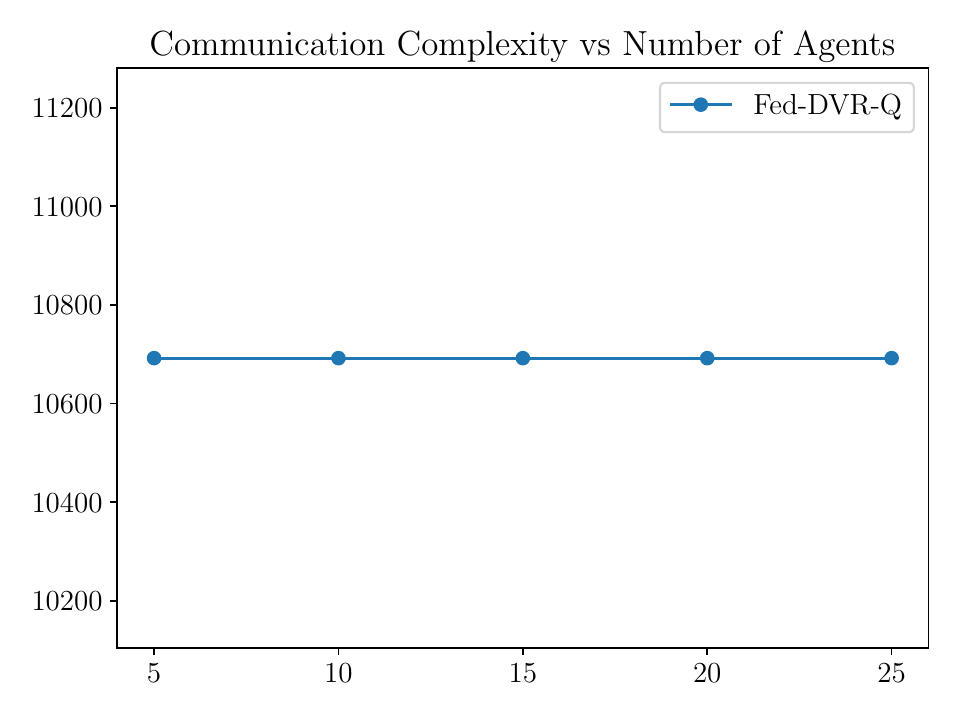}}
~
\caption{Dependence of sample and communication complexities of \DVR \ on the number of agents.}
\label{fig:comparison_agents}
\end{figure}

\begin{figure}[ht]
\centering
\includegraphics[scale = 0.4]{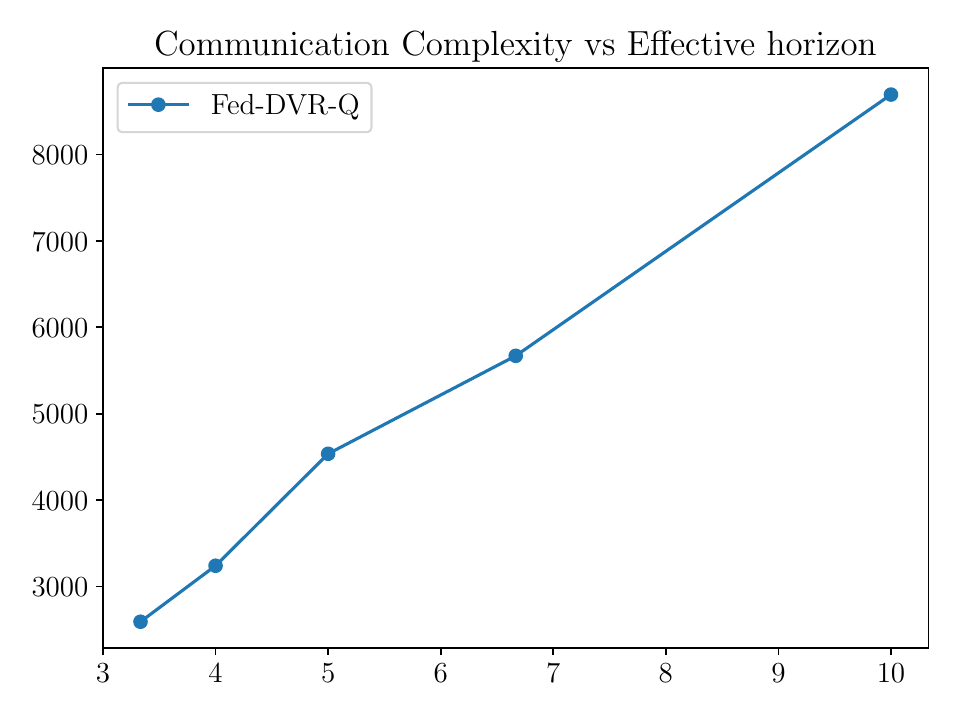}
\caption{Communication complexity of \DVR \  as a function of effective horizon, i.e., $\frac{1}{1-\gamma}$.}
\label{fig:comm_comp_vs_eff_horizon}
\end{figure}

We perform three empirical studies. In the first study, we compare the proposed algorithm \DVR \ to the \textsf{Fed-SynQ} algorithm proposed in~\citet{Woo2023FedSynQ}. We consider a Federated Q-learning setting with $5$ agents. The parameters for both the algorithms were set to the suggested values in the respective papers. Both the algorithms were run with $10^7$ samples at each agent. For the communication cost of \textsf{Fed-SynQ} we assume that each real number is expressed using $32$ bits. In Fig~\ref{fig:samp_complex_comp}, we plot the error rate of the algorithm as a function of the number of samples used. In Fig.~\ref{fig:comm_complex_comp} we plot the corresponding communication complexities. 
As evident from Fig~\ref{fig:samp_complex_comp}, \DVR \ achieves a smaller error than Fed-SynQ under the same sample budget. Similarly, as suggested by Fig.~\ref{fig:comm_complex_comp}, \DVR \ also requires much less communication (measured in terms of the number of bits transmitted) than \textsf{Fed-SynQ}, demonstrating the effectiveness of the proposed approach and corroborating our theoretical results. 

In the second study, we examine the effect of the number of agents on the sample and communication complexity of \DVR. We vary the number of agents from $5$ to $25$ in multiples of $5$ and record the sample and communication complexity to achieve an error rate of $\varepsilon = 0.03$. The sample and communication complexities as a function of number of agents are plotted in Figs.~\ref{fig:samp_comp_vs_n_agents} and~\ref{fig:comm_comp_vs_n_agents} respectively. The sample complexity decreases as $1/M$ while the communication complexity is independent of the number of agents. This corroborates the linear speedup phenomenon suggested by our theoretical results and the independence between communication complexity and the number of agents.

In the last study, we compare the communication complexity of \DVR \  as function of the discount parameter $\gamma$. We consider the same setup as in the first study and vary the values of $\gamma$ from $0.7$ to $0.9$ in steps of $0.05$. We run the algorithm to achieve an accuracy of $\varepsilon = 0.1$ with parameter choices prescribed in Sec.~\ref{ssub:parameter_setting}. We plot the communication cost of \DVR \ against the effective horizon, i.e., $\frac{1}{1-\gamma}$ in Fig.~\ref{fig:comm_comp_vs_eff_horizon}. As evident from the figure, the communication scales linearly with the effective horizon, which matches the theoretical claim in Theorem~\ref{thm:fed_dvr_performance}.

%% file: conclusion.tex
\section{Conclusion}
\label{sec:conclusion}

We presented a complete characterization of the sample-communication trade-off for federated Q-learning algorithms with intermittent communication. We showed that no federated Q-learning algorithm with intermittent communication can achieve {\em any} speedup with respect to the number of agents if its number of communication rounds are sublinear in $\frac{1}{1-\gamma}$. We also proposed a new federated Q-learning algorithm called \DVR \ that uses variance reduction along with minibatching to achieve optimal-order sample and communication complexities. In particular, we showed that \DVR \ has a per-agent sample complexity of $\tilde{\cO}\left( \frac{|\cS| |\cA|}{M (1-\gamma)^3 \varepsilon^2}\right)$, which is order-optimal in all salient problem parameters, and a communication complexity of $\tilde{\cO}\left(\frac{1}{1-\gamma}\right)$ rounds and $\tilde{\cO}\left(\frac{|\cS| |\cA|}{1-\gamma}\right)$ bits. 

The results in this work raise several interesting questions that are worth exploring. While we focus on the tabular setting in this work, it is of great interest to investigate to the trade-off in other settings where we use function approximation to model the $\starQ$ and $V^{\star}$ functions. Moreover, it is interesting to explore the trade-off in the finite-horizon setting, where there is no discount factor. Furthermore, it is also worthwhile to explore if the communication complexity can be further reduced by going beyond the class of intermittent communication algorithms.



%% file: appendix.tex
\section{Proof of Theorem~\ref{thm:lower_bound}}
\label{appendix:proof_lower_bound}

In this section, we prove the main result of the paper, the lower bound on the communication complexity of federated Q-learning algorithms. At a high level, the proof consists of the following three steps.

\paragraph{Introducing the ``hard'' MDP instance.} The proof builds upon analyzing the behavior of a generic algorithm $\sA$ outlined in Algorithm~\ref{alg:general_fed_alg_with_inter_comm} over a particular instance of MDP. The particular choice of MDP is inspired by, and borrowed from, other lower bound proofs in the single-agent setting~\citep{Li2023QLMinimax} and helps highlight core issues that lie at the heart of the sample-communication complexity trade-off. Following \citet{Li2023QLMinimax}, the construction is  first over a small state-action space that allows us to focus on a simpler problem before generalizing it to larger state-action spaces.

\paragraph{Establishing the performance of intermittent communication algorithms.} In the second step, we analyze the error of the iterates generated by an intermittent communication algorithm $\sA$. The analysis is inspired by the single-agent analysis in \cite{Li2023QLMinimax}, which highlights the underlying bias-variance trade-off. Through careful analysis of the algorithm dynamics in the federated setting,
we uncover the impact of communication on the bias-variance trade-off and the resulting error of the iterates to obtain the lower bound on the communication complexity.

\paragraph{Generalization to larger MDPs.} As the last step, we generalize our construction of the ``hard'' instance to more general state-action space and extend our insights to obtain the statement of the theorem.

\subsection{Introducing the ``hard'' instance}

We first introduce an MDP instance $\cM_{h}$ that we will use throughout the proof to establish the result.  Note that this MDP is identical to the one considered in~\cite{Li2023QLMinimax} to establish the lower bounds on the performance of single-agent Q-learning algorithm.  It consists of four states $\cS = \{0,1,2,3\}$. Let $\cA_s$ denote the action set associated with the state $s$. The probability transition kernel and the reward function of $\cM_{h}$ is given as follows:
\begin{subequations}
\label{eqn:hard_MDP}
\begin{align}
    \cA_0 = \{1\}& \quad \quad P(0|0,1) = 1& \quad &\quad r(0,1) = 0, \label{eqn:state_0_dynamics}\\
    \cA_1 = \{1,2\}& \quad\quad P(1|1,1) = p& \quad P(0|1,1) = 1- p& \quad r(1,1) = 1, \label{eqn:state_1_dynamics_1}\\
                   & \quad\quad P(1|1,2) = p& \quad P(0|1,2) = 1- p& \quad r(1,2) = 1, \label{eqn:state_1_dynamics_2}\\
    \cA_2 = \{1\}& \quad\quad P(2|2,1) = p& \quad P(0|2,1) = 1- p& \quad r(2,1) = 1, \\
    \cA_3 = \{1\}& \quad\quad P(3|3,1) = 1&  & \quad r(3,1) = 1,
\end{align}
\end{subequations}
where the parameter $p = \dfrac{4 \gamma - 1}{3\gamma}$. We have the following results about the optimal $Q$ and $V$ functions of this hard MDP instance.
\begin{lemma}[{\citep[Lemma 3]{Li2023QLMinimax}}]
    Consider the MDP $\cM_{h}$ constructed in Eqn.~\eqref{eqn:hard_MDP}. We have,
    \begin{align*}
        V^{\star}(0) & = Q^{\star}(0,1) = 0 \\
        V^{\star}(1) & = Q^{\star}(1,1) = Q^{\star}(1,2) = V^{\star}(2) = Q^{\star}(2,1) = \frac{1}{1 - \gamma p} = \frac{3}{4(1-\gamma)} \\
        V^{\star}(3) & = Q^{\star}(3,1) = \frac{1}{1 - \gamma}.
     \end{align*}
\end{lemma}

Throughout the next section of the proof, we focus on this MDP with four states and two actions. In Appendix~\ref{sec:generalization_larger}, we generalize the proof to larger state-action spaces.

\subsection{Notation and preliminary results}

For convenience, we first define some notation that  will be used throughout the proof. 

\paragraph{Useful relations of the learning rates.} We consider two kinds of step size sequences that are commonly used in Q-learning --- the constant step size schedule, i.e., $\eta_t = \eta$ for all $t \in \{1,2,\dots, T\}$ and the rescaled linear step size schedule, i.e., $\eta_t = \frac{1}{1 + c_{\eta}(1-\gamma)t}$, where $c_\eta > 0$ is a universal constant that is independent of the problem parameters. 

We define the following quantities:
\begin{align}
    \eta_k^{(t)} & = \eta_k \prod_{i = k+1}^t (1- \eta_i(1- \gamma p)) ~~~~~~~~~ \text{ for all } 0 \leq k \leq t, \label{eqn:eta_k_t_defn} 
\end{align}
where we take $\eta_0 = 1$ and use the convention throughout the proof that if a product operation does not have a valid index, we take the value of that product to be $1$. For any integer $0 \leq \tau < t$, we have the following relation, which will be proved at the end of this subsection for completeness:
\begin{align}
    \prod_{k = \tau + 1}^t (1- \eta_k(1-\gamma p)) + (1- \gamma p) \sum_{k = \tau + 1}^t \eta_k^{(t)} = 1. \label{eqn:eta_k_t_recursion}
\end{align}

Similarly, we also define,
\begin{align}
    \widetilde{\eta}_k^{(t)} & = \eta_k \prod_{i = k+1}^t (1- \eta_i) ~~~~~~~~~ \text{ for all } 0 \leq k \leq t, \label{eqn:tilde_eta_k_t_defn} 
\end{align}
which satisfies the relation 
\begin{align}
    \prod_{k = \tau + 1}^t (1- \eta_k) + \sum_{k = \tau + 1}^t \widetilde{\eta}_k^{(t)} = 1. \label{eqn:tilde_eta_k_t_recursion}
\end{align}
for any integer $0 \leq \tau < t$. The claim follows immediately by plugging $p = 0$ in \eqref{eqn:eta_k_t_recursion}. Note that for constant step size, the sequence $\widetilde{\eta}_k^{(t)}$ is clearly increasing. For the rescaled linear step size, we have, 
\begin{align}
    \frac{\widetilde{\eta}_{k-1}^{(t)}}{\widetilde{\eta}_{k}^{(t)}} = \frac{\eta_k}{\eta_{k-1}(1-\eta_k)} = 1- \frac{(1 - c_{\eta}(1-\gamma))\eta_k}{1 - c_{\eta}(1 - \gamma)\eta_k} \leq 1 \label{eqn:tilde_eta_k_t_increasing}
\end{align}
whenever $c_{\eta} \leq \frac{1}{1 - \gamma}$. Thus, $\widetilde{\eta}_k^{(t)}$ is an increasing sequence as long as $c_{\eta} \leq \frac{1}{1 - \gamma}$. Similarly, $\eta_k^{(t)}$ is also clearly increasing for the constant step size schedule. For the rescaled linear step size schedule, we have, 
\begin{align*}
    \frac{\eta_{k-1}^{(t)}}{\eta_{k}^{(t)}} = \frac{\eta_k}{\eta_{k-1}(1-\eta_k(1- \gamma p))} \leq \frac{\eta_k}{\eta_{k-1}(1-\eta_k)} \leq 1, 
\end{align*}
whenever $c_{\eta} \leq \frac{1}{1-\gamma}$. The last bound follows from Eqn.~\eqref{eqn:tilde_eta_k_t_increasing}.


\paragraph{Proof of \eqref{eqn:eta_k_t_recursion}.}
We can show the claim using backward induction. For the base case, note that,
\begin{align*}
    (1-\gamma p)\eta_t^{(t)} + (1-\gamma p)\eta_{t-1}^{(t)} & = (1- \gamma p)\eta_t + (1-\gamma p)\eta_{t-1}(1 - (1- \gamma p)\eta_t) \\
    & = 1 - (1- \eta_t(1-\gamma p))(1-\eta_{t-1}(1 -\gamma p)) = 1 - \prod_{k = t-1}^t (1- \eta_k(1-\gamma p)),
\end{align*}
as required. Assume \eqref{eqn:eta_k_t_recursion} is true for some $\tau$. We have,
\begin{align*}
    (1- \gamma p) \sum_{k = \tau }^t \eta_k^{(t)} & = (1- \gamma p)\eta_{\tau}^t + (1- \gamma p) \sum_{k = \tau + 1}^t \eta_k^{(t)} \\
    & = (1-\gamma p)\eta_{\tau}\prod_{k = \tau + 1}^t (1- \eta_k(1-\gamma p))  + 1 - \prod_{k = \tau + 1}^t (1- \eta_k(1-\gamma p)) \\
    & = 1 - \prod_{k = \tau}^t (1- \eta_k(1-\gamma p)),
\end{align*}
thus completing the induction step.

\paragraph{Sample transition matrix.} Recall $Z \in \cS^{|\cS| |\cA|}$ is a  random vector whose $(s,a)$-th coordinate is drawn from the distribution $P( \cdot | s,a)$.
We use $\hatP_t^m$ to denote the sample transition at time $t$ and agent $m$ obtained by averaging $B$ i.i.d. samples from the generative model. 
Specifically let $\{Z_{t,b}^m\}_{b = 1}^{B}$ denote a collection of $B$ i.i.d. copies of $Z$ collected at time $t$ at agent $m$. Then, for all $s, a, s'$,  
\begin{align}
    \hatP_t^m(s'| s,a) = \frac{1}{B} \sum_{b = 1}^{B} P_{t, b}^m(s'|s,a),
\end{align}
where $ P_{t, b}^m(s'|s,a) = \1\{Z_{t,b}^m(s,a) = s'\}$ for $s' \in \cS$.

\paragraph{Preliminary relations of the iterates.}
We state some preliminary relations regarding the evolution of the Q-function and the value function across different agents that will be helpful for the analysis later. 

We begin with the state $0$, where we have $Q_t^m(0,1) = V_t^m(0) = 0$ for all agents $m \in [M]$ and $t \in [T]$. This follows almost immediately from the fact that state $0$ is an absorbing state with zero reward. Note that $Q_0^m(0,1) = V_0^m(0) = 0$ holds for all clients $m \in [M]$. Assuming that $Q_{t-1}^m(0,1) = V_{t-1}^m(0) = 0$ for all clients for some time instant $t - 1$, by induction, we have,
\begin{align*}
    Q_{t-1/2}^m(0,1) = (1- \eta_t) Q_{t-1}^m(0,1) + \eta_t (\gamma V_{t-1}^m(0)) = 0.
\end{align*}
Consequently, $Q_t^m(0,1) = 0$ and $V_{t}^m(0) = 0$, for all agents $m$, irrespective of whether there is averaging. \\

For state $3$, the iterates satisfy the following relation:
\begin{align*}
    Q_{t - 1/2}^m(3,1) & = (1 - \eta_t)Q_{t-1}^m(3,1) + \eta_t(1+ \gamma V_{t-1}^m(3)) \\
    & = (1 - \eta_t)Q_{t-1}^m(3,1) + \eta_t(1+ \gamma Q_{t-1}^m(3,1)) \\
    & = (1 - \eta_t(1-\gamma))Q_{t-1}^m(3,1) + \eta_t,
\end{align*}
where the second step follows by noting $V_t^m(3) = Q_{t}^m(3,1)$. Once again, one can note that averaging step does not affect the update rule implying that the following holds for all $m \in [M]$ and $t \in [T]$:
\begin{align}
    V_t^m(3) =  Q_{t}^m(3,1) = \sum_{k = 1}^t \eta_k \left( \prod_{i = k + 1}^t (1 - \eta_i(1-\gamma)) \right) = \frac{1}{1-\gamma}\left[ 1 - \prod_{i = 1}^t (1 - \eta_i(1-\gamma))\right], \label{eqn:state_3_recursion}
\end{align}
where the last step follows from Eqn.~\eqref{eqn:eta_k_t_recursion} with $p = 1$. \\

Similarly, for state $1$ and $2$, we have,
\begin{align}
    Q_{t-1/2}^m(1,1) & = (1 - \eta_t)Q_{t-1}^m(1,1) + \eta_t (1 + \gamma \hatP_t^m(1|1,1) V_{t-1}^m(1)), \label{eqn:state_1_1_update} \\
    Q_{t-1/2}^m(1,2) & = (1 - \eta_t)Q_{t-1}^m(1,2) + \eta_t (1 + \gamma \hatP_t^m(1|1,2) V_{t-1}^m(1)), \label{eqn:state_1_2_update} \\
    Q_{t-1/2}^m(2,1) & = (1 - \eta_t)Q_{t-1}^m(2,1) + \eta_t (1 + \gamma \hatP_t^m(2|2,1) V_{t-1}^m(2)). \label{eqn:state_2_update}
\end{align}
Since the averaging makes a difference in the update rule, we further analyze the update as required in later proofs.

\subsection{Main analysis}

We first focus  on establishing a bound on the number of communication rounds, i.e., $\CC_{\textsf{round}}(\sA)$ (where we drop the dependency with other parameters for notational simplicity), and then use this lower bound to establish the bound on the bit level communication complexity $\CC_{\textsf{bit}}(\sA)$. 

To establish the lower bound on $\CC_{\textsf{round}}(\sA)$ for any intermittent communication algorithm $\sA$, we analyze the convergence behavior of $\sA$ on the MDP $\cM_h$. 
We assume that the averaging step in line $6$ of Algorithm~\ref{alg:general_fed_alg_with_inter_comm} is carried out exactly. Since the use of compression only makes the problem harder, it is sufficient for us to consider the case where there is no loss of information in the averaging step for establishing a lower bound. 
Lastly, throughout the proof, without loss of generality we assume that  
\begin{align}
    \log N \leq \frac{1}{1 - \gamma}, \label{eqn:log_n_bound}
\end{align}
otherwise, the lower bound in Theorem~\ref{thm:lower_bound} reduces to the trivial lower bound.

We divide the proof into following three parts based on the choice of learning rates and batch sizes:
\begin{enumerate}
    \item Small learning rates: For constant learning rates, $0 \leq \eta < \frac{1}{(1- \gamma)T}$ and for rescaled linear learning rates, the constant $c_{\eta}$ satisfies $c_{\eta} \geq \log T$.
    \item Large learning rates with small $\eta_T/(BM)$: For constant learning rates, $\eta \geq \frac{1}{(1- \gamma)T}$ and for rescaled linear learning rates, the constant $c_{\eta}$ satisfies $0 \leq c_{\eta} \leq \log T \leq \frac{1}{1-\gamma}$ (c.f.~\eqref{eqn:log_n_bound}). Additionally, the ratio $\frac{\eta_T}{BM}$ satisfies $\frac{\eta_T}{BM} \leq \frac{1-\gamma}{100}$.
    \item  Large learning rates with large $\eta_T/(BM)$: We have the same condition on the learning rates as above. However, in this case the ratio $\frac{\eta_T}{BM}$ satisfies $\frac{\eta_T}{BM} > \frac{1-\gamma}{100}$.
\end{enumerate}
We consider each of the cases separately in the following three subsections.




\subsubsection{Small learning rates}

In this subsection, we prove the lower bound for small learning rates, which follow from similar arguments in \cite{Li2023QLMinimax}. 
 
For this case, we focus on the dynamics of state $2$. We claim that the same relation established in \cite{Li2023QLMinimax} continues to hold, which will be established momentarily: 
\begin{align} \label{eq:mean_V_state2}
    \E[V_T^m(2)] = \left(\frac{1}{M} \sum_{j = 1}^M  \E[V_{T}^j(2)]\right) = \sum_{k = 1}^T \eta_k^{(t)} = \frac{1 - \eta_0^{(T)}}{1 - \gamma p}.
\end{align}
Consequently, for all $m \in [M]$, we have
\begin{align}
    V^{\star}(2) - \E[V_T^m(2)] = \frac{\eta_0^{(T)}}{1- \gamma p}. \label{eqn:v_minus_exp_v_state_2}
\end{align}
To obtain lower bound on $V^{\star}(2) - \E[V_T^m(2)]$, we need to obtain a lower bound on $\eta_0^{(T)}$, which from \cite[Eqn.~(120)]{Li2023QLMinimax} we have 
\begin{align*}
    \log(\eta_0^{(T)}) \geq -1.5\sum_{t = 1}^T \eta(1- \gamma p) \geq -2\sum_{t = 1}^T \frac{1}{t \log T} \geq -2 \qquad \Longrightarrow \qquad \eta_0^{(T)} \geq e^{-2}
\end{align*}
when $T \geq 16$ for both choices of learning rates. On plugging this bound in \eqref{eqn:v_minus_exp_v_state_2}, we obtain,
\begin{align}
     \E[\| Q_T^m - Q^{\star}\|_{\infty}] \geq \E[ |Q^{\star}(2) - Q_T^m(2)| ] \geq  V^{\star}(2) - \E[V_T^m(2)] \geq \frac{3}{4e^2(1- \gamma)\sqrt{N}} 
     \label{eqn:q_err_small_lr_final_bound}
\end{align}
holds for all $m \in [M]$, $N \geq 1$ and $M \geq 2$. Thus, it can be noted that the error rate  $\ER(\sA; N, M)$   is bounded away from a constant value irrespective of the number of agents and the number of communication rounds. Thus, even with $\CC_{\textsf{round}} = \Omega(T)$, we will not observe any collaborative gain if the step size is too small.


\paragraph{Proof of \eqref{eq:mean_V_state2}.} Recall that from \eqref{eqn:state_2_update}, we have,
\begin{align*}
    Q_{t-1/2}^m(2,1) & = (1 - \eta_t)V_{t-1}^m(2) + \eta_t (1 + \gamma \hatP_t^m(2|2,1) V_{t-1}^m(2)).
\end{align*}
Here, $Q_{t-1}^m(2,1) = V_{t-1}^m(2) $ as the second state has only a single action. 
\begin{itemize}
\item When $t$ is not an averaging instant, we have,
\begin{align}
    V_t^m(2) = Q_{t}^m(2,1) = (1 - \eta_t)V_{t-1}^m(2) + \eta_t (1 + \gamma \hatP_t^m(2|2,1) V_{t-1}^m(2)). \label{eqn:v_t_state_2_client_update_non_averaged}
\end{align} 
On taking expectation on both sides of the equation, we obtain,
\begin{align}
    \E[V_t^m(2)] & = (1 - \eta_t)\E[V_{t-1}^m(2)] + \eta_t (1 + \gamma \E[ \hatP_t^m(2|2,1) V_{t-1}^m(2)]) \nonumber \\
    & = (1 - \eta_t)\E[V_{t-1}^m(2)] + \eta_t \left(1 + \gamma \E[ \hatP_t^m(2|2,1) ] \E[V_{t-1}^m(2)]\right) \nonumber \\
    & = (1 - \eta_t)\E[V_{t-1}^m(2)] + \eta_t \left(1 + \gamma p \E[V_{t-1}^m(2)]\right) \nonumber \\
    & = (1 - \eta_t(1 -\gamma p))\E[V_{t-1}^m(2)] + \eta_t. \label{eqn:state_2_value_update_case_1}
\end{align} 
In the second step, we used the fact that $\hatP_t^m(2|2,1)$ is independent of $V_{t-1}^m(2)$. 

\item Similarly, if $t$ is an averaging instant, we have,
\begin{align}
    V_t^m(2) & = Q_{t}^m(2,1) = \frac{1}{M} \sum_{j = 1}^M Q_{t-1/2}^j(2,1) \nonumber \\
    & = (1 - \eta_t)\frac{1}{M} \sum_{j = 1}^M  V_{t-1}^j(2) + \frac{1}{M} \sum_{j = 1}^M \eta_t (1 + \gamma \hatP_t^j(2|2,1) V_{t-1}^j(2)). \label{eqn:v_t_state_2_client_update_averaged}
\end{align}
Once again, upon taking expectation we obtain,
\begin{align}
    \E[V_t^m(2)] & = (1 - \eta_t)\frac{1}{M} \sum_{j = 1}^M  \E[V_{t-1}^j(2)] + \frac{1}{M} \sum_{j = 1}^M \eta_t (1 + \gamma \E[\hatP_t^j(2|2,1) V_{t-1}^j(2)]) \nonumber \\
    & = (1 - \eta_t)\frac{1}{M} \sum_{j = 1}^M  \E[V_{t-1}^j(2)] + \frac{1}{M} \sum_{j = 1}^M \eta_t (1 + \gamma p \E[V_{t-1}^j(2)])\nonumber \\
    & = (1 - \eta_t(1-\gamma p)) \left(\frac{1}{M} \sum_{j = 1}^M  \E[V_{t-1}^j(2)]\right) + \eta_t. \label{eqn:state_2_value_update_case_2}
\end{align}
\end{itemize}
Eqns.~\eqref{eqn:state_2_value_update_case_1} and~\eqref{eqn:state_2_value_update_case_2} together imply that for all $t \in [T]$, 
\begin{align}
    \left(\frac{1}{M} \sum_{m = 1}^M  \E[V_{t}^m(2)]\right) = (1 - \eta_t(1-\gamma p)) \left(\frac{1}{M} \sum_{m = 1}^M  \E[V_{t-1}^m(2)]\right) + \eta_t. \label{eqn:avg_v_state_2_recursion}
\end{align}
On unrolling the above recursion with $V_0^m = 0$ for all $m \in [M]$, we obtain the desired claim \eqref{eq:mean_V_state2}.


\input{large_learning_rates}

\subsection{Generalizing to larger state action spaces}
\label{sec:generalization_larger}
We now elaborate on how we can extend the result to general state-action spaces along with the obtaining the lower bound on the bit level communication complexity. For the general case, we instead consider the following MDP. For the first four states $\{0,1,2, 3\}$, the probability transition kernel and reward function are given as follows.
\begin{subequations}
\label{eqn:hard_MDP_general}
\begin{align}
    \cA_0 = \{1\}& \quad \quad P(0|0,1) = 1& \quad &\quad r(0,1) = 0, \\
    \cA_1 = \{1,2, \dots, |\cA|\}& \quad\quad P(1|1,a) = p& \quad P(0|1,a) = 1- p& \quad r(1,a) = 1, \forall \ a \in \cA \\
    \cA_2 = \{1\}& \quad\quad P(2|2,1) = p& \quad P(0|2,1) = 1- p& \quad r(2,1) = 1, \\
    \cA_3 = \{1\}& \quad\quad P(3|3,1) = 1&  & \quad r(3,1) = 1,
\end{align}
\end{subequations}
where the parameter $p = \dfrac{4 \gamma - 1}{3\gamma}$. The overall MDP is obtained by creating $|\cS|/4$ copies of the above MDP for all sets of the form $\{4r, 4r + 1, 4r + 2, 4r + 3\}$ for $r \leq |\cS|/4 - 1$. It is straightforward to note that the lower bound on the number of communication rounds immediately transfers to the general case as well. Moreover, note that the bound on $\CC_{\textsf{round}}$ implies the bound $\CC_{\textsf{bit}} = \Omega\left(\frac{1}{(1-\gamma)\log^2 N}\right)$ as every communication entails sending $\Omega(1)$ bits.

To obtain the general lower bound on bit level communication complexity, note that we can carry out the analysis in the previous section for all $|\cA|/2$ pairs of actions in state $1$ corresponding to the set of states $\{0,1,2,3\}$. Moreover, the algorithm $\sA$, needs to ensure that the error is low across all the $|\cA|/2$ pairs. Since the errors are independent across all these pairs, each of them require $\Omega\left(\frac{1}{(1-\gamma)\log^2 N}\right)$ bits of information to be transmitted during the learning horizon leading to a lower bound of $\Omega\left(\frac{|\cA|}{(1-\gamma)\log^2 N}\right)$. Note that since we require a low $\ell_{\infty}$ error, $\sA$ needs to ensure that the error is low across all the pairs, resulting in a communication cost linearly growing with $|\cA|$. Upon repeating the argument across all $|\cS|/4$ copies of the MDP, we arrive at the lower bound of $\CC_{\textsf{bit}} = \Omega\left(\frac{|\cS| |\cA|}{(1-\gamma)\log^2 N}\right)$.

\input{auxiliary_lower_bound}

%% file: large_learning_rates.tex
\subsubsection{Large learning rates with small $\frac{\eta_T}{BM}$}

In this subsection, we prove the lower bound for case of large learning rates when the ratio $\frac{\eta_T}{BM}$ is small.
For the analysis in this part, we focus on the dynamics of state $1$. Unless otherwise specified, throughout the section we implicitly assume that the state is $1$. 

We further define a key parameter that will play a key role in the analysis:
\begin{align}
    \tau := \min \{ k \in \N : \forall \ t \geq k, \eta_t \leq \eta_{k} \leq 3 \eta_{t} \}. \label{eqn:tau_def}
\end{align}
It can be noted that for constant step size sequence $\tau = 1$ and for rescaled linear stepsize $\tau = T/3$.

\paragraph{Step 1: introducing an auxiliary sequence.} We define an auxiliary sequence $\hatQ^m_t(a)$ for $a \in \{1,2\}$ and all $t = 1,2,\dots, T$ to aid our analysis, where we drop the dependency with state $s=1$ for simplicity. The evolution of the sequence $\hatQ_t^m$ is defined in Algorithm~\ref{alg:hat_Q}, where $\hatV_t^m = \max_{a \in \{1,2\}} \hatQ_t^m(a)$. In other words, the iterates $\{\hatQ_t^m\}$ evolve exactly as the iterates of Algorithm~\ref{alg:general_fed_alg_with_inter_comm} except for the fact that sequence $\{\hatQ_t^m\}$ is initialized at the optimal $Q$-function of the MDP. We would like to point out that we assume that the underlying stochasticity is also identical in the sense that the evolution of both $Q_{t}^m$ and $\hatQ_t^m$ is governed by the same $\hatP_t^m$ matrices. The following lemma controls the error between the iterates $Q_t^m$ and $\hatQ_t^m$, allowing us to focus only on $\hatQ_t^m$. 
 
\begin{algorithm}[!h]
    \caption{Evolution of $\hatQ$}
    \label{alg:hat_Q}
    \begin{algorithmic}[1]
        \STATE Input : $T, R, \{\eta_t\}_{t = 1}^T, \cC = \{t_r\}_{r = 1}^R, B$
        \STATE Set $\hatQ_0^m(a) \leftarrow Q^{\star}(1, a)$ for $a \in \{1,2\}$ and all agents $m$ \qquad \qquad \texttt{// Different initialization}
        \FOR{$t = 1,2,\dots, T$}
        \FOR{$m = 1,2,\dots, M$}
        \STATE Compute $\hatQ_{t-\frac{1}{2}}^m$ according to Eqn.~\eqref{eqn:generic_algo_update}
        \STATE Compute $\hatQ_{t}^m$ according to Eqn.~\eqref{eqn:generic_algo_averaging}
        \ENDFOR
        \ENDFOR
    \end{algorithmic}
\end{algorithm}


\begin{lemma}
    The following relation holds for all agents $m \in [M]$, all $t \in[T]$ and $a \in \{1,2\}$:
    \begin{align*}
        Q_{t}^m(1, a) - \hatQ_t^m(a) \geq -\frac{1}{1 - \gamma} \prod_{i = 1}^t (1- \eta_i(1-\gamma)).
    \end{align*}
    \label{lemma:Q_hatQ_difference}
\end{lemma}

By Lemma~\ref{lemma:Q_hatQ_difference}, bounding the error of the sequence $\hatQ_t^m$ allows us to obtain a bound on the error of $Q_t^m$. To that effect,
we define the following terms for any $t \leq T$ and all $m \in [M]$: 
\begin{align*}
    \Delta_t^m(a) & := \hatQ_t^m(a) - Q^{\star}(1,a); \quad a = 1,2; \\
    \Delta_{t, \max}^m & = \max_{a \in \{1,2\}} \Delta_t^m(a).
\end{align*}
In addition, we use $\overline{\Delta}_t = \frac{1}{M} \sum_{m=1}^M \Delta_t^m$ to denote the error of the averaged iterate\footnote{We use this different notation in appendix as opposed to the half-time indices used in the main text to improve readability of the proof.}, and similarly,
\begin{align}
    \barDelta_{t, \max} := \max_{a \in \{1,2\}} \barDelta_t(a). \label{eqn:bar_delta_max_defn}
\end{align}

We first derive a basic recursion regarding $\Delta_t^m(a)$. From the iterative update rule in Algorithm~\ref{alg:hat_Q}, we have,
\begin{align*}
    \Delta_t^m(a) & = (1- \eta_t)\Delta_{t-1}^m(a) + \eta_t(1 +\gamma \widehat{P}_t^m(1|1, a)\hatV_{t-1}^m - Q^{\star}(1,a)) \\
    & = (1- \eta_t)\Delta_{t-1}^m(a) + \eta_t\gamma( \hatP_t^m(1|1, a)\hatV_{t-1}^m - p V^{\star}(1)) \\
    & = (1- \eta_t)\Delta_{t-1}^m(a) + \eta_t\gamma( p(\hatV_{t-1}^m - V^{\star}(1)) + (\hatP_t^m(1|1, a) - p)\hatV_{t-1}) \\
    & = (1- \eta_t)\Delta_{t-1}^m(a) + \eta_t\gamma( p\Delta_{t-1, \max}^m + (\hatP_t^m(1|1, a) - p)\hatV_{t-1}^m).
\end{align*}
Here in the last line, we used the following relation:
\begin{align*}
    \Delta_{t, \max}^m & = \max_{a \in \{1,2\}} (\hatQ_t^m(a) - Q^{\star}(1,a)) = \max_{a \in \{1,2\}} \hatQ_t^m(a) - V^{\star}(1) = \hatV_{t-1}^m - V^{\star}(1),
\end{align*}
as $Q^{\star}(1,1) = Q^{\star}(1,2) = V^{\star}(1)$.

Recursively unrolling the above expression, and using the expression \eqref{eqn:tilde_eta_k_t_defn}, we obtain the following relation: for any $t' < t$ when there is no averaging during the interval $(t',t)$
\begin{align}
    \Delta_t^m(a) & = \left(\prod_{k = t' +1}^t (1- \eta_k) \right) \Delta_{t'}^m(a)  +  \sum_{k = t' + 1}^t \widetilde{\eta}_k^{(t)}  \gamma( p\Delta_{k-1, \max}^m + (\hatP_k^m(1|1, a) - p)\hatV_{k-1}^m) . \label{eqn:delta_t_recursive_eqn_orig}
\end{align}
For any $t', t$ with $t' < t$, we define the notation
\begin{align}
    \varphi_{t', t} & := \prod_{k = t' + 1}^t (1- \eta_k) , \label{eqn:varphi_defn} \\
    \xi_{t', t}^m(a) & := \sum_{k = t' + 1}^t \widetilde{\eta}_k^{(t)} \gamma (\hatP_k^m(1|1, a) - p)\hatV_{k-1}^m, \quad a = 1,2; \label{eqn:xi_def} \\
    \xi_{t', t, \max}^m & := \max_{a \in \{1,2\}} \xi_{t', t}^{m}(a). \label{eqn:xi_max_def}
\end{align}
Note that by definition, $\E[\xi_{t',t}^m(a)] = 0$ for $a \in \{1,2\}$ and all $m$, $t'$ and $t$. Plugging them into the previous expression leads to the simplified expression
\begin{align*}
    \Delta_t^m(a) & =  \varphi_{t', t} \Delta_{t'}^m(a) +  \left[\sum_{k = t' + 1}^t \widetilde{\eta}_k^{(t)} \gamma p\Delta_{k-1, \max}^m \right] + \xi_{t',t}^m(a).
\end{align*}
We specifically choose $t'$ and $t$ to be the consecutive averaging instants to analyze the behaviour of $\Delta_t^m$ across two averaging instants.  Consequently, we can rewrite the above equation as
\begin{align}
    \Delta_t^m(a) & = \varphi_{t', t}  \barDelta_{t'}(a) +  \left[\sum_{k = t' + 1}^t  \widetilde{\eta}_k^{(t)}\gamma p\Delta_{k-1, \max}^m \right] + \xi_{t',t}^m(a). \label{eqn:delta_m_unaveraged}
\end{align}
Furthermore, after averaging, we obtain,
\begin{align}
    \barDelta_t(a) & =  \varphi_{t', t} \barDelta_{t'}(a) +  \frac{1}{M} \sum_{m = 1}^M \left[\sum_{k = t' + 1}^t  \widetilde{\eta}_k^{(t)} \gamma p\Delta_{k-1, \max}^m \right] + \frac{1}{M} \sum_{m = 1}^M \xi_{t',t}^m(a). \label{eqn:delta_m_averaged}
\end{align}

\paragraph{Step 2: deriving a recursive bound for $ \E[\barDelta_{t, \max}] $.} 
Bounding \eqref{eqn:delta_m_unaveraged}, we obtain,
\begin{subequations}\label{eqn:delta_max_unaveraged}
\begin{align}
    \Delta_{t, \max}^m & \geq \varphi_{t', t}  \barDelta_{t', \max} +  \left[\sum_{k = t' + 1}^t  \widetilde{\eta}_k^{(t)}\gamma  p\Delta_{k-1, \max}^m  \right] + \xi_{t',t, \max}^m - \varphi_{t', t} |\barDelta_{t'}(1) - \barDelta_{t'}(2)| , \label{eqn:delta_max_unaveraged_lower_bound} \\
    \Delta_{t, \max}^m & \leq \varphi_{t', t}  \barDelta_{t', \max} +  \left[\sum_{k = t' + 1}^t  \widetilde{\eta}_k^{(t)}\gamma  p\Delta_{k-1, \max}^m  \right] + \xi_{t',t, \max}^m   ,\label{eqn:delta_max_unaveraged_upper_bound}
\end{align}
\end{subequations}
where in the first step we used the fact that
\begin{align}
    \max\{a_1 + b_1, a_2 + b_2\} \geq \min\{a_1, a_2\} + \max\{b_1, b_2\} = \max\{a_1, a_2\} + \max\{b_1, b_2\} - |a_1 - a_2|. \label{eqn:max_decomposition}
\end{align}
On taking expectation, we obtain,
\begin{subequations} \label{eqn:delta_max_unaveraged_expectation}
\begin{align}
    \E[\Delta_{t, \max}^m] & \geq \varphi_{t', t}  \E[\barDelta_{t', \max}] +  \left[\sum_{k = t' + 1}^t  \widetilde{\eta}_k^{(t)}\gamma  p\E[\Delta_{k-1, \max}^m]  \right] + \E[\xi_{t',t, \max}^m] - \varphi_{t', t} \E[|\barDelta_{t'}(1) - \barDelta_{t'}(2)|] , \label{eqn:delta_max_unaveraged_expectation_lower_bound} \\
    \E[\Delta_{t, \max}^m] & \leq \varphi_{t', t}  \E[\barDelta_{t', \max}] +  \left[\sum_{k = t' + 1}^t  \widetilde{\eta}_k^{(t)}\gamma  p\E[\Delta_{k-1, \max}^m]  \right] + \E[\xi_{t',t, \max}^m]   .\label{eqn:delta_max_unaveraged_expectation_upper_bound}
\end{align}
\end{subequations}


Similarly, using \eqref{eqn:delta_m_averaged} and~\eqref{eqn:max_decomposition} we can write,
\begin{subequations}
\begin{align}
    \barDelta_{t, \max} & \geq \varphi_{t', t}  \barDelta_{t', \max} +  \frac{1}{M} \sum_{m = 1}^M \left[\sum_{k = t' + 1}^t  \widetilde{\eta}_k^{(t)}\gamma  p\Delta_{k-1, \max}^m  \right] - \varphi_{t', t} |\barDelta_{t'}(1) - \barDelta_{t'}(2)|   \nonumber \\
    & ~~~~~~~~~~~~~~~~~~~~~~~~~~~~~ + \max\left\{ \frac{1}{M}\sum_{m = 1}^M \xi_{t',t}^m(1) , \frac{1}{M}\sum_{m = 1}^M \xi_{t',t}^m(2) \right\}  \label{eqn:delta_max_averaged} \\ 
    \implies \E[\barDelta_{t, \max}] & \geq \varphi_{t', t}  \E[\barDelta_{t', \max}] + \frac{1}{M} \sum_{m = 1}^M \left[\sum_{k = t' + 1}^t  \widetilde{\eta}_k^{(t)}\gamma  p \E[\Delta_{k-1, \max}^m]  \right] - \varphi_{t', t} \E[|\barDelta_{t'}(1) - \barDelta_{t'}(2)|] \nonumber  \\
    & ~~~~~~~~~~~~~~~~~~~~~~~~~~~~~ + \E\left[\max\left\{ \frac{1}{M}\sum_{m = 1}^M \xi_{t',t}^m(1), \frac{1}{M}\sum_{m = 1}^M \xi_{t',t}^m(2)  \right\}\right] .\label{eqn:delta_max_averaged_expectation} 
\end{align}
\end{subequations}
On combining \eqref{eqn:delta_max_unaveraged_expectation_upper_bound} and~\eqref{eqn:delta_max_averaged_expectation}, we obtain,
\begin{align}
    \E[\barDelta_{t, \max}] & \geq \frac{1}{M} \sum_{m = 1}^M \left[\E[\Delta_{t, \max}^m] - \E[\xi_{t',t, \max}^m]\right] - \varphi_{t', t} \E[|\barDelta_{t'}(1) - \barDelta_{t'}(2)|] \nonumber \\
    & ~~~~~~~~~~~~~~~~~~~~~~~~~~~~~~~~~~ + \E\left[\max\left\{ \frac{1}{M}\sum_{m = 1}^M \xi_{t',t}^m(1), \frac{1}{M}\sum_{m = 1}^M \xi_{t',t}^m(2) \right\}\right]. \label{eqn:barDelta_max_as_func_clients} 
\end{align}


In order to simplify \eqref{eqn:barDelta_max_as_func_clients}, we make use of the following lemmas.

\begin{lemma}
    Let $t' < t$ be two consecutive averaging instants.
    Then for all $m \in [M]$, 
    \begin{align*}
        \E[\Delta_{t, \max}^m] - \E[\xi_{t',t, \max}^m] & \geq  \left(\prod_{k = t' + 1}^t (1 - \eta_k(1-\gamma p))\right) \E[\barDelta_{t', \max}] + \E[\xi_{t',t, \max}^m] \left[\sum_{k = t' + 1}^t \eta_k^{(t)} - 1  \right]_{+} \nonumber \\
        &~~~~~~~~~~~~~~~~~~~~~~~~~~~~~~ - \varphi_{t', t} \E[|\barDelta_{t'}(1) - \barDelta_{t'}(2)|], 
    \end{align*}
    where $[x]_{+} = \max\{x, 0\}$.
    \label{lemma:delta_t_max_minus_xi_tt_max}
\end{lemma}

\begin{lemma}
For all consecutive averaging instants $t', t$ satisfying $t - \max\{t' , \tau\} \geq 1/\eta_{\tau}$ and all $m \in [M]$,  we have,
    \begin{align*}
        \E[\xi_{t',t, \max}^m] & \geq \frac{1}{240 \log\left(\frac{180B}{\eta_T(1-\gamma)} \right)} \cdot \frac{\nu}{\nu + 1} , \\ 
        \E\left[\max\left\{ \frac{1}{M}\sum_{m = 1}^M \xi_{t',t}^m(1), \frac{1}{M}\sum_{m = 1}^M \xi_{t',t}^m(2) \right\}\right] & \geq \frac{1}{240 \log\left(\frac{180BM}{\eta_T(1-\gamma)} \right)} \cdot \frac{\nu}{\nu + \sqrt{M}}, 
    \end{align*}
    where $\nu := \sqrt{\dfrac{20\eta_T}{B(1-\gamma)}}$.
    \label{lemma:xi_lower_bound}
\end{lemma}

\begin{lemma}
    For all $t \in \{t_r\}_{r = 1}^R$, we have
    \begin{align*}
         \E[|\barDelta_{t}(1) - \barDelta_{t}(2)|] \leq \sqrt{ \frac{8\eta_T}{3BM(1-\gamma)}}.
    \end{align*}
    \label{lemma:barDelta_action_diff}
\end{lemma}


Thus, on combining the results from Lemmas~\ref{lemma:delta_t_max_minus_xi_tt_max},~\ref{lemma:xi_lower_bound}, and~\ref{lemma:barDelta_action_diff} and plugging them into \eqref{eqn:barDelta_max_as_func_clients}, we obtain the following relation for $t, t' \geq \tau$:  
\begin{align}
  \E[\barDelta_{t, \max}]    & \geq    \left(\prod_{k = t' + 1}^t (1 - \eta_k(1-\gamma p))\right) \E[\barDelta_{t', \max}] + \E[\xi_{t',t, \max}^m] \left[\sum_{k = t' + 1}^t \eta_k^{(t)} - 1  \right]_{+} - 2\varphi_{t', t} \E[|\barDelta_{t'}(1) - \barDelta_{t'}(2)|] \nonumber \\
    & ~~~~~~~~~~~~~~~~~~~~~~~~~~~~~~~~~~ + \E\left[\max\left\{ \frac{1}{M}\sum_{m = 1}^M \xi_{t',t}^m(1), \frac{1}{M}\sum_{m = 1}^M \xi_{t',t}^m(2) \right\}\right] \nonumber \\ 
 & \geq  (1 - \eta_{\tau}(1-\gamma p))^{t - t'} \E[\barDelta_{t', \max}]  + \left(\frac{ 1-  (1- \eta_{\tau} (1- \gamma p))^{t - t'}}{5760 \log\left(\frac{180B}{\eta_T(1-\gamma)} \right)(1 -\gamma p)}\right) \cdot \frac{\nu}{\nu + 1} \cdot \1\left\{t - t' \geq \frac{8}{\eta_{\tau}} \right\} \nonumber \\
    & ~~~~~~~ - 2(1 - \eta_T)^{t - t'} \sqrt{ \frac{8\eta_T}{3BM(1-\gamma)}} +  \frac{1}{240 \log\left(\frac{180BM}{\eta_T(1-\gamma)} \right)} \cdot \frac{\nu}{\nu + \sqrt{M}} \cdot \1\left\{t - t' \geq \frac{8}{\eta_{\tau}} \right\},\label{eqn:barDelta_recursion}
\end{align}
where we used the relation $\varphi_{t',t} \leq (1 - \eta_T)^{t- t'}$, as well as the value of $\nu$ as defined in Lemma~\ref{lemma:xi_lower_bound} along with the fact    
\begin{align} \label{eq:lr_sour}
        \sum_{k = t' + 1}^t \eta_k^{(t)} - 1 \geq \frac{1 - (1-\eta_{\tau}(1-\gamma p))^{t - t'}}{24(1-\gamma p)}
    \end{align} 
for all $t, t' \geq \tau$ such that $t - t' \geq 8/\eta_{\tau}$.

\paragraph{Proof of \eqref{eq:lr_sour}.}
We have,
\begin{align}
    \sum_{k  = t' + 1}^{t} \eta_k^{(t)} - 1 & = \sum_{k  = t' + 1}^{t} \left(\eta_k \prod_{i = k + 1}^t(1 - \eta_i(1-\gamma p))\right) - 1 \nonumber \\
    & \geq \sum_{k  = t' + 1}^{t} \left(\eta_t \prod_{i = k + 1}^t(1 - \eta_{\tau}(1-\gamma p))\right) - 1 \nonumber \\
    & \geq \eta_t  \sum_{k  = t' + 1}^{t} (1 - \eta_{\tau}(1-\gamma p))^{t - k} - 1 \nonumber \\
    & \geq \eta_t \cdot \left( \frac{1 - (1-\eta_{\tau}(1 - \gamma p))^{t - t'}}{\eta_{\tau}(1- \gamma p)}\right) - 1 \nonumber \\
    & \geq \frac{1 - (1-\eta_{\tau}(1 - \gamma p))^{t - t'}}{3(1- \gamma p)} - 1.
\end{align}
To show \eqref{eq:lr_sour}, it is sufficient to show that $\displaystyle \frac{1 - (1-\eta_{\tau}(1 - \gamma p))^{t - t'}}{3(1- \gamma p)} \geq \frac{8}{7}$ for $t - t' \geq 8/\eta_{\tau}$. Thus, for $t - t' \geq 8/\eta_{\tau}$ we have,
\begin{align}
    \frac{1 - (1-\eta_{\tau}(1 - \gamma p))^{t - t'}}{3(1- \gamma p)} & \geq \frac{1 - \exp(-\eta_{\tau}(1 - \gamma p)\cdot(t - t'))}{3(1- \gamma p)} \nonumber \\
    & \geq \frac{1 - \exp(-8(1 - \gamma p))}{3(1- \gamma p)}.
\end{align}
Since $\gamma \geq 5/6$, $1 - \gamma p \leq 2/9$. For $x \leq 2/9$, the function $f(x) = \frac{1 - e^{-8x}}{3x} \geq 8/7$, proving the claim.


\paragraph{Step 3: lower bounding $\E[\barDelta_{T, \max}]$.} We are now interested in evaluating $\E[\barDelta_{T, \max}]$ based on the recursion \eqref{eqn:barDelta_recursion}. To this effect, we introduce some notation to simplify the presentation.
Let 
\begin{align}\label{eq:lemonade}
R_{\tau} & := \min \{r : t_r \geq \tau \}.
\end{align} 
 For $r = R_{\tau},\dots, R$, we define the following terms:
\begin{align*}
    x_r & := \E[\barDelta_{t_r, \max}] ,\\
    \alpha_r & := (1 - \eta_{\tau}(1-\gamma p))^{t_r - t_{r-1}}, \\
    \beta_r & := (1 - \eta_T)^{t_r - t_{r-1}}, \\
    \cI_r & := \{r \geq r' > R_{\tau} : t_{r'} - t_{r'-1} \geq 8/\eta_{\tau}\} ,\\
    C_1 & :=  \frac{1}{5760 \log\left(\frac{180B}{\eta_T(1-\gamma)} \right)(1-\gamma p)} \cdot \frac{\nu}{\nu + 1} , \\
    C_2 & := \sqrt{ \frac{32\eta_T}{3BM(1-\gamma)}} ,\\
    C_3 & := \frac{1}{240 \log\left(\frac{180BM}{\eta_T(1-\gamma)} \right)} \cdot \frac{\nu}{\nu + \sqrt{M}}.
\end{align*}
With these notations in place, the recursion in \eqref{eqn:barDelta_recursion} can be rewritten as
\begin{align}
    x_r \geq \alpha_r x_{r-1} - \beta_r C_2 + C_3 \1 \{r \in \cI_r \} + (1 - \alpha_r) C_1 \1 \{r \in \cI_r \}, \label{eqn:x_recursion}
\end{align}
for all $r \geq R_{\tau}$. We claim that $x_r$ satisfies the following relation for all $r \geq R_{\tau} + 1$ (whose proof is deferred to the end of this step):
\begin{align}
    x_r & \geq \left( \prod_{i = R_{\tau}+1}^r \alpha_i \right) x_{R_{\tau}} - \sum_{k = R_{\tau}+1}^r  \beta_k \left( \prod_{i = k+1}^{r} \alpha_i \right) C_2 + \sum_{k = R_{\tau} +1 }^r   \left( \prod_{i = k+1}^{r} \alpha_i \right)  \1\{k \in \cI_k\} C_3 \nonumber\\
    &~~~~~~~~~~~~~~~~~~~~~~~~~~~~~~~~~~~~ + C_1 \left( \prod_{i \notin \cI_r} \alpha_i \right) \left( 1 - \prod_{i \in \cI_r} \alpha_i \right), \label{eqn:x_recursion_solved}
\end{align}
where we recall that if there is no valid index for a product, its value is taken to be $1$.

Invoking \eqref{eqn:x_recursion_solved} for $r = R$ and using the relation $x_{R_{\tau} -1 } \geq 0$, we obtain,
\begin{align}
    x_R & \geq - \sum_{k = R_{\tau}}^{R}  \beta_k \left( \prod_{i = k+1}^{R} \alpha_i \right) C_2 + \sum_{k = R_{\tau} }^R   \left( \prod_{i = k+1}^{R} \alpha_i \right) C_3 \1\{k \in \cI_k\} + C_1 \left( \prod_{i \notin \cI_R} \alpha_i \right) \left( 1 - \prod_{i \in \cI_R} \alpha_i \right) \nonumber \\
    & \geq - R C_2  + C_1 \left( \prod_{i \notin \cI_R} \alpha_i \right) \left( 1 - \prod_{i \in \cI_R} \alpha_i \right) \nonumber \\
    & \geq - R \cdot \sqrt{ \frac{32\eta_T}{3BM(1-\gamma)}}  +  \left( \prod_{i \notin \cI_R} \alpha_i \right) \left( 1 - \prod_{i \in \cI_R} \alpha_i \right) \cdot \frac{1}{5760 \log\left(\frac{180B}{\eta_T(1-\gamma)} \right)(1-\gamma p)} \cdot \frac{\nu}{\nu + 1}, \label{eqn:x_R_final}
\end{align}
where we used the fact $\beta_k \left( \prod_{i = k+1}^{R} \alpha_i \right) \leq 1$ and that $C_3 \geq 0$. Consider the expression
\begin{align}
    \prod_{i \notin \cI_R} \alpha_i & = \prod_{i \notin \cI_R} (1 - \eta_{\tau}(1- \gamma p))^{t_{i} - t_{i-1}} \geq 1 - \eta_{\tau}(1- \gamma p) \cdot \underbrace{\sum_{i \notin \cI_R} (t_{i} - t_{i-1})}_{ =: T_1} . \label{eqn:alpha_i_prod}
\end{align}
Consequently, 
\begin{align}
    \left( 1 - \prod_{i \in \cI_R} \alpha_i \right) = 1 - (1 - \eta_{\tau}(1- \gamma p))^{T -\tau - T_1} \geq 1 - \exp \left( - \eta_{\tau} (1- \gamma p) \left(T - \tau - T_1 \right)\right) .\label{eqn:one_minus_alpha_i_prod}
\end{align}
Note that $T_1$ satisfies the following bound
\begin{align}
    T_1 : = \sum_{i \notin \cI_R} (t_{i} - t_{i-1}) \leq (R - |\cI_R|) \cdot \frac{8}{\eta_{\tau}} \leq \frac{8R}{\eta_{\tau}}. \label{eqn:T_1_bound}
\end{align}

We split the remainder of the analysis based on the step size schedule. 
\begin{itemize}
\item 
For the constant step size schedule, i.e., $\eta_t = \eta \geq \frac{1}{(1-\gamma)T}$, we have, $R_{\tau} = 0$, with $\tau=0$ and $t_0 = 0$ (as all agents start at the same point). 
  If $R \leq \frac{1}{96000(1-\gamma) \log\left(\frac{180 B}{\eta (1-\gamma)}\right)}$, then, \eqref{eqn:alpha_i_prod},~\eqref{eqn:one_minus_alpha_i_prod} and~\eqref{eqn:T_1_bound} yield the following relations:
\begin{align*}
    T_1 & \leq \frac{8R}{\eta} \leq \frac{T}{12000\log(180 N)}, \\
    \prod_{i \notin \cI_R} \alpha_i & \geq 1 - \eta (1- \gamma p) \cdot T_1 \geq 1 - \frac{32R(1 -\gamma)}{3} \geq 1 - \frac{1}{9000 \log(180 N)} ,\\
    \left( 1 - \prod_{i \in \cI_R} \alpha_i \right) & \geq 1 - \exp \left( - \eta (1- \gamma p) \left(T  - T_1 \right)\right) \geq 1 - \exp \left( -  \frac{4}{3}\left(1  - \frac{1}{9000\log(180 N)} \right)\right).
\end{align*}
On plugging the above relations into \eqref{eqn:x_R_final}, we obtain
\begin{align}
    x_R & \geq \frac{\sqrt{40}}{96000\log\left(\frac{180 B}{\eta(1-\gamma)} \right)(1-\gamma)} \cdot \left( \frac{\nu}{\nu + 1} - \frac{\nu}{5\sqrt{M}}  \right) \label{eqn:x_R_bound_constant_generic} 
\end{align}
where recall that $\nu := \sqrt{\dfrac{20 \eta}{3B(1-\gamma)}}$. Consider the function $f(x) = \frac{x}{x + 1} - \frac{x}{5\sqrt{M}}$ . We claim that for $x \in [0, \sqrt{M}]$ and all $M \geq 2$, 
\begin{align}
    f(x) \geq \frac{7}{20} \min\{x, 1\}. \label{eqn:x_R_constant_additional_claim}
\end{align}
The proof of the above claim is deferred to the end of the section. In light of the above claim, we have,
\begin{align}
    x_R & \geq \frac{\sqrt{40}}{96000\log\left(\frac{180 B}{\eta(1-\gamma)} \right)(1-\gamma)} \cdot \frac{7}{20} \cdot \min\left\{1, \sqrt{\frac{20\eta}{3B(1-\gamma)}}\right\} \nonumber \\
    & \geq \frac{\sqrt{40}}{96000\log\left(180 N \right)} \cdot \frac{7}{20} \cdot  \min\left\{ \frac{1}{1-\gamma}, \sqrt{\frac{20}{3(1-\gamma)^4 N}}\right\}, \label{eqn:x_R_bound_constant}
\end{align}
where we used the fact that $M \geq 2$, $\frac{\sqrt{x}}{\log(1/x)}$ is an increasing function and the relation $\dfrac{\nu}{M} = \dfrac{20\eta}{3BM(1-\gamma)} \leq \dfrac{1}{15} \leq 1$.

\item Next, we consider the rescaled linear step size schedule, where $\tau = T/3$ (cf.~\eqref{eqn:tau_def}). To begin, we  assume $t_{R_{\tau}} \leq \max\{ \frac{3T}{4}, T - \frac{1}{6\eta_{\tau}(1-\gamma p)}\}$.   It is straightforward to note that 
\begin{align*}
    \max\left\{ \frac{3T}{4}, T - \frac{1}{6\eta_{\tau}(1-\gamma p)}\right\} = \begin{cases} \frac{3T}{4} & \text{ if } c_{\eta} \geq 3 \\ T - \frac{1}{6\eta_{\tau}(1-\gamma p)} & \text{ if } c_{\eta} < 3. \end{cases}
\end{align*}
If $R \leq \frac{1}{384000(1-\gamma) \log\left(\frac{180 B}{\eta_T (1-\gamma)}\right) \cdot (5 + c_{\eta})}$ then, \eqref{eqn:alpha_i_prod},~\eqref{eqn:one_minus_alpha_i_prod} and~\eqref{eqn:T_1_bound} yield the following relations:
\begin{align*}
    T_1 & \leq \frac{8R}{\eta_{\tau}} , \qquad 
    \prod_{i \notin \cI_R} \alpha_i  \geq 1 - \eta_{\tau} (1- \gamma p) \cdot T_1 \geq 1 - \frac{32R(1 -\gamma)}{3} \geq 1 - \frac{1}{36000}.
\end{align*}
For $c_{\eta} \geq 3$, we have,
\begin{align*}
    \left( 1 - \prod_{i \in \cI_R} \alpha_i \right) & \geq 1 - \exp \left( - \eta_{\tau} (1- \gamma p) \left(T - t_{R_{\tau}}  - T_1 \right)\right) \\
    & \geq 1 - \exp \left( -\frac{(1-\gamma)T}{(3 + c_{\eta}(1-\gamma)T)} + \frac{32R(1-\gamma)}{3} \right) \\
    & \geq \frac{1}{2(3 + c_{\eta})},
\end{align*}
where we used $T \geq \frac{1}{1- \gamma}$ in the second step. Similarly, for $c_{\eta} < 3$, we have,
\begin{align*}
    \left( 1 - \prod_{i \in \cI_R} \alpha_i \right) & \geq 1 - \exp \left( - \eta_{\tau} (1- \gamma p) \left(T - t_{R_{\tau}}  - T_1 \right)\right) \\
    & \geq 1 - \exp \left( -\frac{1}{6} + \frac{32R(1-\gamma)}{3} \right) \\
    & \geq \frac{1}{10}.
\end{align*}
On plugging the above relations into \eqref{eqn:x_R_final}, we obtain
\begin{align}
    x_R & \geq \frac{18\sqrt{1.6}}{384000\log\left(\frac{180B}{\eta_T(1-\gamma)} \right)(1-\gamma)(5+ c_{\eta})} \cdot  \left( \frac{\nu}{\nu  +1} - \frac{\nu}{18{\sqrt{M}}}  \right) \nonumber \\
    & \geq \frac{18\sqrt{1.6}}{384000\log\left(\frac{180B}{\eta_T(1-\gamma)} \right)(1-\gamma)(5+ c_{\eta})} \cdot \frac{7}{20}  \cdot \min\left\{1, \sqrt{\frac{20\eta_T}{3B(1-\gamma)}}\right\} \nonumber \\
    & \geq \frac{18\sqrt{1.6}}{384000\log\left(\frac{180B}{\eta_T(1-\gamma)} \right)(5+ c_{\eta})} \cdot \frac{7}{20}  \cdot \min\left\{\frac{1}{1-\gamma}, \sqrt{\frac{20\eta_T}{3B(1-\gamma)^3}}\right\} \nonumber \\
    & \geq \frac{18\sqrt{1.6}}{384000\log\left(180N(1 + \log N) \right)(5+ \log N)} \cdot \frac{7}{20}  \cdot \min\left\{\frac{1}{1-\gamma}, \sqrt{\frac{20}{3B(1+\log N)(1-\gamma)^4N}}\right\}, 
    \label{eqn:x_R_bound_linear}
\end{align}
where we again used the facts that $M \geq 2$, $c_{\eta} \leq \log N$, $\frac{\sqrt{x}}{\log(1/x)}$ is an increasing function and the relation $\dfrac{\nu}{M} = \dfrac{20\eta_T}{3BM(1-\gamma)} \leq 1$. 

\item Last but not least, let us consider the rescaled linear step size schedule case when $t_{R_{\tau}} > \max\{ \frac{3T}{4}, T - \frac{1}{6\eta_{\tau}(1-\gamma p)}\}$. The condition implies that the time between the communication rounds $R_{\tau} - 1$ and $R_{\tau}$ is at least $T_0 := \max\{ \frac{5T}{12},  \frac{2T}{3} - \frac{1}{6\eta_{\tau}(1-\gamma p)}\}$. Thus, \eqref{eqn:barDelta_recursion} yields that
\begin{align}
    \E[\barDelta_{t_{R_{\tau}}}] & \geq \left(\frac{ 1-  (1- \eta_{\tau} (1- \gamma p))^{T_0}}{5760 \log\left(\frac{180}{B\eta_T(1-\gamma)} \right)(1 -\gamma p)}\right) \cdot \frac{\nu}{\nu + 1}  - 2(1 - \eta_T)^{T_0} \sqrt{ \frac{8\eta_T}{3BM(1-\gamma)}}. 
\end{align}
Using the above relation along with \eqref{eqn:x_recursion_solved}, we can conclude that
\begin{align}
    x_R & \geq  (1- \eta_{\tau} (1- \gamma p))^{T - t_{R_{\tau}}}\left(\frac{ 1-  (1- \eta_{\tau} (1- \gamma p))^{T_0}}{5760 \log\left(\frac{180}{B\eta_T(1-\gamma)} \right)(1 -\gamma p)}\right) \cdot \frac{\nu}{\nu + 1} \nonumber \\
    & ~~~~~~~~~~~~ - 2(1 - \eta_T)^{T_0} \cdot (1- \eta_{\tau} (1- \gamma p))^{T - t_{R_{\tau}}} \sqrt{ \frac{8\eta_T}{3BM(1-\gamma)}}  - RC_2. \label{eqn:x_R_final_case_2}
\end{align}
In the above relation, we used the trivial bounds $C_1, C_3 \geq 0$ and a crude bound on the term corresponding to $C_2$, similar to \eqref{eqn:x_R_final}. Let us first consider the case of $c_{\eta} \geq 3$. We have,
\begin{align*}
    1 - (1- \eta_{\tau} (1- \gamma p))^{T_0} & \geq 1 - \exp\left(-\eta_{\tau} (1- \gamma p) 5T/12\right) \geq 1 - \exp\left(-\frac{ 5(1- \gamma)T}{3(3 + c_{\eta}(1-\gamma)T)}\right)  \geq \frac{1}{3 + c_{\eta}} ,\\
    (1- \eta_{\tau} (1- \gamma p))^{T - t_{R_{\tau}}} & \geq 1 - \eta_{\tau} (1-\gamma p) \frac{T}{4} \geq 1 - \frac{ (1- \gamma)T}{(3 + c_{\eta}(1-\gamma)T)} \geq 1 - \frac{1}{c_{\eta}} \geq \frac{2}{3}.
\end{align*}
Similarly, for $c_{\eta} < 3$, we have,
\begin{align*}
    1 - (1- \eta_{\tau} (1- \gamma p))^{T_0} & \geq 1 - \exp\left(-\eta_{\tau} (1- \gamma p) \frac{2T}{3} + \frac{1}{6} \right) \nonumber \\
    & \geq 1 - \exp\left(-\frac{ 8(1- \gamma)T}{3(3 + c_{\eta}(1-\gamma)T)} + \frac{1}{6} \right)  \geq 1- e^{-5/18} ,\\
    (1- \eta_{\tau} (1- \gamma p))^{T - t_{R_{\tau}}} & \geq 1 - \frac{\eta_{\tau} (1-\gamma p)}{6\eta_{\tau} (1-\gamma p)}  \geq \frac{5}{6}.
\end{align*}
The above relations implies that $(1- \eta_{\tau} (1- \gamma p))^{T - t_{R_{\tau}}}(1-  (1- \eta_{\tau} (1- \gamma p))^{T_0}) \geq c$ for some constant $c$, which only depends on $c_{\eta}$. On plugging this into \eqref{eqn:x_R_final_case_2}, we obtain a relation that is identical to that in \eqref{eqn:x_R_final} up to leading constants. Thus, by using a similar sequence of argument as used to obtain \eqref{eqn:x_R_bound_linear}, we arrive at the same conclusion as for the case of $t_{R_{\tau}} \leq \max\{ \frac{3T}{4}, T - \frac{1}{6\eta_{\tau}(1-\gamma p)}\}$.
\end{itemize}

\paragraph{Step 4: finishing up the proof.}
Thus, \eqref{eqn:x_R_bound_constant},~\eqref{eqn:x_R_bound_linear} along with the above conclusion together imply that there exists a numerical constant $c_0 > 0$ such that 
\begin{align}
    \E[|\hatV_T^m(1) - V^{\star}(1)|] \geq \E[\barDelta_{T, \max}] \geq \frac{c_0}{\log^3 N} \cdot \min\left\{\frac{1}{1-\gamma}, \sqrt{ \frac{1}{(1-\gamma)^4 N}} \right\}.
\end{align}
The above equation along with Lemma~\ref{lemma:Q_hatQ_difference} implies
\begin{align}
    \E[|V_T^m - V^{\star}(1)|] \geq \frac{c_0}{\log^3 N} \cdot \min\left\{\frac{1}{1-\gamma}, \sqrt{ \frac{1}{(1-\gamma)^4 N}} \right\} - \frac{1}{1 - \gamma} \prod_{i = 1}^T (1- \eta_i(1-\gamma)). \label{eqn:v_1_final_lower_bound}
\end{align}
On the other hand, from \eqref{eqn:state_3_recursion} we know that
\begin{align}
    \E[|V_T^m(3) - V^{\star}(3)|] \geq \frac{1}{1 - \gamma} \prod_{i = 1}^T (1- \eta_i(1-\gamma)).\label{eqn:v_3_final_lower_bound}
\end{align}
Hence,
\begin{align}
    \E[\|Q_T^m - Q^{\star}\|_{\infty}] & \geq \E\left[ \max\left\{|V_T^m(3) - V^{\star}(3)|, |V_T^m(1) - V^{\star}(1)| \right\} \right] \nonumber \\
    & \geq \max\left\{ \E\left[ |V_T^m(3) - V^{\star}(3)| \right], \E\left[ |V_T^m(1) - V^{\star}(1)|  \right] \right\} \nonumber \\
    & \geq \max\left\{ \frac{1}{1 - \gamma} \prod_{i = 1}^T (1- \eta_i(1-\gamma)), \min\left\{\frac{1}{1-\gamma}, \sqrt{ \frac{1}{(1-\gamma)^4 N}} \right\} - \frac{1}{1 - \gamma} \prod_{i = 1}^T (1- \eta_i(1-\gamma)) \right\} \nonumber \\
    & \geq \frac{1}{2} \min\left\{\frac{1}{1-\gamma}, \sqrt{ \frac{1}{(1-\gamma)^4 N}} \right\}, \label{eqn:q_err_large_lr_final_bound}
\end{align}
where the third step follows from \eqref{eqn:v_1_final_lower_bound} and~\eqref{eqn:v_3_final_lower_bound} and the fourth step uses $\max\{a,b\} \geq (a +b)/2$. \\

Thus, from \eqref{eqn:q_err_small_lr_final_bound} and~\eqref{eqn:q_err_large_lr_final_bound} we can conclude that whenever $\CC_{\textsf{round}} = \cO\left( \frac{1}{(1-\gamma) \log^2 N} \right)$, $\ER(\sA; N, M) = \Omega\left(\frac{1}{\log^3 N \sqrt{N}} \right)$ for all values of $M \geq 2$.
In other words, for any algorithm to achieve any collaborative gain, its communication complexity should satisfy $\CC_{\textsf{round}} = \Omega\left( \frac{1}{(1-\gamma) \log^2 N} \right)$, as required.


\paragraph{Proof of \eqref{eqn:x_recursion_solved}.} We now return to establish \eqref{eqn:x_recursion_solved} using induction. For the base case, \eqref{eqn:x_recursion} yields
\begin{align}
    x_{R_{\tau}+1} & \geq \alpha_{R_{\tau}+1} x_{R_{\tau}} - \beta_{R_{\tau}+1} C_2 + C_3 \1\{R_{\tau} + 1 \in \cI_{R_{\tau} + 1}\} + (1 - \alpha_{R_{\tau}+1}) C_1 \1 \{{R_{\tau}+1} \in \cI_{R_{\tau}+1} \}.
\end{align}
Note that this is identical to the expression in \eqref{eqn:x_recursion_solved} for $r = R_{\tau} +1$ as 
\begin{align*}
    \left( \prod_{i \notin \cI_{R_{\tau}+1}} \alpha_i \right) \left( 1 - \prod_{i \in \cI_{R_{\tau}+1}} \alpha_i \right) = (1 - \alpha_{R_{\tau}+1}) \1 \{{R_{\tau}+1} \in \cI_{R_{\tau}+1} \}
\end{align*}
based on the adopted convention for products with no valid indices. For the induction step, assume \eqref{eqn:x_recursion_solved} holds for some $r \geq R_\tau +1$. On combining \eqref{eqn:x_recursion} and~\eqref{eqn:x_recursion_solved}, we obtain,
\begin{align}
    x_{r+1} & \geq \alpha_{r+1} x_{r} - \beta_{r+1} C_2 + C_3 \1 \{(r+1) \in \cI_{r+1} \}  + (1 - \alpha_{r+1}) C_1 \1 \{r+1 \in \cI_{r+1} \} \nonumber \\
    & \geq \alpha_{r+1} \left( \prod_{i = R_{\tau} + 1}^r \alpha_i \right) x_{R_{\tau}} - \alpha_{r+1}  \sum_{k = R_{\tau} + 1}^r  \beta_k \left( \prod_{i = k+1}^{r} \alpha_i \right) C_2 + \alpha_{r+1} \sum_{k = R_{\tau} + 1}^r   \left( \prod_{i = k+1}^{r} \alpha_i \right) C_3 \1\{k \in \cI_k\}  \nonumber \\
    &  + \alpha_{r+1} C_1 \left( \prod_{i \notin \cI_r} \alpha_i \right) \left( 1 - \prod_{i \in \cI_r} \alpha_i \right) - \beta_{r+1} C_2 + C_3 \1 \{(r+1) \in \cI_{r+1} \} + (1 - \alpha_{r+1}) C_1 \1 \{(r+1) \in \cI_{r+1} \} \nonumber \\
    & \geq \left( \prod_{i = R_{\tau} + 1}^{r+1} \alpha_i \right) x_{R_{\tau}} -  \sum_{k = R_{\tau} + 1}^{r+1}  \beta_k \left( \prod_{i = k+1}^{r+1} \alpha_i \right) C_2 + \sum_{k = R_{\tau} + 1}^{r+1}   \left( \prod_{i = k+1}^{r+1} \alpha_i \right) C_3 \1\{k \in \cI_k\} \nonumber \\
    & ~~~~~~~~~ + \alpha_{r+1} C_1 \left( \prod_{i \notin \cI_r} \alpha_i \right) \left( 1 - \prod_{i \in \cI_r} \alpha_i \right)  + (1 - \alpha_{r+1}) C_1 \1 \{(r+1) \in \cI_{r+1} \}. \label{eqn:x_induction_step}
\end{align}
If $(r+1) \notin \cI_{r+1}$, then $\left( 1 - \prod_{i \in \cI_r} \alpha_i \right) = \left( 1 - \prod_{i \in \cI_{r+1}} \alpha_i \right)$ and $\alpha_{r+1} \left( \prod_{i \notin \cI_r} \alpha_i \right) = \left( \prod_{i \notin \cI_{r+1}} \alpha_i \right)$. Consequently, 
\begin{align}
    \alpha_{r+1} C_1 \left( \prod_{i \notin \cI_r} \alpha_i \right) \left( 1 - \prod_{i \in \cI_r} \alpha_i \right)  + (1 - \alpha_{r+1}) C_1 \1 \{(r+1) \in \cI_{r+1} \} = C_1 \left( \prod_{i \notin \cI_{r+1}} \alpha_i \right) \left( 1 - \prod_{i \in \cI_{r+1}} \alpha_i \right). \label{eqn:r_plus_one_not_in_I}
\end{align}
On the other hand, if $(r+1) \in \cI_{r+1}$, then $\left( \prod_{i \notin \cI_r} \alpha_i \right) = \left( \prod_{i \notin \cI_{r+1}} \alpha_i \right)$. Consequently, we have,
\begin{align}
    &\alpha_{r+1} C_1 \left( \prod_{i \notin \cI_r} \alpha_i \right) \left( 1 - \prod_{i \in \cI_r} \alpha_i \right)  + (1 - \alpha_{r+1}) C_1 \1 \{(r+1) \in \cI_{r+1} \} \nonumber \\
    &  = \alpha_{r+1} C_1 \left( \prod_{i \notin \cI_{r+1}} \alpha_i \right) \left( 1 - \prod_{i \in \cI_r} \alpha_i \right)  + (1 - \alpha_{r+1}) C_1 \nonumber \\
    &  \geq  C_1 \left( \prod_{i \notin \cI_{r+1}} \alpha_i \right) \left[\alpha_{r+1}\left( 1 - \prod_{i \in \cI_r} \alpha_i \right)  + (1 - \alpha_{r+1}) \right] \nonumber \\
    &  \geq  C_1 \left( \prod_{i \notin \cI_{r+1}} \alpha_i \right) \left( 1 - \prod_{i \in \cI_{r+1}} \alpha_i \right). \label{eqn:r_plus_one_in_I}
\end{align}
Combining \eqref{eqn:x_induction_step},~\eqref{eqn:r_plus_one_not_in_I} and~\eqref{eqn:r_plus_one_in_I} proves the claim. 

\paragraph{Proof of~\eqref{eqn:x_R_constant_additional_claim}.} To establish this result, we separately consider the cases $x\leq 1$ and $x \geq 1$.

\begin{itemize}
    \item When $x \leq 1$, we have
    \begin{align}
        f(x) = \frac{x}{x + 1} - \frac{1}{5\sqrt{M}} \geq x \cdot \left(\frac{1}{2} - \frac{x}{5\sqrt{M}}\right) \geq \frac{7x}{20}, \label{eqn:f_x_lower_bound_case_1}
    \end{align}
    where in the last step, we used the relation $M \geq 2$.
    \item Let us now consider the case $x\geq 1$. The second derivative of $f$ is given by $f''(x) = -\frac{1}{2(x+1)^3}$. Clearly, for all $x \geq 1$, $f'' < 0$ implying that $f$ is a concave function. It is well-known that a continuous, bounded, concave function achieves its minimum values over a compact interval at the end points of the interval (Bauer's minimum principle). For all $M \geq 2$, we have,
    \begin{align*}
        f(1)  = \frac{1}{2} - \frac{1}{5\sqrt{M}} \geq \frac{7}{20}; \quad
        f(\sqrt{M})  = \frac{\sqrt{M}}{\sqrt{M} + 1} - \frac{1}{5} \geq \frac{7}{20}.
    \end{align*}
    Consequently, we can conclude that for all $x \in [1,\sqrt{M}]$,
    \begin{align}
        f(x) \geq \frac{7}{20}. \label{eqn:f_x_lower_bound_case_2}
    \end{align}
    Combining~\eqref{eqn:f_x_lower_bound_case_1} and~\eqref{eqn:f_x_lower_bound_case_2} proves the claim.
\end{itemize}

\subsubsection{Large learning rates with large $\frac{\eta_T}{BM}$}


In order to bound the error in this scenario, note that $\frac{\eta_T}{BM}$ controls the variance of the stochastic updates in the fixed point iteration. Thus, when $\frac{\eta_T}{BM}$ is large, the variance of the iterates is large, resulting in a large error. To demonstrate this effect, we focus on the dynamics of state $2$. This part of the proof is similar to the large learning rate case of~\cite{Li2023QLMinimax}. For all $t \in [T]$, define:
\begin{align}
    \overline{V}_t(2) := \frac{1}{M} \sum_{m = 1}^M V_t^m(2). \label{eqn:bar_v_state_2_defn}
\end{align}
Thus, from~\eqref{eqn:avg_v_state_2_recursion}, we know that $\E[\overline{V}_t(2)]$ obeys the following recursion:
\begin{align*}
    \E[\overline{V}_t(2)] = (1-\eta_t(1-\gamma p)) \E[\overline{V}_{t-1}(2)] + \eta_t.
\end{align*}
Upon unrolling the recursion, we obtain,
\begin{align*}
    \E[\overline{V}_T(2)] = \left(\prod_{k = t+1}^T (1-\eta_k(1-\gamma p))\right) \E[\overline{V}_{t}(2)] + \sum_{k = t + 1}^T \eta_k^{(T)}.
\end{align*}
Thus, the above relation along with~\eqref{eqn:eta_k_t_recursion} and the value of $V^{\star}(2)$ yields us,
\begin{align}
    V^{\star}(2) - \E[\overline{V}_T(2)] = \prod_{k = t+1}^T (1-\eta_k(1-\gamma p)) \left(\frac{1}{1-\gamma p} - \E[\overline{V}_{t}(2)]\right). \label{eqn:bar_v_expected_err}
\end{align}

Similar to~\cite{Li2023QLMinimax}, we define
\begin{align*}
    \tau' := \min\left\{ 0 \leq t' \leq T-2 \ \bigg| \ \E[(\overline{V}_t)^2] \geq \frac{1}{4(1-\gamma)^2} \text{ for all } t' + 1 \leq t \leq T\right\}.
\end{align*}
If such a $\tau'$ does not exist, it implies that either $\E[(\overline{V}_T)^2] < \frac{1}{4(1-\gamma)^2}$ or $\E[(\overline{V}_{T-1})^2] < \frac{1}{4(1-\gamma)^2}$. If the former is true, then, 
\begin{align}
    V^{\star}(2) - \E[\overline{V}_T(2)] = \frac{3}{4(1-\gamma)} - \sqrt{\E[(\overline{V}_T)^2]} > \frac{1}{4(1-\gamma)}. \label{eqn:small_V_T}
\end{align}
Similarly, if $\E[(\overline{V}_{T-1})^2] < \frac{1}{4(1-\gamma)^2}$, it implies $\E[\overline{V}_{T-1}] < \frac{1}{2(1-\gamma)}$. Using~\eqref{eqn:avg_v_state_2_recursion}, we have,
\begin{align*}
    \E[\overline{V}_T(2)] = (1-\eta_T(1-\gamma p)) \E[\overline{V}_{T-1}(2)] + \eta_T \leq \E[\overline{V}_{T-1}(2)] +  1 < \frac{1}{2(1-\gamma)} + \frac{1}{6(1-\gamma)} = \frac{2}{3(1-\gamma)}.
\end{align*}
Consequently, 
\begin{align}
    V^{\star}(2) - \E[\overline{V}_T(2)] > \frac{3}{4(1-\gamma)} - \frac{2}{3(1-\gamma)}  > \frac{1}{12(1-\gamma)}. \label{eqn:small_V_T_minus_1}
\end{align}

For the case when $\tau'$ exists, we divide the proof into two cases.

\begin{itemize}
    \item We first consider the case when the learning rates satisfy:
    \begin{align}
        \prod_{k = \tau' + 1}^T (1-\eta_k(1-\gamma p)) \geq \frac{1}{2}. \label{eqn:tau_prime_to_T_more_than_half}
    \end{align}
    The analysis for this case is identical to that considered in~\cite{Li2023QLMinimax}. We explicitly write the steps for completeness. Specifically,
    \begin{align}
        V^{\star}(2) - \E[\overline{V}_T(2)] & = \left(\prod_{k = \tau'+1}^T (1-\eta_k(1-\gamma p)) \right)\left(\frac{1}{1-\gamma p} - \E[\overline{V}_{\tau'}(2)]\right) \nonumber \\
        & \geq \frac{1}{2} \cdot \left(\frac{3}{4(1-\gamma)} - \sqrt{\E[(\overline{V}_{\tau'}(2))^2]}\right) \nonumber \\
        & \geq \frac{1}{2} \cdot \left(\frac{3}{4(1-\gamma)} - \frac{1}{2(1-\gamma)}\right) \geq \frac{1}{8(1-\gamma)}, \label{eqn:v_bound_tau_prime_to_T_more_than_half}
    \end{align}
    where the first line follows from~\eqref{eqn:bar_v_expected_err}, the second line from the condition on step sizes and the third line from the definition of $\tau'$.
    \item We now consider the other case where,
    \begin{align}
        0\leq \prod_{k = \tau' + 1}^T (1-\eta_k(1-\gamma p)) < \frac{1}{2}. \label{eqn:tau_prime_to_T_less_than_half}
    \end{align}
    Using~\citep[Eqn.(134)]{Li2023QLMinimax}, for any $t' < t$ and all agents $m$, we have the relation
    \begin{align*}
        V_t^m(2) = \frac{1}{1-\gamma p} - \prod_{k = t'+1}^t (1-\eta_k(1-\gamma p)) \left(\frac{1}{1-\gamma p} - V_{t'}^m(2)\right) + \sum_{k = t' + 1} \eta_k^{(t)} \gamma (\hat{P}_k^m(2|2) - p) V_{k-1}^m(2).
    \end{align*}
    The above equation is directly obtained by unrolling the recursion in~\eqref{eqn:state_2_update} along with noting that $Q_t(2,1) = V_t(2)$ for all $t$. Consequently, we have, 
    \begin{align}
        \overline{V}_T(2) = \frac{1}{1-\gamma p} - \prod_{k = t'+1}^T (1-\eta_k(1-\gamma p)) \left(\frac{1}{1-\gamma p} - \overline{V}_{t'}(2)\right) + \frac{1}{M}\sum_{m = 1}^M\sum_{k = t' + 1}^T \eta_k^{(T)} \gamma (\hat{P}_k^m(2|2) - p) V_{k-1}^m(2).
    \end{align}
    Let $\{\sF_t\}_{t = 0}^T$ be a filtration such that $\sF_t$ is the $\sigma$-algebra corresponding to $\{\{\hat{P}^m_s(2|2)\}_{m = 1}^M\}_{s = 1}^t$. It is straightforward to note that $\left\{\frac{1}{M}\sum_{m = 1}^M\eta_k^{(T)} \gamma (\hat{P}_k^m(2|2) - p) V_{k-1}^m(2)\right\}_k$ is a martingale sequence adapted to the filtration $\sF_k$. Thus, using the result from~\citep[Eqn.(139)]{Li2023QLMinimax}, we can conclude that
    \begin{align}
        \var(\overline{V}_T(2)) \geq \E\left[ \sum_{k = \tau'+ 2}^T \var\left( \frac{1}{M}\sum_{m = 1}^M\eta_k^{(T)} \gamma (\hat{P}_k^m(2|2) - p) V_{k-1}^m(2) \ \bigg| \sF_{k-1} \right)\right].\label{eqn:variance_v_bar_T_formula}
    \end{align}
    We have, 
    \begin{align}
        \var\left( \frac{1}{M}\sum_{m = 1}^M\eta_k^{(T)} \gamma (\hat{P}_k^m(2|2) - p) V_{k-1}^m(2) \ \bigg| \sF_{k-1} \right) & = \frac{1}{M^2} \sum_{m = 1}^M \var\left(\eta_k^{(T)} \gamma (\hat{P}_k^m(2|2) - p) V_{k-1}^m(2) \ \bigg| \sF_{k-1}\right)\nonumber \\
        & =  \frac{(\eta_k^{(T)})^2}{BM} \gamma^2 p(1-p) \left(\frac{1}{M} \sum_{m = 1}^M(V_{k-1}^m(2))^2 \right)\nonumber \\
        & \geq  \frac{(1-\gamma)(4\gamma - 1)}{9BM} \cdot (\eta_k^{(T)})^2 \cdot   (\overline{V}_{k-1}(2))^2, \label{eqn:variance_v_bar_martingale_terms}
    \end{align}
    where the first line follows from that fact that variance of sum of i.i.d. random variables is the sum of their variances, the second line from variance of Binomial random variable and the third line from Jensen's inequality. Thus,~\eqref{eqn:variance_v_bar_T_formula} and~\eqref{eqn:variance_v_bar_martingale_terms} together yield,
    \begin{align}
        \var(\overline{V}_T(2)) \geq  \frac{(1-\gamma)(4\gamma - 1)}{9BM} \cdot \sum_{k = \tau'+ 2}^T  (\eta_k^{(T)})^2 \cdot   \E[(\overline{V}_{k-1}(2))^2] \nonumber \\
        \geq  \frac{(1-\gamma)(4\gamma - 1)}{9BM} \cdot   \frac{1}{4(1-\gamma)^2} \cdot \sum_{k = \max\{\tau, \tau'\}+ 2}^T  (\eta_k^{(T)})^2,\label{eqn:variance_v_bar_T_bound}
    \end{align}
    where the second line follows from the definition of $\tau'$. We focus on bounding the third term in the above relation. We have,
    \begin{align}
    \sum_{k = \max\{\tau', \tau\} + 2}^T \left(\eta_{k}^{(T)}\right)^2 & \geq \sum_{k = \max\{\tau', \tau\} + 2}^T \left( \eta_k \prod_{i = k+1}^T (1- \eta_i(1-\gamma p) \right)^2 \nonumber  \\
    & \geq \sum_{k = \max\{\tau', \tau\} + 2}^T \left( \eta_T \prod_{i = k+1}^t (1- \eta_{\tau}(1-\gamma p)) \right)^2 \nonumber  \\
    & = \eta_T^2 \sum_{k = \max\{\tau', \tau\} + 2}^T (1- \eta_{\tau}(1-\gamma p))^{2(t-k)} \nonumber  \\
    & \geq \eta_T^2 \cdot \frac{1 - (1-\eta_{\tau}(1-\gamma p))^{2(T-\max\{\tau', \tau\}- 1)}}{\eta_{\tau}(1 -\gamma p)(2 - \eta_{\tau}(1-\gamma p))}  \nonumber \\
    & \geq \eta_T \cdot \frac{1}{4(1 - \gamma)} \cdot c',
    \label{eqn:eta_k_t_square_sum_lower_bound_variance_v_2}
\end{align}
where the second line follows from monotonicity of $\eta_t$ and the numerical constant $c'$ in the fifth step is given by the following claim whose proof is deferred to the end of the section: 
\begin{align}
    1 - (1-\eta_{\tau}(1-\gamma p))^{2(T-\max\{\tau', \tau\}- 1)} \geq \begin{cases} 1- e^{-8/9} & \text{ for constant step sizes}, \\ 1 - \exp\left(-\frac{8}{3\max\{1,c_{\eta}\}}\right) & \text{ for linearly rescaled step sizes} \end{cases} . \label{eqn:numerical_constant_variance_v_2}
\end{align}
Thus,~\eqref{eqn:variance_v_bar_T_bound} and~\eqref{eqn:eta_k_t_square_sum_lower_bound_variance_v_2} together imply
\begin{align}
        \var(\overline{V}_T(2)) & \geq  \frac{(4\gamma - 1)}{36BM(1-\gamma)} \cdot \sum_{k = \tau'+ 2}^T  (\eta_k^{(T)})^2 \nonumber \\
        &\geq  \frac{c'(4\gamma - 1)}{144(1-\gamma)} \cdot   \frac{\eta_T}{BM(1- \gamma)} \geq  \frac{c'(4\gamma - 1)}{144(1-\gamma)} \cdot   \frac{1}{100}, \label{eqn:v_bound_tau_prime_to_T_less_than_half}
    \end{align}
    where the last inequality follows from the bound on $\frac{\eta_T}{BM}$. 
\end{itemize}
Thus, for all $N \geq 1$, we have,
\begin{align*}
    \E[(V^{\star}(2) - \overline{V}_T(2))^2] = \E[(V^{\star}(2) - \E[\overline{V}_T(2)])^2] + \var(\overline{V}_T(2)) \geq \frac{c''}{(1-\gamma)N},
\end{align*}
for some numerical constant $c''$. Similar to the small learning rate case, the error rate is bounded away from a constant value irrespective of the number of agents and the number of communication rounds. Thus, even with $\CC_{\textsf{round}} = \Omega(T)$, we will not observe any collaborative gain in this scenario.

\paragraph{Proof of~\eqref{eqn:numerical_constant_variance_v_2}.} To establish the claim, we consider two cases:
\begin{itemize}
    \item $\tau' \geq \tau$: Under this case, we have,
    \begin{align}
        (1-\eta_{\tau}(1-\gamma p))^{2(T-\max\{\tau', \tau\}- 1)}  & = (1-\eta_{\tau}(1-\gamma p))^{2(T-\tau'- 1)} \nonumber \\
        & \leq (1-\eta_{\tau}(1-\gamma p))^{T-\tau'} \leq \prod_{k = \tau' + 1}^T (1-\eta_{k}(1-\gamma p)) \leq \frac{1}{2}, \label{eqn:tau_prime_greater_than_tau}
    \end{align}
    where the last inequality follows from~\eqref{eqn:tau_prime_to_T_less_than_half}.
    \item $\tau \geq \tau'$: For this case, we have
    \begin{align}
        (1-\eta_{\tau}(1-\gamma p))^{2(T-\max\{\tau', \tau\}- 1)}  &= (1-\eta_{\tau}(1-\gamma p))^{2(T-\tau- 1)} \nonumber \\
        & \leq (1-\eta_{\tau}(1-\gamma p))^{T-\tau} \leq \exp\left(-\frac{2T\eta_{\tau}(1-\gamma p)}{3}\right). \label{eqn:tau_prime_less_than_tau}
    \end{align}
    For the constant stepsize schedule, we have,
    \begin{align}
        \exp\left(-\frac{2T\eta_{\tau}(1-\gamma p)}{3}\right) \leq \exp\left(-\frac{2T}{3} \cdot \frac{1}{(1-\gamma)T} \cdot \frac{4(1-\gamma)}{3}\right) = \exp\left(-\frac{8}{9}\right) \label{eqn:constant_step_size_bound_variance_v_2}
    \end{align}
    For linearly rescaled stepsize schedule, we have,
    \begin{align}
        \exp\left(-\frac{2T\eta_{\tau}(1-\gamma p)}{3}\right) \leq \exp\left(-\frac{2T}{3} \cdot \frac{1}{1+ c_{\eta}(1-\gamma)T/3} \cdot \frac{4(1-\gamma)}{3}\right) = \exp\left(-\frac{8}{3\max\{1,c_{\eta}\}}\right) \label{eqn:rescaled_step_size_bound_variance_v_2}
    \end{align}
    On combining~\eqref{eqn:tau_prime_greater_than_tau},~\eqref{eqn:tau_prime_less_than_tau},~\eqref{eqn:constant_step_size_bound_variance_v_2} and~\eqref{eqn:rescaled_step_size_bound_variance_v_2}, we arrive at the claim.
\end{itemize}

%% file: auxiliary_lower_bound.tex
\subsection{Proofs of auxiliary lemmas}
\label{sec:aux_lemmas_proof_lower_bound}

\subsubsection{Proof of Lemma~\ref{lemma:Q_hatQ_difference}}

Note that a similar relationship is also derived in \cite{Li2023QLMinimax}, but needing to take care of the averaging over multiple agents, we present the entire arguments for completeness.
We prove the claim using an induction over $t$. It is straightforward to note that the claim is true for $t = 0$ and all agents $m \in \{1,2,\dots, M\}$. For the inductive step, we assume that the claim holds for $t - 1$ for all clients. Using the induction hypothesis, we have the following relation between $V_{t-1}^m(1)$ and $\hatV_{t-1}^m$:
\begin{align}
    V_{t-1}^m(1) = \max_{a \in \{1,2\}} Q_{t-1}^m(1,a) \geq \max_{a \in \{1,2\}} \hatQ_{t-1}^m(a) - \frac{1}{1-\gamma} \prod_{i = 1}^{t-1}(1 - \eta_{i}(1-\gamma)) = \hatV_{t-1}^m - \frac{1}{1-\gamma} \prod_{i = 1}^{t-1}(1 - \eta_{i}(1-\gamma)). \label{eqn:V_hatV_relation}
\end{align}
For $t \notin \{t_r\}_{r = 1}^R$ and $a \in \{1,2\}$, we have,
\begin{align}
    Q_{t}^m(1,a) - \hatQ_{t}^m(a) & = Q_{t-1/2}^m(1,a) - \hatQ_{t - 1/2}^m(a) \nonumber \\
    & = (1- \eta_t)Q_{t-1}^m(1,a) + \eta_t(1 + \gamma \hatP_{t}^m(1|1,a) V_{t-1}^m(1)) \nonumber\\
    & ~~~~~~~~~~~~~~~~~~~~~~~  - \left[(1- \eta_t)\hatQ_{t-1}^m(a) + \eta_t(1 + \gamma \hatP_{t}^m(1|1,a) \hatV_{t-1}^m)\right] \nonumber \\
    & = (1- \eta_t)(Q_{t-1}^m(1|1,a) - \hatQ_{t-1}^m(a)) + \eta_t \gamma \hatP_{t}^m(1|1,a) (V_{t-1}^m(1) - \hatV_{t-1}^m) \nonumber \\
    & \geq - \frac{(1- \eta_t)}{1-\gamma} \prod_{i = 1}^{t-1}(1 - \eta_{i}(1-\gamma)) - \hatP_{t}^m(1|1,a) \cdot \frac{\eta_t \gamma }{1-\gamma} \prod_{i = 1}^{t-1}(1 - \eta_{i}(1-\gamma)) \nonumber \\
    & \geq - \frac{(1- \eta_t)}{1-\gamma} \prod_{i = 1}^{t-1}(1 - \eta_{i}(1-\gamma))  -  \frac{\eta_t \gamma}{1-\gamma} \prod_{i = 1}^{t-1}(1 - \eta_{i}(1-\gamma))  \nonumber \\
    & \geq - \frac{1}{1-\gamma} \prod_{i = 1}^{t}(1 - \eta_{i}(1-\gamma)). \label{eqn:Q_hatQ_diff_case_2}
\end{align}
For $t \in \{t_r\}_{r = 1}^R$ and $a \in \{1,2\}$, we have,
\begin{align}
    Q_{t}^m(1,a) - \hatQ_{t}^m(a) & = \frac{1}{M} \sum_{m = 1}^M Q_{t-1/2}^m(1,a) - \frac{1}{M} \sum_{m = 1}^M \hatQ_{t - 1/2}^m(a) \nonumber \\
    & = \frac{1}{M} \sum_{m = 1}^M \left[(1- \eta_t)Q_{t-1}^m(1,a) + \eta_t(1 + \gamma \hatP_{t}^m(1|1,a) V_{t-1}^m(1))\right] \nonumber \\
    & ~~~~~~~~~~~~~~~~~ - \frac{1}{M} \sum_{m = 1}^M \left[(1- \eta_t)\hatQ_{t-1}^m(a) + \eta_t(1 + \gamma \hatP_{t}^m(1|1,a) \hatV_{t-1}^m)\right] \nonumber \\
    & = \frac{1}{M} \sum_{m = 1}^M \left[(1- \eta_t)(Q_{t-1}^m(1,a) - \hatQ_{t-1}^m(a)) + \eta_t \gamma \hatP_{t}^m(1|1,a) (V_{t-1}^m(1) - \hatV_{t-1}^m)\right] \nonumber \\
    & \geq - \frac{1}{1-\gamma} \prod_{i = 1}^{t}(1 - \eta_{i}(1-\gamma)), \label{eqn:Q_hatQ_diff_case_3}
\end{align}
where the last step follows using the same set of arguments as used in \eqref{eqn:Q_hatQ_diff_case_2}. The inductive step follows from \eqref{eqn:Q_hatQ_diff_case_2} and~\eqref{eqn:Q_hatQ_diff_case_3}.


\subsubsection{Proof of Lemma~\ref{lemma:delta_t_max_minus_xi_tt_max}}

In order to bound the term $\E[\Delta_{t, \max}^m] - \E[\xi_{t',t, \max}^m]$, we make use of the relation in \eqref{eqn:delta_max_unaveraged_expectation_lower_bound}, which we recall
\begin{align*}
    \E[\Delta_{t, \max}^m] & \geq \varphi_{t', t}  \E[\barDelta_{t', \max}] +  \left[\sum_{k = t' + 1}^t  \widetilde{\eta}_k^{(t)}\gamma  p\E[\Delta_{k-1, \max}^m]  \right] + \E[\xi_{t',t, \max}^m] - \varphi_{t', t} \E[|\barDelta_{t'}(1) - \barDelta_{t'}(2)|].
\end{align*}
\begin{itemize}
\item To aid the analysis, we consider the following recursive relation for any fixed agent $m$:
\begin{align}
    y_t = (1- \eta_t)y_{t-1} + \eta_t(\gamma p y_{t-1} + \E[\xi_{t',t, \max}^m]). \label{eqn:y_recursion}
\end{align}
Upon unrolling the recursion, we obtain,
\begin{align}
    y_t & = \left(\prod_{k = t' +1}^t (1- \eta_k) \right) y_{t'}  +  \sum_{k = t' + 1}^t \left( \eta_k  \prod_{i = k+1}^t (1- \eta_i) \right) \gamma p y_{k-1}  + \sum_{k = t' + 1}^t \left( \eta_k  \prod_{i = k+1}^t (1- \eta_i) \right) \E[\xi_{t',t, \max}^m] \nonumber \\
    & = \varphi_{t', t} y_{t'}  +  \sum_{k = t' + 1}^t \widetilde{\eta}_k^{(t)} \gamma p y_{k-1}  + \sum_{k = t' + 1}^t \widetilde{\eta}_k^{(t)} \E[\xi_{t',t, \max}^m].\label{eqn:y_recursion_unrolled}
\end{align}
Initializing $y_{t'} = \E[\barDelta_{t', \max}]$ in \eqref{eqn:y_recursion_unrolled} and plugging this into \eqref{eqn:delta_max_unaveraged_expectation_lower_bound}, we have
$$\E[\Delta_{t, \max}^m] \geq y_t - \varphi_{t', t} \E[|\barDelta_{t'}(1) - \barDelta_{t'}(2)|],$$ 
where we used $\sum_{k = t' + 1}^t \widetilde{\eta}_k^{(t)}  \leq 1$ (cf.~\eqref{eqn:tilde_eta_k_t_recursion}). We now further simply the expression of $y_t$. By rewriting \eqref{eqn:y_recursion} as
\begin{align*}
    y_t = (1- \eta_t(1-\gamma p))y_{t-1} + \eta_t \E[\xi_{t',t, \max}^m],
\end{align*}
it is straight forward to note that $y_t$ is given as
\begin{align}
    y_t =  \left(\prod_{k = t' + 1}^t (1 - \eta_k(1-\gamma p))\right) y_{t'} + \E[\xi_{t',t, \max}^m] \left[\sum_{k = t' + 1}^t  \eta_k^{(t)} \right]. \label{eqn:y_recursion_solved}
\end{align}
Consequently, we have,
\begin{align}
    \E[\Delta_{t, \max}^m] - \E[\xi_{t',t, \max}^m] & \geq  \left(\prod_{k = t' + 1}^t (1 - \eta_k(1-\gamma p))\right) \E[\barDelta_{t', \max}] + \E[\xi_{t',t, \max}^m] \left[\sum_{k = t' + 1}^t \eta_k^{(t)} - 1  \right] \nonumber \\
    & ~~~~~~~~~~~~~~~~~~~~~~~~~~~~~~ - \varphi_{t', t} \E[|\barDelta_{t'}(1) - \barDelta_{t'}(2)|]. \label{eqn:delta_minus_xi_eqn_1}
\end{align}

\item We can consider a slightly different recursive sequence defined as
\begin{align}
    w_t = (1- \eta_t)w_{t-1} + \eta_t(\gamma p  w_{t-1}). \label{eqn:w_recursion}
\end{align}
Using a similar sequence of arguments as outlined in \eqref{eqn:y_recursion}-\eqref{eqn:y_recursion_solved}, we can conclude that if $w_{t'} = \E[\barDelta_{t', \max}]$, then $\E[\Delta_{t, \max}^m] \geq w_t + \E[\xi_{t',t, \max}^m] -  \varphi_{t', t} \E[|\barDelta_{t'}(1) - \barDelta_{t'}(2)|]$ and consequently,
\begin{align}
    \E[\Delta_{t, \max}^m] & \geq \left(\prod_{k = t' + 1}^t (1 - \eta_k(1-\gamma p))\right) \E[\barDelta_{t', \max}] + \E[\xi_{t',t, \max}^m] - \varphi_{t', t} \E[|\barDelta_{t'}(1) - \barDelta_{t'}(2)|]. \label{eqn:delta_minus_xi_eqn_2}
\end{align}
\end{itemize}
On combining \eqref{eqn:delta_minus_xi_eqn_1} and~\eqref{eqn:delta_minus_xi_eqn_2}, we arrive at the claim.

\subsubsection{Proof of Lemma~\ref{lemma:xi_lower_bound}}

We begin with bounding the first term $\E[\xi_{t',t, \max}^m]$; the second bound follows in an almost identical derivation. 

\paragraph{Step 1: applying Freedman's inequality.} Using the relation $\max\{a,b\} = \frac{a + b + |a - b|}{2}$, we can rewrite $\E[\xi_{t',t,\max}^m]$ as 
\begin{align}
    \E[\xi_{t',t, \max}^m] & = \E\left[ \frac{\xi_{t',t}^m(1) + \xi_{t',t}^m(2)}{2} + \left| \frac{\xi_{t',t}^m(1) - \xi_{t',t}^m(2)}{2} \right| \right] \nonumber \\
    & = \frac{1}{2} \E\left[\left| \frac{\xi_{t',t}^m(1) - \xi_{t',t}^m(2)}{2} \right| \right] \nonumber \\
    & = \frac{1}{2} \E\Bigg[\Bigg| \underbrace{ \sum_{k = t' + 1}^t \widetilde{\eta}_{k}^{(t)} \gamma (\hatP_k^m(1|1, 1) - \hatP_k^m(1|1, 2))\hatV_{k-1}^m}_{=:   \zeta_{t',t}^m } \Bigg| \Bigg], \label{eq:apricot}
\end{align}
where we used the definition in \eqref{eqn:xi_def} and the fact that $\E[\xi_{t',t}^m(1)] = \E[\xi_{t',t}^m(2)] = 0$. Decompose $   \zeta_{t',t}^m$ as
\begin{align}
    \zeta_{t',t}^m     & = \sum_{k = t' + 1}^t \sum_{b = 1}^{B} \widetilde{\eta}_{k}^{(t)} \frac{\gamma}{B} (P_{k,b}^m(1|1, 1) - P_{k,b}^m(1|1, 2))\hatV_{k-1}^m  =: \sum_{l = 1}^L z_l, \label{eqn:zeta_def}
\end{align}
where for all $1 \leq l \leq L$
\begin{align*}
    z_l := \frac{\gamma}{B} (P_{k(l),b(l)}^m(1|1, 1) - P_{k(l),b(l)}^m(1|1, 2))\hatV_{k(l) - 1}^m 
\end{align*}
with 
\begin{align*}
    k(l) := \lfloor l/B \rfloor + t' + 1; \quad b(l) = ((l-1) \mod B) + 1; \quad L = (t - t')B.
\end{align*}
Let $\{\sF_l\}_{l = 1}^{L}$ be a filtration such that $\sF_l$ is the $\sigma$-algebra corresponding to $\{P_{k(j),b(j)}^m(1|1, 1), P_{k(j),b(j)}^m(1|1, 2)\}_{j = 1}^l$. It is straightforward to note that $\{z_l\}_{l = 1}^L$ is a martingale sequence adapted to the filtration $\{\sF\}_{l =1}^L$. We will use the Freedman's inequality~\citep{Freedman1975Martingale, Li2023QLMinimax} to obtain a high probability bound on $|\zeta_{t',t}^m|$. 
\begin{itemize}
\item To that effect, note that
\begin{align}
    \sup_{l} |z_l| & \leq \sup_{l} \left|\widetilde{\eta}_{k(l)}^{(t)} \cdot  \frac{\gamma}{B} \cdot (P_{k(l),b(l)}^m(1|1, 1) - P_{k(l),b(l)}^m(1|1, 2)) \cdot \hatV_{k(l)-1}^m \right| \nonumber \\
    & \leq \widetilde{\eta}_{k(l)}^{(t)} \cdot \frac{\gamma}{B(1-\gamma)} \nonumber \\
    & \leq \frac{\eta_t}{B(1 - \gamma)}, \label{eqn:z_l_abs_bound}
\end{align}
where the second step follows from the bounds $|(P_{k(l),b(l)}^m(1|1, 1) - P_{k(l),b(l)}^m(1|1, 2))| \leq 1$ and $\hatV_{k(l)-1}^m  \leq \frac{1}{1-\gamma}$ and the third step uses $c_{\eta}   \leq \frac{1}{1-\gamma}$ and the fact that $\widetilde{\eta}_{k}^{(T)}$ is increasing in $k$ in this regime. (cf.~\eqref{eqn:tilde_eta_k_t_increasing}). 
\item Similarly, 
\begin{align}
    \var(z_l| \sF_{l-1}) & \leq \left(\widetilde{\eta}_{k(l)}^{(t)}\right)^2 \frac{\gamma^2}{B^2}\cdot \left( \hatV_{k(l)-1}^m \right)^2 \cdot \var(P_{k(l),b(l)}^m(1|1, 1) - P_{k(l),b(l)}^m(1|1, 2))  \nonumber \\
    & \leq \left(\widetilde{\eta}_{k(l)}^{(t)}\right)^2 \frac{\gamma^2}{B^2(1-\gamma)^2} \cdot 2p(1-p)  \nonumber \\
    & \leq  \frac{2\left(\widetilde{\eta}_{k(l)}^{(t)}\right)^2}{3B^2(1-\gamma)}. \label{eqn:z_l_var_bound}
\end{align}
\end{itemize}
Using the above bounds~\eqref{eqn:z_l_abs_bound} and~\eqref{eqn:z_l_var_bound} along with Freedman's inequality yield that 
\begin{align}
    \Pr \left( |\zeta_{t',t}^m| \geq \sqrt{ \frac{8 \log(2/\delta)}{3B^2(1-\gamma)} \sum_{l = 1}^L \left(\widetilde{\eta}_{k(l)}^{(t)}\right)^2 } + \frac{4\eta_t \log(2/\delta)}{3B(1 - \gamma)}\right) \leq \delta .
\end{align}
Setting $\delta_0 = \frac{(1-\gamma)^2}{2} \cdot \E[|\zeta_{t',t}^m|^2]$, with probability at least $1-\delta_0$, it holds
\begin{align} \label{eqn:D_def} 
  |\zeta_{t',t}^m| \geq \sqrt{ \frac{8 \log(2/\delta_0)}{3B(1-\gamma)} \sum_{k = t' + 1}^t \left(\widetilde{\eta}_{k}^{(t)}\right)^2 } + \frac{4\eta_t \log(2/\delta_0)}{3B(1 - \gamma)} =: D . 
\end{align}

Consequently, plugging this back to \eqref{eq:apricot}, we obtain 
\begin{align}
    \E[\xi_{t',t, \max}^m] & = \frac{1}{2} \E[|\zeta_{t',t}^m|] \nonumber \\
    & \geq  \frac{1}{2} \E[|\zeta_{t',t}^m| \1\{ |\zeta_{t',t}^m| \leq D\}] \nonumber \\
    & \geq  \frac{1}{2D} \E[|\zeta_{t',t}^m|^2 \1\{ |\zeta_{t',t}^m| \leq D\}] \nonumber \\
    & \geq  \frac{1}{2D} \left(\E[|\zeta_{t',t}^m|^2] - \E[|\zeta_{t',t}^m|^2 \1\{ |\zeta_{t',t}^m| > D\}] \right) \nonumber \\
    & \geq  \frac{1}{2D} \left(\E[|\zeta_{t',t}^m|^2] - \frac{\Pr( |\zeta_{t',t}^m| > D)}{(1-\gamma)^2} \right)   \geq  \frac{1}{4D} \cdot \E[|\zeta_{t',t}^m|^2]. \label{eqn:xi_in_terms_of_zeta}
\end{align}
Here, the penultimate step  used the fact that $\displaystyle |\zeta_{t',t}^m| \leq \sum_{k = t' + 1}^t \frac{\widetilde{\eta}_{k}^{(t)}}{(1-\gamma)} \leq \frac{1}{(1-\gamma)}$, and the last step used the definition of $\delta_0$. Thus, it is sufficient to obtain a lower bound on $\E[|\zeta_{t',t}^m|^2]$ in order obtain a lower bound for $\E[\xi_{t',t, \max}^m]$. 

\paragraph{Step 2: lower bounding $\E[|\zeta_{t',t}^m|^2]$.} To proceed, we introduce the following lemma pertaining to lower bounding $\hatV_{t}^m$ that will be useful later.
\begin{lemma}
    For all time instants $t \in [T]$ and all agent $m \in [M]$:
    \begin{align*}
        \E\left[ \left( \hatV_{t}^m \right)^2 \right] \geq \frac{1}{2(1 -\gamma)^2}.
    \end{align*}
    \label{lemma:V_t_lower_bound}
\end{lemma}
We have,
\begin{align}
    \E[|\zeta_{t',t}^m|^2] & = \E\left[\sum_{l = 1}^L \var\left(z_l | \sF_{l-1}\right)\right] = \E\left[\sum_{l = 1}^L \E\left[z_l^2 | \sF_{l-1}\right]\right] \nonumber \\
    & \geq  \sum_{l = 1}^L \left(\widetilde{\eta}_{k(l)}^{(t)}\right)^2 \frac{\gamma^2}{B^2}\cdot 2p(1-p)  \cdot \E\left[\left( \hat{V}_{k(l)-1}^m\right)^2  \right] \nonumber \\
    & \geq  \sum_{l = 1}^L \left(\widetilde{\eta}_{k(l)}^{(t)}\right)^2 \frac{\gamma^2}{B^2}\cdot 2p(1-p)  \cdot \frac{1}{2(1-\gamma)^2}\nonumber \\
     & \geq \frac{2}{9B(1-\gamma)} \cdot \sum_{k = \max\{t', \tau\} + 1}^t \left(\widetilde{\eta}_{k}^{(t)}\right)^2, \label{eqn:zeta_sq_lower_bound}
\end{align}
where the third line follows from Lemma~\ref{lemma:V_t_lower_bound} and the fourth line uses $\gamma \geq 5/6$.

\paragraph{Step 3: finishing up.} We finish up the proof by bounding $\sum_{k = \max\{t', \tau\} + 1}^t \left(\widetilde{\eta}_{k}^{(t)}\right)^2$ for $t - \max\{t', \tau\} \geq 1/\eta_{\tau}$. We have
\begin{align}
    \sum_{k = \max\{t', \tau\} + 1}^t \left(\widetilde{\eta}_{k}^{(t)}\right)^2 & \geq \sum_{k = \max\{t', \tau\} + 1}^t \left( \eta_k \prod_{i = k+1}^t (1- \eta_i) \right)^2 \nonumber  \\
    & \overset{\mathrm{(i)}}{\geq} \sum_{k = \max\{t', \tau\} + 1}^t \left( \eta_t \prod_{i = k+1}^t (1- \eta_{\tau}) \right)^2 \nonumber  \\
    & = \eta_t^2 \sum_{k = \max\{t', \tau\} + 1}^t (1- \eta_{\tau})^{2(t-k)} \nonumber  \\
    & \geq \eta_t^2 \cdot \frac{1 - (1-\eta_{\tau})^{2(t-\max\{t', \tau\})}}{\eta_{\tau}(2 - \eta_{\tau})}  \nonumber \\
    & \geq \eta_t \cdot \frac{1 - \exp(-2)}{6}  \geq \frac{\eta_t}{10} \geq \frac{\eta_T}{10},
    \label{eqn:lambda_square_sum_lower_bound}
\end{align}
where (i) follows from the monotonicity of $\eta_k$. Plugging \eqref{eqn:lambda_square_sum_lower_bound} into the expressions of $D$ (cf.~\eqref{eqn:D_def}) we have
\begin{align*}
    D = & \sqrt{ \frac{8 \log(2/\delta_0)}{3B(1-\gamma)} \sum_{k = t' + 1}^t \left(\widetilde{\eta}_{k}^{(t)}\right)^2 } + \frac{4\eta_t \log(2/\delta_0)}{3B(1 - \gamma)}  \\
    & \leq \frac{9}{2} \E[|\zeta_{t',t}^m|^2] \cdot \sqrt{\frac{8 \log(2/\delta_0)}{3}} \left( \frac{1}{B(1-\gamma)} \sum_{k = t' + 1}^t \left(\widetilde{\eta}_{k}^{(t)}\right)^2 \right)^{-1/2} + 60 \cdot \E[|\zeta_{t',t}^m|^2] \cdot \log(2/\delta_0) \\
    & \leq 3 \E[|\zeta_{t',t}^m|^2] \cdot \log(2/\delta_0) \left[\sqrt{\frac{60B(1-\gamma)}{\eta_t}}  + 20 \right] \\
    & \leq 60 \E[|\zeta_{t',t}^m|^2] \cdot \log(2/\delta_0) \left[\sqrt{\frac{3B(1-\gamma)}{20\eta_T}}  + 1 \right] ,
\end{align*}
where the second line follows from \eqref{eqn:zeta_sq_lower_bound} and \eqref{eqn:lambda_square_sum_lower_bound}, and the third line follows from \eqref{eqn:lambda_square_sum_lower_bound}.
On combining the above bound with \eqref{eqn:xi_in_terms_of_zeta}, we obtain,
\begin{align}
    \E[\xi_{t',t, \max}^m] \geq \frac{1}{240 \log(2/\delta_0)} \cdot \frac{\nu}{\nu + 1},  \label{eqn:xi_lower_bound_penultimate_step} 
\end{align}
where $\nu := \sqrt{ \dfrac{20\eta_T}{3B(1-\gamma)}}$. Note that we have,
\begin{align*}
    \delta_0 & = \frac{ (1-\gamma)^2}{2} \cdot \E[|\zeta_{t',t}^m|^2]  \geq \frac{(1 - \gamma)}{9B} \cdot \sum_{k = t' + 1}^t \left(\widetilde{\eta}_{k}^{(t)}\right)^2  \geq \frac{\eta_T(1 - \gamma)}{90 B}.
\end{align*}
Combining the above bound with \eqref{eqn:xi_lower_bound_penultimate_step} yields us the required bound.
 
\paragraph{Step 4: repeating the argument for the second claim.}
 We note that second claim in the theorem, i.e., the lower bound on $\E\left[\max\left\{ \frac{1}{M}\sum_{m = 1}^M \xi_{t',t}^m(1), \frac{1}{M}\sum_{m = 1}^M \xi_{t',t}^m(2) \right\}\right]$ follows through an identical series of arguments where the bounds in 
 Eqns.~\eqref{eqn:z_l_abs_bound} and~\eqref{eqn:z_l_var_bound} contain an additional factor of $M$ in the denominator (effectively replacing $B$ with $BM$), which is carried through in all the following steps.

\subsubsection{Proof of Lemma~\ref{lemma:barDelta_action_diff}}

Using Eqns.~\eqref{eqn:delta_m_averaged} and~\eqref{eqn:xi_def}, we can write
\begin{align*}
    \barDelta_t(1) - \barDelta_t(2) & = \left(\prod_{k = t' +1}^t (1- \eta_k)\right) (\barDelta_{t'}(1) - \barDelta_{t'}(2)) \\
    & ~~~~~~~~~ + \frac{1}{M} \sum_{m = 1}^M \sum_{k = t' + 1}^t \left( \eta_k \prod_{i = k+1}^t (1- \eta_i) \right) \gamma (\hatP_k^m(1|1, 1) - \hatP_k^m(1|1, 2))\hatV_{k-1}^m.
\end{align*}
Upon unrolling the recursion, we obtain,
\begin{align*}
    \barDelta_t(1) - \barDelta_t(2) =  \sum_{k = 1}^t  \sum_{m = 1}^M  \left( \eta_k \prod_{i = k+1}^t (1- \eta_i) \right) \frac{\gamma}{M}  (\hatP_k^m(1|1, 1) - \hatP_k^m(1|1, 2))\hatV_{k-1}^m.
\end{align*}
If we define a filtration $\sF_k$ as the $\sigma$-algebra corresponding to $\{\hatP_l^1(1|1, 1), \hatP_l^1(1|1, 2), \dots, \hatP_l^M(1|1, 1), \hatP_l^M(1|1, 2)\}_{l = 1}^k$, then it is straightforward to note that $\{\barDelta_t(1) - \barDelta_t(2)\}_{t}$ is a martingale sequence adapted to the filtration $\{\sF_t\}_{t}$. Using Jensen's inequality, we know that if $\{Z_t\}_t$ is a martingale adapted to a filtration $\{\sG_t\}_t$, then for a convex function $f$ such that $f(Z_t)$ is integrable for all $t$, $\{f(Z_t)\}_t$ is a sub-martingale adapted to $\{\sG_t\}_t$. Since $f(x) = |x|$ is a convex function, $\{|\barDelta_t(1) - \barDelta_t(2)|\}_{t}$ is a submartingale adapted to the filtration $\{\sF_t\}_{t}$.  As a result, 
\begin{align}
    \sup_{1 \leq t \leq T} \E[|\barDelta_t(1) - \barDelta_t(2)|] & \leq \E[|\barDelta_T(1) - \barDelta_T(2)|]  \leq \left(\E[(\barDelta_T(1) - \barDelta_T(2))^2] \right)^{1/2}. \label{eqn:sub_martingale} 
\end{align}

We use the following observation about a martingale sequence $\{X_i\}_{i = 1}^{t}$ adapted to a filtration $\{\sG_i\}_{i = 1}^{t}$ to evaluate the above expression. We have,
\begin{align}
    \E\left[ \left(\sum_{i = 1}^{t} X_i\right)^2 \right] & = \E\left[ \E\left[  \left(\sum_{i = 1}^{t} X_i\right)^2 \bigg| \sG_{t-1}  \right] \right] \nonumber \\
    & = \E\left[ \E\left[  X_t^2 + 2X_t \left(\sum_{i = 1}^{t-1} X_i\right) + \left(\sum_{i = 1}^{t-1} X_i\right)^2 \bigg| \sG_{t-1}  \right] \right] \nonumber \\
    & = \E\left[ X_t^2 \right] + \E\left[  \left(\sum_{i = 1}^{t-1} X_i\right)^2   \right] \nonumber \\
    & = \sum_{i = 1}^t \E\left[ X_i^2 \right], \label{eqn:martingale_sum_of_squares}
\end{align}
where the third step uses the facts that $\left(\sum_{i = 1}^{t-1} X_i\right)$ is $\sG_{t-1}$ measure and $\E[X_t|\sG_{t-1}] = 0$ and fourth step is obtained by recursively applying second and third steps. Using the relation in Eqn.~\eqref{eqn:martingale_sum_of_squares} in Eqn.~\eqref{eqn:sub_martingale}, we obtain,
\begin{align}
    \sup_{1 \leq t \leq T} \E[|\barDelta_t(1) - \barDelta_t(2)|] & \leq \left(\E[(\barDelta_T(1) - \barDelta_T(2))^2] \right)^{1/2} \nonumber  \\
    & \leq \left(\sum_{k = 1}^T \E\left[\left( \sum_{m = 1}^M  \widetilde{\eta}_k^{(T)} \cdot \frac{\gamma}{M} \cdot  (\hatP_k^m(1|1, 1) - \hatP_k^m(1|1, 2))\hat{V}_{k-1}^m\right)^2\right] \right)^{1/2} \nonumber \\
    & \leq \left(\sum_{k = 1}^T   \left( \widetilde{\eta}_k^{(T)} \right)^2 \cdot \frac{2\gamma^2 p(1-p)}{BM^2} \cdot \sum_{m = 1}^M \E\left[ \left(\hat{V}_{k-1}^m\right)^2\right] \right)^{1/2} \nonumber \\
    & \leq \left(\sum_{k = 1}^T   \left( \widetilde{\eta}_k^{(T)} \right)^2 \cdot \frac{2\gamma^2 p(1-p)}{BM(1-\gamma)^2} \right)^{1/2}. \label{eqn:delta_diff_bound}
\end{align}
Let us focus on the term involving the step sizes. We separately consider the scenario for constant step sizes and linearly rescaled step sizes. For constant step sizes, we have,
\begin{align}
    \sum_{k = 1}^T   \left( \widetilde{\eta}_k^{(T)} \right)^2 & = \sum_{k = 1}^T \left( \eta_k \prod_{i = k+1}^T (1- \eta_i) \right)^2 = \sum_{k = 1}^T \eta^2 (1- \eta)^{2(T-k)} \leq  \frac{\eta^2}{1 - (1-\eta)^2}  \leq \eta. \label{eqn:eta_squares_sum_constant}
\end{align}
Similarly, for linearly rescaled step sizes, we have,
\begin{align}
    \sum_{k = 1}^T   \left( \widetilde{\eta}_k^{(T)} \right)^2 & = \sum_{k = 1}^{\tau}   \left( \widetilde{\eta}_k^{(T)} \right)^2 + \sum_{k = \tau+1}^{T}   \left( \eta_k \prod_{i = k+1}^T (1- \eta_i) \right)^2 \nonumber \\
    & \leq \sum_{k = 1}^{\tau}  \left( \widetilde{\eta}_{\tau}^{(T)} \right)^2 + \sum_{k = \tau+1}^{T}    \eta_k^2 (1- \eta_T)^{2(T-k)}  \nonumber \\
    & \leq  \eta_{\tau}^2 (1 - \eta_T)^{2(T- \tau)} \cdot \tau +  \eta_{\tau}^2 \cdot  \frac{1}{\eta_{T}(2 -\eta_T)}  \nonumber \\
    & \leq  3\eta_T \cdot \eta_T \cdot T \cdot \exp\left(-\frac{4T\eta_T}{3}\right) + 3\eta_{T}   \nonumber \\
    & \leq  \frac{9}{4e} \eta_T + 3\eta_{T}   \nonumber \\
    & \leq  4\eta_{T}, \label{eqn:eta_squares_sum_linear}
\end{align}
where the second step uses $c_{\eta} \leq \log N \leq \frac{1}{1-\gamma}$ and the fact that $\widetilde{\eta}_{k}^{(T)}$ is increasing in $k$ in this regime. (See Eqn.~\eqref{eqn:tilde_eta_k_t_increasing}) and fifth step uses $xe^{-4x/3} \leq 3/4e$. On plugging results from Eqns.~\eqref{eqn:eta_squares_sum_constant} and~\eqref{eqn:eta_squares_sum_linear} into Eqn.~\eqref{eqn:delta_diff_bound} along with the value of $p$, we obtain,
\begin{align}
    \sup_{1 \leq t \leq T} \E[|\barDelta_t(1) - \barDelta_t(2)|] & \leq \sqrt{ \frac{8\eta_T}{3BM(1-\gamma)}},
\end{align}
as required.

\subsubsection{Proof of Lemma~\ref{lemma:V_t_lower_bound}}

For the proof, we fix an agent $m$. In order to obtain the required lower bound on $\hatV_t^m$, we define an auxiliary sequence $\barQ_t^m$ that evolves as described in Algorithm~\ref{alg:bar_Q}. Essentially, $\barQ_t^m$ evolves in a manner almost identical to $\hatQ_t^m$ except for the fact that there is only one action and hence there is no maximization step in the update rule. 

\begin{algorithm}[!h]
    \caption{Evolution of $\barQ$}
    \label{alg:bar_Q}
    \begin{algorithmic}[1]
        \STATE $r \leftarrow 1$, $\barQ^m_0 = Q^{\star}(1, 1)$ for all $m \in \{1,2,\dots, M\}$
        \FOR{$t = 1,2,\dots, T$}
        \FOR{$m = 1,2,\dots, M$}
        \STATE $\barQ^{m}_{t - 1/2} \leftarrow (1 - \eta_t)\barQ^{m}_{t - 1}(a) + \eta_t(1 + \widehat{P}_{t}^m(1| 1,1) \barQ_{t-1}^m)$
        \STATE Compute $\barQ_{t}^m$ according to Eqn.~\eqref{eqn:generic_algo_averaging}
        \ENDFOR
        \ENDFOR
    \end{algorithmic}
\end{algorithm}

It is straightforward to note that $\hatQ_t^m(1) \geq \barQ_t^m$, which can be shown using induction. From the initialization, it follows that $\hatQ_0^m(1) \geq \barQ_0^m$. Assuming the relation holds for $t - 1$, we have,
\begin{align*}
    \hatQ_{t-1/2}^m(1) & = (1 - \eta_t) \hatQ_{t-1}^m(1) + \eta_t(1 + \gamma \hatP_t^m(1|1,1) \hatV_{t-1}^m) \\
    & \geq (1 - \eta_t) \hatQ_{t-1}^m(1) + \eta_t(1 + \gamma \hatP_t^m(1|1,1) \hatQ_{t-1}^m(1)) \\
    & \geq (1 - \eta_t) \barQ_{t-1}^m + \eta_t(1 + \gamma \hatP_t^m(1|1,1) \barQ_{t-1}^m) \\
    & = \barQ_{t-1/2}^m.
\end{align*}
Since $\hatQ_t^m$ and $\barQ_t^m$ follow the same averaging schedule, it immediately follows from the above relation that $\hatQ_t^m(1) \geq \barQ_t^m$. Since $\hatV_t^m \geq  \hatQ_t^m(1) \geq \barQ_t^m$, we will use the sequence $\barQ_t^m$ to establish the required lower bound on $\hatV_t^m$. 

We claim that for all time instants $t$ and all agents $m$, 
\begin{align}
    \E[\barQ_t^m] = \frac{1}{1-\gamma p}. \label{eqn:bar_Q_expected_value_claim}
\end{align}
Assuming~\eqref{eqn:bar_Q_expected_value_claim} holds, we have
\begin{align*}
    \E[(\hatV_t^m)^2] \geq \left(\E[\hatV_t^m] \right)^2 \geq \left(\E[\barQ_t^m] \right)^2 \geq \left(\frac{1}{1-\gamma p} \right)^2 \geq \frac{1}{2(1-\gamma)^2},
\end{align*}
as required. In the above expression, the first inequality follows from Jensen's inequality, the second from the relation $\hatV_t^m \geq \barQ_t^m \geq 0$ and the third from~\eqref{eqn:bar_Q_expected_value_claim}.

We now move now to prove the claim~\eqref{eqn:bar_Q_expected_value_claim} using induction. For the base case, $\E[\barQ_0^m] = \frac{1}{1- \gamma p}$ holds by choice of initialization. Assume that $\E[\barQ_{t-1}^m] = \frac{1}{1- \gamma p}$ holds for some $t - 1$ for all $m$. 

\begin{itemize}
    \item If $t$ is not an averaging instant, then for any client $m$,
    \begin{align}
        \barQ_{t}^m & = (1 - \eta_t) \barQ_{t-1}^m + \eta_t(1 + \gamma \hatP_t^m(1|1,1) \barQ_{t-1}^m)  \nonumber \\
        \implies \E[\barQ_{t}^m] & = (1 - \eta_t) \E[\barQ_{t-1}^m] + \eta_t(1 + \gamma \E[\hatP_t^m(1|1,1) \barQ_{t-1}^m])  \nonumber \\
        & = (1 - \eta_t) \E[\barQ_{t-1}^m] + \eta_t(1 + \gamma p \E[\barQ_{t-1}^m])  \nonumber \\
        & = \frac{(1 - \eta_t)}{1-\gamma p} + \eta_t\left(1 + \frac{\gamma p}{1-\gamma p}\right)  = \frac{1}{1-\gamma p}. \label{eqn:bar_Q_non_avg_instant_case}
    \end{align}
    The third line follows from the independence of $\hatP_t^m(1|1,1)$ and $\barQ_{t-1}^m$ and the fourth line uses the inductive hypothesis.
    \item If $t$ is an averaging instant, then for all clients $m$,
    \begin{align}
    \barQ_{t}^m  & =   \frac{(1 - \eta_t)}{M} \sum_{j = 1}^M \barQ_{t-1}^j + \eta_t \frac{1}{M} \sum_{j = 1}^M (1 + \gamma \hatP_t^j(1|1,1) \barQ_{t-1}^j) \nonumber \\
    \implies \E[\barQ_{t}^m]  & =   \frac{(1 - \eta_t)}{M} \sum_{j = 1}^M \E[\barQ_{t-1}^j] + \eta_t \frac{1}{M} \sum_{j = 1}^M (1 + \gamma \E[\hatP_t^j(1|1,1) \barQ_{t-1}^j]) \nonumber \\
    & =   \frac{(1 - \eta_t)}{M} \sum_{j = 1}^M \frac{1}{1-\gamma p} + \eta_t \frac{1}{M} \sum_{j = 1}^M \left(1 + \frac{\gamma p}{1-\gamma p}\right) = \frac{1}{1-\gamma p}, \label{eqn:bar_Q_avg_instant_case}
    \end{align}
    where we again make use of independence and the inductive hypothesis.
\end{itemize}

Thus,~\eqref{eqn:bar_Q_non_avg_instant_case} and~\eqref{eqn:bar_Q_avg_instant_case} taken together complete the inductive step.

%% file: analysis_upper_bound.tex
\section{Analysis of \DVR}
\label{appendix:dvr_analysis}

In this section, we prove Theorem~\ref{thm:fed_dvr_performance} that outlines the performance guarantees of \DVR. There are two main parts of the proof. The first part deals with establishing that for the given choice of parameters described in Section~\ref{ssub:parameter_setting}, the output of the algorithm is an $\varepsilon$-optimal estimate of $\starQ$ with probability $1 - \delta$. The second part deals with deriving the bounds on the sample and communication complexity based on the choice of prescribed parameters. We begin with the second part, which is easier of the two.

\subsection{Establishing the sample and communication complexity bounds}

\paragraph{Establishing the communication complexity.} We begin with bounding $\CC_{\textsf{round}}$. From the description of \DVR, it is straightforward to note that each epoch, i.e., each call to the \RE\ routine, involves $I + 1$ rounds of communication, one for estimating $\cT\barQ$ and the remaining ones during the iterative updates of the Q-function. Since there are a total of $K$ epochs,
\begin{align*}
    \CC_{\textsf{round}}(\DVR; \varepsilon, M, \delta) \leq (I+ 1)K \leq \frac{16}{\eta(1-\gamma)} \log_2\left(\frac{1}{(1-\gamma)\varepsilon}\right),
\end{align*}
where the second bound follows from the prescribed choice of parameters in~\eqref{eq:def_K}. Similarly, since the quantization step is designed to compress each coordinate into $J$ bits, each message transmitted by an agent has a size of no more than $J \cdot |\cS| |\cA|$ bits. Consequently,
\begin{align*}
    \CC_{\textsf{bit}}(\DVR; \varepsilon, M, \delta) & \leq J \cdot |\cS| |\cA| \cdot \CC_{\textsf{round}}(\DVR; \varepsilon, M, \delta) \\
    & \leq \frac{32|\cS|\cA|}{\eta(1-\gamma)} \log_2\left(\frac{1}{(1-\gamma)\varepsilon}\right) \log_2\left(\frac{70}{\eta(1-\gamma)}\sqrt{\frac{4}{M} \log\left(\frac{8KI|\cS||\cA|}{\delta} \right)}\right),
\end{align*}
where once again in the second step we plugged in the choice of $J$ from~\eqref{eq:def_J}. 

\paragraph{Establishing the sample complexity.} In order to establish the bound on the sample complexity, note that during epoch $k$, each agent takes a total of $\lceil L_k/M\rceil + I \cdot B $ samples, where the first term corresponds to approximating $\tildeT_L(Q^{(k-1)})$ and the second term corresponds to the samples taken during the iterative update scheme. Thus, the total sample complexity is obtained by summing up over all the $K$ epochs. We have,  
\begin{align*}
    \SC(\DVR; \varepsilon, M, \delta) & \leq \sum_{k = 1}^K  \left(\left\lceil \frac{L_k}{M}\right\rceil + I \cdot B\right) 
     \leq I\cdot B \cdot K + \frac{1}{M}\sum_{k = 1}^K L_k + K .
\end{align*}
To continue, notice that
\begin{align*}    
 \frac{1}{M}\sum_{k = 1}^K L_k   & \leq   \frac{39200}{M(1-\gamma)^2} \log\left(\frac{8KI|\cS||\cA|}{\delta} \right) \left( \sum_{k = 1}^{K_0} 4^k + \sum_{k = K_0 + 1}^K 4^{k- K_0} \right)\\
    & \leq \frac{39200}{3M(1-\gamma)^2} \log\left(\frac{8KI|\cS||\cA|}{\delta} \right) \left( 4^{K_0} +  4^{K - K_0} \right)\\
    & \leq \frac{156800}{3M(1-\gamma)^2} \log\left(\frac{8KI|\cS||\cA|}{\delta} \right) \left( \frac{1}{1-\gamma} +  \frac{1}{(1-\gamma)\varepsilon^2} \right) ,
\end{align*}
where the first line follows from the choice of $L_k$ in~\eqref{eq:choice_Lk} and the last line follows from $K_0 = \lceil \frac{1}{2}\log_2 (\frac{1}{1-\gamma}) \rceil$. Plugging this relation and the choices of $I$ and $B$ (cf.~\eqref{eq:def_I} and~\eqref{eq:def_B}) into the previous bound yields
\begin{align*}    
   \SC(\DVR; \varepsilon, M, \delta)  
    & \leq \frac{4608}{\eta M(1-\gamma)^3}\log_2\left(\frac{1}{(1-\gamma)\varepsilon}\right)\log\left(\frac{8KI|\cS||\cA|}{\delta} \right) + K \\
    & ~~~~~~~~~~~~ + \frac{156800}{3M(1-\gamma)^2} \log\left(\frac{8KI|\cS||\cA|}{\delta} \right) \left( \frac{1}{1-\gamma} +  \frac{1}{(1-\gamma)\varepsilon^2} \right)\\
    & \leq \frac{313600}{\eta M(1-\gamma)^3\varepsilon^2}\log_2\left(\frac{1}{(1-\gamma)\varepsilon}\right)\log\left(\frac{8KI|\cS||\cA|}{\delta} \right) + K.
\end{align*}
Plugging in the choice of $K$ finishes the proof.

\subsection{Establishing the error guarantees}

In this section, we show that the Q-function estimate returned by the \DVR \ algorithm is $\varepsilon$-optimal with probability at least $1-\delta$. We claim that the estimates of the Q-function generated by the algorithm across different epochs satisfy the following relation for all $k \leq K$ with probability $1 - \delta$:
\begin{align}
    \|Q^{(k)} - \starQ\|_{\infty} \leq  \dfrac{2^{-k}}{1-\gamma}.
    \label{eqn:q_k_relation}
\end{align}
The required bound on $\|Q^{(K)} - \starQ\|_{\infty}$ immediately follows by plugging in the value of $K$. Thus, for the remainder of the section, we focus on establishing the above claim. 

\paragraph{Step 1: fixed-point contraction of \RE.} Firstly, note that the variance-reduced update scheme carried out during the \RE\ routine resembles that of the classic Q-learning scheme, i.e., fixed-point iteration, with a different operator defined as follows:
\begin{align}
    \cH(Q) := \cT(Q) - \cT(\barQ) + \tildeT_L(\barQ), \quad \mbox{for some fixed } \barQ.  \label{eqn:operator_h_definition}
\end{align}
Thus, the update scheme at step $i \geq 1$ in \eqref{eqn:q_l_update_rule} can then be written as
\begin{align}
    Q_{i- \frac{1}{2}}^{m} = (1-\eta) Q_{i-1} + \eta \widehat{\cH}_i^{(m)}(Q_{i-1}), \label{eqn:q_l_update_rule_with_h}
\end{align}
where $\widehat{\cH}_i^{(m)}(Q) := \hatT_i^{(m)}(Q) - \hatT_i^{(m)}(\barQ) + \tildeT_L(\barQ)$ is a stochastic, unbiased estimate of the operator $\cH$, similar to $\hatT_i^{(m)}(Q)$. Let $Q_{\cH}^{\star}$ denote the fixed point of $\cH$. Then the update scheme in \eqref{eqn:q_l_update_rule_with_h} drives the sequence $\{Q_{i}^m\}_{i\geq 0}$ to $Q_{\cH}^{\star}$; further, as long as $\|Q^{\star} - Q_{\cH}^{\star}\|_{\infty}$ is small, the required error $\|Q_i - Q^{\star}\|_{\infty}$ can also be controlled. The following lemmas formalize these ideas and pave the path to establish the claim in \eqref{eqn:q_k_relation}. The proofs are deferred to Appendix~\ref{sec:aux_lemma_proofs_fed_dvr}. 
 
\begin{lemma}
Let $\delta \in (0,1)$.  Consider the \normalfont{\RE} routine described in Algorithm~\ref{alg:refine_estimate} and let $Q_{\cH}^{\star}$ denote the fixed point of the operator $\cH$ defined in \eqref{eqn:operator_h_definition} for some fixed $\barQ$.  Then the iterates generated by \RE \ $Q_I$ satisfy
    \begin{align*}
        \|Q_I - \starQH\|_{\infty} & \leq \frac{1}{6} \left(\| \barQ - \starQ\|_{\infty} + \| \starQ - \starQH\|_{\infty} \right) + \frac{D}{70}  
    \end{align*}
    with probability $1- \frac{\delta}{2K}$. 
    \label{lemma:q_l_star_q_h_err}
\end{lemma}

\begin{lemma}
    Consider the \normalfont{\RE} routine described in Alg.~\ref{alg:refine_estimate} and let $Q_{\cH}^{\star}$ denote the fixed point of the operator $\cH$ defined in Eqn.~\eqref{eqn:operator_h_definition} for a fixed $\barQ$. The following relation holds with probability $1- \frac{\delta}{2K}$:
    \begin{align*}
        \|\starQH - \starQ\|_{\infty} \leq \| \barQ - \starQ\|_{\infty} \cdot \sqrt{\frac{16\kappa'}{L(1-\gamma)^2} } +  \sqrt{\frac{64\kappa'}{L(1-\gamma)^3}} + \frac{2\kappa'\sqrt{2}}{3L(1- \gamma)^2} + \frac{D}{70},
    \end{align*}
    whenever $L \geq 32\kappa'$, where $\kappa' = \log\left(\frac{12K|\cS||\cA|}{\delta} \right)$.
    \label{lemma:star_q_h_star_q_err}
\end{lemma}

\paragraph{Step 2: establishing the linear contraction.}  We now leverage the above lemmas to establish the desired contraction in \eqref{eqn:q_k_relation}. Instantiating the operator \eqref{eqn:operator_h_definition} at each $k$-th epoch by setting $\barQ := Q^{(k-1)}$ and $L :=L_k$,  we define
\begin{align}
    \cH_k(Q) := \cT(Q) - \cT(Q^{(k-1)}) + \tildeT_{L_k}(Q^{(k-1)}),   \label{eqn:operator_hk_definition}
\end{align}
whose fixed point is denoted as $\starQ_{\cH_k}$.
Using the results from Lemmas~\ref{lemma:q_l_star_q_h_err} and~\ref{lemma:star_q_h_star_q_err} with $D := D_k$ and $\cH = \cH_k$, we obtain
\begin{align}
    \|Q^{(k)} - \starQ\|_{\infty} & \leq \|Q^{(k)} - \starQ_{\cH_k}\|_{\infty} + \|\starQH - \starQ_{\cH_k}\|_{\infty} \nonumber \\
    & \leq \frac{1}{6} \left(\| Q^{(k-1)} - \starQ\|_{\infty} + \| \starQ - \starQ_{\cH_k}\|_{\infty} \right) + \frac{D_k}{70} + \|\starQ_{\cH_k} - \starQ\|_{\infty} \nonumber \\
    & = \frac{1}{6} \left(\| Q^{(k-1)} - \starQ\|_{\infty} + 7\| \starQ - \starQ_{\cH_k}\|_{\infty} \right) + \frac{D_k}{70} \nonumber \\
    & \leq \| Q^{(k-1)} - \starQ\|_{\infty}  \left( \frac{1}{6} + \frac{7}{6}\sqrt{\frac{16\kappa'}{L_k(1-\gamma)^2} } \right) + \frac{7}{6}\left(\sqrt{\frac{64\kappa'}{L_k(1-\gamma)^3}} + \frac{2\sqrt{2}\kappa'}{3L_k(1- \gamma)^2}\right)+ \frac{13D_k}{420} \nonumber  \\
    & \leq \| Q^{(k-1)} - \starQ\|_{\infty}  \left( \frac{1}{6} + \frac{7}{6}\sqrt{\frac{16\kappa'}{L_k(1-\gamma)^2} } \right) + \frac{7}{6} \sqrt{\frac{100\kappa'}{L_k(1-\gamma)^3}}+ \frac{13D_k}{420}, \label{eqn:q_k_err_recursive_relation}
\end{align}
holds with probability $1- \frac{\delta}{K}$. Here, we invoke Lemma~\ref{lemma:q_l_star_q_h_err} in the second step and Lemma~\ref{lemma:star_q_h_star_q_err} in the fourth step corresponding to the \normalfont{\RE} routine during the $k$-th epoch. In the last step, we used the fact that $\frac{L_k(1-\gamma)^2}{\kappa'} \geq 1$. 

We now use induction along with the recursive relation in \eqref{eqn:q_k_err_recursive_relation} to establish the required claim \eqref{eqn:q_k_relation}. Let us first consider the case $0 \leq k \leq K_0$. The base case, $\|Q^{(0)} - \starQ\|_{\infty} \leq \frac{1}{1-\gamma}$, holds by definition. Let us assume the relation holds for $k - 1$. Then, from \eqref{eqn:q_k_err_recursive_relation} and \eqref{eq:choice_Lk}, we have
\begin{align}
    \|Q^{(k)} - \starQ\|_{\infty} & \leq \| Q^{(k-1)} - \starQ\|_{\infty}  \left( \frac{1}{6} + \frac{7}{6}\sqrt{\frac{16\kappa'}{L_k(1-\gamma)^2} } \right) + \frac{7}{6} \sqrt{\frac{100\kappa'}{L_k(1-\gamma)^3}}+ \frac{13D_k}{420} \nonumber \\
    & \leq \frac{2^{-(k-1)}}{1-\gamma}  \left( \frac{1}{6} + 2^{-k} \cdot \frac{7}{6}\sqrt{\frac{8}{19600} } \right) + 2^{-k} \cdot \frac{7}{6} \sqrt{\frac{50}{19600(1-\gamma)}}+ \frac{104}{420} \cdot \frac{2^{-(k-1)}}{1-\gamma} \nonumber \\
    & \leq \frac{2^{-(k-1)}}{1-\gamma}  \left( \frac{1}{6} +  \frac{7}{6}\sqrt{\frac{91}{39200} } + \frac{1}{4} \right) \nonumber \\
    & \leq \frac{2^{-k}}{1-\gamma}. \label{eqn:q_k_induction_case_1}
\end{align}
Now we move to the second case, for $k > K_0$.
From \eqref{eqn:q_k_err_recursive_relation} and \eqref{eq:choice_Lk}, we have
\begin{align}
    \|Q^{(k)} - \starQ\|_{\infty} & \leq \| Q^{(k-1)} - \starQ\|_{\infty}  \left( \frac{1}{6} + \frac{7}{6}\sqrt{\frac{16\kappa'}{L_k(1-\gamma)^2} } \right) + \frac{7}{6} \sqrt{\frac{100\kappa'}{L_k(1-\gamma)^3}}+ \frac{13D_k}{420} \nonumber \\
    & \leq \frac{2^{-(k-1)}}{1-\gamma}  \left( \frac{1}{6} + 2^{-(k-K_0)} \cdot \frac{7}{6}\sqrt{\frac{8}{19600} } \right) + 2^{-(k-K_0)} \cdot \frac{7}{6} \sqrt{\frac{50}{19600(1-\gamma)}}+ \frac{104}{420} \cdot \frac{2^{-(k-1)}}{1-\gamma} \nonumber \\
    & \leq \frac{2^{-(k-1)}}{1-\gamma}  \left( \frac{1}{6} + \frac{7}{6}\sqrt{\frac{1}{196} } + \frac{1}{4} \right) \nonumber \\
    & \leq \frac{2^{-k}}{1-\gamma}. \label{eqn:q_k_induction_case_2}
\end{align}
By a union bound argument, we can conclude that the relation $\|Q^{(k)} - \starQ\|_{\infty} \leq \frac{2^{-k}}{1-\gamma}$ holds for all $k \leq K$ with probability at least $1- \delta$. 

\paragraph{Step 3: confirm the compressor bound.}
The only thing left to verify is that the inputs to the compressor are always bounded by $D_k$ during the $k$-th epoch, for all $1\leq k\leq K$. The following lemma provides a bound on the input to the compressor during any run of the \RE \ routine.
\begin{lemma}
    Consider the \normalfont{\RE} routine described in Algorithm~\ref{alg:refine_estimate} with some  for some fixed $\barQ$. For all $i \leq I$ and all agents $m$, the following bound holds with probability $1- \frac{\delta}{2K}$:
    \begin{align*}
        \|Q_{i -\frac{1}{2}}^{m} - Q_{i-1}\|_{\infty} \leq \eta\|\barQ- \starQH\|_{\infty}\left( \frac{7}{6} \cdot (1+\gamma)  + 2\gamma \right) + \frac{\eta D(1+\gamma)}{70}.
    \end{align*}
    \label{lemma:compressor_input_bound}
\end{lemma}

For the $k$-th epoch, it follows that
\begin{align*}
    \eta\|Q^{(k-1)}- \starQ_{\cH_k}\|_{\infty}\left( \frac{7}{6} \cdot (1+\gamma)  + 2\gamma \right) + \frac{\eta D_k(1+\gamma)}{70} & \leq \frac{13}{3} \left(\|Q^{(k-1)}- \starQ\|_{\infty} + \|\starQ- \starQ_{\cH_k}\|_{\infty} \right) + \frac{ D_k(1+\gamma)}{70} \\
    & \leq \frac{13}{3} \cdot \frac{15}{14} \cdot \|Q^{(k-1)}- \starQ\|_{\infty}  + \frac{2D_k}{70} \\
    & \leq \left(\frac{195}{42}  + \frac{16}{70}\right) \cdot \frac{2^{-(k-1)}}{1-\gamma} \\
    & \leq 8 \cdot \frac{2^{-(k-1)}}{1-\gamma} := D_k.
\end{align*}
In the third step, we used the same sequence of arguments as used in \eqref{eqn:q_k_induction_case_1} and~\eqref{eqn:q_k_induction_case_2} and, in the fourth step, we used the bound on $\|Q^{(k-1)}- \starQ\|_{\infty}$ from \eqref{eqn:q_k_relation} and the prescribed value of $D_k$.



\subsection{Proof of auxiliary lemmas}
\label{sec:aux_lemma_proofs_fed_dvr}

\subsubsection{Proof of Lemma~\ref{lemma:q_l_star_q_h_err}}

Let us begin with analyzing the evolution of the sequence $\{Q_i\}_{i= 1}^I$ during a run of the \RE \ routine. The sequence $\{Q_i\}_{i= 1}^I$ satisfies the following recursion:
\begin{align}
    Q_i & = Q_{i-1} + \frac{1}{M} \sum_{m = 1}^M \sC\left( Q_{i- \frac{1}{2}}^{m} - Q_{i-1}; D, J \right ) \nonumber \\
    & = Q_{i-1} + \frac{1}{M} \sum_{m = 1}^M \left( Q_{i- \frac{1}{2}}^{m} - Q_{i-1} + \zeta_i^{m} \right) \nonumber \\
    & = \frac{1}{M} \sum_{m = 1}^M \left( Q_{i- \frac{1}{2}}^{m} + \zeta_i^{m} \right)   = (1-\eta)Q_{i-1} + \frac{\eta}{M} \sum_{m = 1}^M \hatH_i^{(m)}(Q_{i-1})  + \underbrace{\frac{1}{M} \sum_{m = 1}^M \zeta_i^{m}}_{ =: \zeta_i }. \label{eqn:q_l_recursive_update}
\end{align}
In the above expression, $\zeta_i^{m}$ denotes the quantization noise introduced at agent $m$ in the $i$-th update.

Subtracting $\starQH$ from both sides of \eqref{eqn:q_l_recursive_update}, we obtain
\begin{align}
    Q_i - \starQH & = (1-\eta)(Q_{i-1} - \starQH) + \frac{\eta}{M} \sum_{m = 1}^M \left(\hatH_i^{(m)}(Q_{i-1}) - \starQH \right) + \zeta_i \nonumber \\
    & = (1-\eta)(Q_{i-1} - \starQH) + \frac{\eta}{M} \sum_{m = 1}^M \left(\hatH_i^{(m)}(Q_{i-1}) - \hatH_i^{(m)}(\starQH) \right) \nonumber \\
    & ~~~~~~~~~~~~~~~~~~~~~~~~~~~~~~~~~~~~~~~~~~~~~~~~~~~~~~~~~~~~ + \frac{\eta}{M} \sum_{m = 1}^M \left(\hatH_i^{(m)}(\starQH) - \cH(\starQH) \right) + \zeta_i. \label{eqn:q_l_minus_star_q_h}
\end{align}
Consequently, 
\begin{align}
    \|Q_i - \starQH\|_{\infty} & \leq (1-\eta)\|Q_{i-1} - \starQH\|_{\infty} + \frac{\eta}{M} \sum_{m = 1}^M \left\|\hatH_i^{(m)}(Q_{i-1}) - \hatH_i^{(m)}(\starQH) \right\|_{\infty} \nonumber \\
    & ~~~~~~~~~~~~~~~~~~~~~~~~~~~~~~~~~~~~~~~~~~~~ + \left\| \frac{\eta}{M} \sum_{m = 1}^M \left(\hatH_i^{(m)}(\starQH) - \cH(\starQH) \right) \right\|_{\infty} + \left\|\zeta_i\right\|_{\infty}  , \label{eqn:Q_l_recursion_inf_norm}
\end{align}
which we shall proceed to bound each term separately.
\begin{itemize}
\item Regarding the second term, it follows that
\begin{align} \label{eq:contraction_hatH}
    \left\|\hatH_i^{(m)}(Q) - \hatH_i^{(m)}(\starQH)\right\|_{\infty} = \left\| \hatT_i^{(m)}(Q) - \hatT_i^{(m)}(\starQH) \right\|_{\infty} \leq \gamma \left\| Q - \starQH \right\|_{\infty},
\end{align}
which holds for all $Q$ since $\hatT_i^{(m)}$ is a $\gamma$-contractive operator.
\item Regarding the third term, notice that 
\begin{align*}
    \frac{1}{M} \sum_{m = 1}^M \left(\hatH_i^{(m)}(\starQH) - \cH(\starQH) \right) = \frac{1}{MB} \sum_{m = 1}^M \sum_{z \in \cZ_i^{(m)}} \left(\cT_z(\starQH) - \cT_z(\barQ) - \cT(\starQH) + \cT(\barQ) \right).
\end{align*}
Note that $\cT_z(\starQH) - \cT_z(\barQ) - \cT(\starQH) + \cT(\barQ)$ is a zero-mean random vector satisfying 
\begin{equation} \label{eq:empiral_H_dev}
\|\cT_z(\starQH) - \cT_z(\barQ) - \cT(\starQH) + \cT(\barQ)\|_{\infty} \leq 2\gamma \| \barQ - \starQH\|_{\infty}.
\end{equation}
Thus, each of its coordinate is a $(2\gamma \| \barQ - \starQH\|_{\infty})^2$-sub-Gaussian vector. Applying the tail bounds for a maximum of sub-Gaussian random variables \citep{vershynin2018high}, we obtain that 
\begin{align}
    \left\|\frac{1}{M} \sum_{m = 1}^M \left(\hatH_i^{(m)}(\starQH) - \cH(\starQH) \right) \right\|_{\infty} \leq 2\gamma \| \barQ - \starQH\|_{\infty} \cdot \sqrt{\frac{2}{MB} \log\left(\frac{8KI|\cS||\cA|}{\delta} \right)} \label{eqn:H_err_bound}
\end{align}
holds with probability at least $1 - \frac{\delta}{4KI}$. 

\item Turning to the last term, by the construction of the compression routine described in Section~\ref{ssub:quantization}, it is straightforward to note that $\zeta_i^{m}$ is a zero-mean random vector whose coordinates are independent, $D^2 \cdot 4^{-J}$-sub-Gaussian random variables. Thus, $\zeta_i$ is also a zero-mean random vector whose coordinates are independent, $\frac{D^2}{M \cdot 4^{J}}$-sub-Gaussian random variables. Hence, we can similarly conclude that
\begin{align}
    \|\zeta_i\|_{\infty} \leq D \cdot 2^{-J} \cdot \sqrt{\frac{2}{M} \log\left(\frac{8KI|\cS||\cA|}{\delta} \right)} \label{eqn:zeta_err_bound}
\end{align}
holds with probability at least $1 - \frac{\delta}{4KI}$.
\end{itemize}

Combining the above bounds into \eqref{eqn:Q_l_recursion_inf_norm}, and introducing the short-hand notation $\kappa := \log\left(\frac{8KI|\cS||\cA|}{\delta} \right)$, we obtain with probability at least $1- \frac{\delta}{2KI}$, 
\begin{align*}
    \|Q_i - \starQH\|_{\infty} & \leq (1-\eta(1-\gamma))\|Q_{i-1} - \starQH\|_{\infty}  + 2\eta \gamma \| \barQ - \starQH\|_{\infty} \cdot \sqrt{\frac{2\kappa}{MB}}  + D \cdot 2^{-J} \cdot \sqrt{\frac{2\kappa}{M}}. 
\end{align*}
Unrolling the above recursion over $i=1,\ldots, I$ yields the following relation, which holds with probability at least $1 - \frac{\delta}{2K}$:
\begin{align}
    \|Q_I - \starQH\|_{\infty} & \leq \left(1-\eta(1-\gamma)\right)^{I}\|Q_{0} - \starQH\|_{\infty}  +   \sqrt{\frac{2\kappa}{M}} \left(\frac{2\eta \gamma}{\sqrt{B}} \| \barQ - \starQH\|_{\infty}  + D \cdot 2^{-J} \right) \cdot \sum_{i = 1}^{I} \left(1-\eta(1-\gamma)\right)^{I - i} \nonumber\\
    & \leq \left(1-\eta(1-\gamma)\right)^{I}\|\barQ - \starQH\|_{\infty}  +  \frac{1}{\eta(1-\gamma)} \sqrt{\frac{2\kappa}{M}} \left(\frac{2\eta \gamma}{\sqrt{B}} \| \barQ - \starQH\|_{\infty}  + D \cdot 2^{-J} \right)  \nonumber\\
    & \leq \|\barQ - \starQH\|_{\infty} \left(\left(1-\eta(1-\gamma)\right)^{I} + \frac{2 \gamma}{(1-\gamma)}\sqrt{\frac{2\kappa}{MB}} \right)  +  \frac{D \cdot 2^{-J}}{\eta(1-\gamma)} \cdot \sqrt{\frac{2\kappa}{M}}   \label{eqn:Q_L_inf_norm_orig_bound_step_3}\\
    & \leq \frac{\|\barQ - \starQH\|_{\infty}}{6} + \frac{D}{70}  
     \leq \frac{1}{6} \left(\| \barQ - \starQ\|_{\infty} + \| \starQ - \starQH\|_{\infty} \right) + \frac{D}{70}.
    \label{eqn:Q_L_inf_norm_orig_bound}
\end{align} 
Here, the fourth step is obtained by plugging in the prescribed values of $B, I$ and $J$ in \eqref{eq:def_IBJ}.

\subsubsection{Proof of Lemma~\ref{lemma:star_q_h_star_q_err}}
\label{proof:star_q_h_star_q_err}
Intuitively, the error $\|\starQH - \starQ\|_{\infty}$ depends on the error term $\tildeT_L(\barQ) - \cT(\barQ)$. If the latter is small, then $\cH(Q)$ is close to $\cT(Q)$ and consequently so are $\starQH$ and $\starQ$. Thus, we begin with bounding the term $\tildeT_L(\barQ) - \cT(\barQ)$. 
We have,
\begin{align}
  &  \tildeT_L(\barQ) - \cT(\barQ) \nonumber \\
   & = \overline{Q} + \frac{1}{M} \sum_{m = 1}^M \sC\left( \tildeT^{(m)}_L(\overline{Q}) - \barQ \right) - \cT(\barQ) \nonumber \\
    & =\frac{1}{M} \sum_{m = 1}^M \left( \tildeT^{(m)}_L(\overline{Q}) + \tilde{\zeta}_L^{(m)} \right)  - \cT(\barQ) \nonumber  \\
    & =\frac{1}{M} \sum_{m = 1}^M \left( \tildeT^{(m)}_L(\overline{Q}) - \tildeT^{(m)}_L(\starQ) - \cT(\barQ) + \cT(\starQ) \right)  + \frac{1}{M} \sum_{m = 1}^M \tilde{\zeta}_L^{(m)} +  \frac{1}{M} \sum_{m = 1}^M \left( \tildeT^{(m)}_L(\starQ) -  \cT(\starQ) \right), \label{eqn:tilde_T_N_err_total}
\end{align}
where once again $\tilde{\zeta}_L^{(m)}:=   \tildeT^{(m)}_L(\overline{Q}) - \barQ  -  \sC\left( \tildeT^{(m)}_L(\overline{Q}) - \barQ \right)  $ denotes the quantization error at agent $m$.
Similar to the arguments of \eqref{eqn:H_err_bound} and~\eqref{eqn:zeta_err_bound}, we can conclude that each of the following relations hold with probability at least $1 - \frac{\delta}{6K}$:
\begin{align}
    \left\|\frac{1}{M} \sum_{m = 1}^M \left( \tildeT^{(m)}_L(\overline{Q}) - \tildeT^{(m)}_L(\starQ) - \cT(\barQ) + \cT(\starQ) \right)\right\|_{\infty} & \leq 2\gamma \| \barQ - \starQ\|_{\infty} \cdot \sqrt{\frac{2}{L} \log\left(\frac{12K|\cS||\cA|}{\delta} \right)}, \label{eqn:tilde_T_N_err_term_1} \\
    \left\|\frac{1}{M} \sum_{m = 1}^M \tilde{\zeta}_L^{(m)}\right\|_{\infty} & \leq D \cdot 2^{-J} \cdot \sqrt{\frac{2}{M} \log\left(\frac{12K|\cS||\cA|}{\delta} \right)}.
    \label{eqn:tilde_T_N_err_term_2}
\end{align}
For the third term, we can rewrite it as
\begin{align*}
    \frac{1}{M} \sum_{m = 1}^M \left( \tildeT^{(m)}_L(\starQ) -  \cT(\starQ) \right) = \frac{1}{M \lceil L/M \rceil} \sum_{m = 1}^M \sum_{l = 1}^{\lceil L/M \rceil} \left( \cT_{Z_l^{(m)}}(\starQ) -  \cT(\starQ) \right).
\end{align*}
We will use Bernstein inequality element wise to bound the above term. Let $\bfsigma^{\star} \in \R^{|\cS| \times |\cA|}$ be such that $[\bfsigma^{\star}(s,a)]^2 = \var(\cT_{Z}(\starQ)(s,a))$, i.e., $(s,a)$-th element of $\bfsigma$ denotes the standard deviation of the random variable $\cT_{Z}(\starQ)(s,a)$. Since $\|\cT_{Z}(\starQ) -  \cT(\starQ)\|_{\infty} \leq \frac{1}{1-\gamma}$ a.s., Bernstein inequality gives us that
\begin{align}
    \left|\frac{1}{M} \sum_{m = 1}^M \left( \tildeT^{(m)}_L(\starQ)(s,a) -  \cT(\starQ)(s,a) \right) \right| \leq \bfsigma^{\star}(s,a) \sqrt{\frac{2}{L}\log\left(\frac{6K|\cS||\cA|}{\delta} \right)} + \frac{2}{3L(1- \gamma)} \log\left(\frac{6K|\cS||\cA|}{\delta} \right). \label{eqn:tilde_T_N_err_term_3}
\end{align}
holds simultaneously for all $(s,a) \in \cS \times \cA$ with probability at least $1 - \frac{\delta}{6K}$. 
On combining \eqref{eqn:tilde_T_N_err_total},~\eqref{eqn:tilde_T_N_err_term_1},~\eqref{eqn:tilde_T_N_err_term_2} and~\eqref{eqn:tilde_T_N_err_term_3}, we obtain that 
\begin{align}
    \left|\tildeT_L(\barQ)(s,a) - \cT(\barQ)(s,a) \right| & =\| \barQ - \starQ\|_{\infty}  \cdot \sqrt{\frac{8\kappa'}{L} } + \bfsigma^{\star}(s,a) \sqrt{\frac{2\kappa'}{L}} + \frac{2\kappa'}{3L(1- \gamma)} + D \cdot 2^{-J} \cdot \sqrt{\frac{2\kappa'}{M}},  \label{eqn:tilde_T_N_err_final_bound}
\end{align}
holds simultaneously for all $(s,a) \in \cS \times \cA$ with probability at least $1 - \frac{\delta}{2K}$, where $\kappa' = \log\left(\frac{12K|\cS||\cA|}{\delta} \right)$. We use this bound in~\eqref{eqn:tilde_T_N_err_final_bound} to obtain a bound on  $\|\starQH - \starQ\|_{\infty}$ using the following lemma. 

\begin{lemma}[\cite{Wainwright2019VarianceReduced}]
    Let $\pi^{\star}$ and $\pi^{\star}_{\cH}$ respectively denote the optimal policies w.r.t. $\starQ$ and $\starQH$. Then,
    \begin{align*}
        \|\starQH - \starQ\|_{\infty} \leq \max\left\{(I - \gamma P^{\pi^{\star}})^{-1}\left|\tildeT_L(\barQ) - \cT(\barQ) \right|, (I - \gamma P^{\pi^{\star}_{\cH}})^{-1}\left|\tildeT_L(\barQ) - \cT(\barQ) \right| \right\}.
    \end{align*}
    Here, for any deterministic policy $\pi$, $P^{\pi} \in \R^{|\cS||\cA| \times |\cS||\cA|}$ is given by $(P^{\pi}Q)(s,a) = \sum_{s' \in \cS}P(s'|s,a)Q(s', \pi(s'))$.\label{lemma:wainwright_err_bound}
\end{lemma}
Furthermore, it was shown in~\citet[Proof of Lemma 4]{Wainwright2019VarianceReduced} that if the error $|\tildeT_L(\barQ)(s,a) - \cT(\barQ)(s,a)|$ satisfies
\begin{align}
    \left|\tildeT_L(\barQ)(s,a) - \cT(\barQ)(s,a) \right|  \leq z_0 \| \barQ - \starQ\|_{\infty} + z_1 \bfsigma^{\star}(s,a) + z_2 \label{eqn:tilde_T_N_err_final_bound_parametric}
\end{align}
for some $z_0, z_1, z_2 \geq 0$ with $z_1 < 1$, then the bound in Lemma~\ref{lemma:wainwright_err_bound} can be simplified to
\begin{align}
    \|\starQH - \starQ\|_{\infty} \leq \frac{1}{1-z_1} \left( \frac{z_0}{1-\gamma} \| \barQ - \starQ\|_{\infty} + \frac{z_1}{(1-\gamma)^{3/2}}  + \frac{z_2}{1-\gamma} \right). \label{eqn:Q_H_Q_star_error_bound_parametric}
\end{align}
On comparing,~\eqref{eqn:tilde_T_N_err_final_bound} with~\eqref{eqn:tilde_T_N_err_final_bound_parametric}, we obtain
\begin{align*}
    z_0 \equiv \sqrt{\frac{8\kappa'}{L}}; \quad z_1 \equiv \sqrt{\frac{2\kappa'}{L}}; \quad z_2 \equiv \frac{2\kappa'}{3L(1- \gamma)} + D \cdot 2^{-J} \cdot \sqrt{\frac{2\kappa'}{M}}.
\end{align*}
Moreover, the condition $L \geq 32\kappa'$ implies that $z_1 < 1$ and $\frac{1}{1 -z_1}\leq \sqrt{2}$. Thus, on plugging in the above values in~\eqref{eqn:Q_H_Q_star_error_bound_parametric}, we can conclude that
\begin{align}
    \|\starQH - \starQ\|_{\infty} & \leq \| \barQ - \starQ\|_{\infty} \cdot \sqrt{\frac{16\kappa'}{L(1-\gamma)^2} } +  \sqrt{\frac{64\kappa'}{L(1-\gamma)^3}} + \frac{2\kappa'\sqrt{2}}{3L(1- \gamma)^2} + \frac{D \cdot 2^{-J}}{(1-\gamma)} \cdot \sqrt{\frac{4\kappa'}{M}} \nonumber \\
    & \leq \| \barQ - \starQ\|_{\infty} \cdot \sqrt{\frac{8\kappa'}{L(1-\gamma)^2} } +  \sqrt{\frac{32\kappa'}{L(1-\gamma)^3}} + \frac{2\sqrt{2}\kappa'}{3L(1- \gamma)^2} + \frac{D}{40},
    \label{eqn:Q_H_Q_star_error_bound}
\end{align}
where once again we use the value of $J$ in the last step.


\subsubsection{Proof of Lemma~\ref{lemma:compressor_input_bound}}

From the iterative update rule in \eqref{eqn:q_l_update_rule_with_h}, for any agent $m$ we have,
\begin{align*}
    Q_{i - \frac{1}{2}}^{m} - Q_{i-1} & = \eta(\hatH_{i-1}^{(m)}(Q_{i-1}) - Q_{i-1}) \\
    & = \eta(\hatH_{i-1}^{(m)}(Q_{i-1}) - \hatH_{i-1}^{(m)}(\starQH) + \hatH_{i-1}^{(m)}(\starQH) - \cH(\starQH) + \starQH - Q_{i-1}).
\end{align*}
Thus,
\begin{align*}
    \|Q_{i - \frac{1}{2}}^{m} - Q_{i-1}\|_{\infty} & \leq \eta\left( \|\hatH_{i-1}^{(m)}(Q_{i-1}) - \hatH_{i-1}^{(m)}(\starQH)\|_{\infty} + \|\hatH_{i-1}^{(m)}(\starQH) - \cH(\starQH)\|_{\infty} + \|\starQH - Q_{i-1}\|_{\infty}\right) \\
    & \leq \eta\left( \gamma \|Q_{i-1} - \starQH\|_{\infty} + 2\gamma\|\barQ- \starQH\|_{\infty} + \|\starQH - Q_{i-1}\|_{\infty}\right) \\
    & = \eta\left( (1+\gamma) \|Q_{i-1} - \starQH\|_{\infty} + 2\gamma\|\barQ- \starQH\|_{\infty} \right) \\
    & \leq \eta\|\barQ- \starQH\|_{\infty}\left( \frac{7}{6} \cdot (1+\gamma)  + 2\gamma \right) + \frac{\eta D(1+\gamma)}{70},
\end{align*}
holds with probability $1- \frac{\delta}{2KI}$. Here, the second inequality follows from \eqref{eq:contraction_hatH} and \eqref{eq:empiral_H_dev},
The last step in the above relation follows from \eqref{eqn:Q_L_inf_norm_orig_bound_step_3} evaluated at a general value of $i$ and the prescribed value of $J$. By a union bound argument, the above relation holds for all $i$ with probability at least $1- \frac{\delta}{2K}$.